\documentclass{article}
\pdfoutput=1

\PassOptionsToPackage{numbers, compress}{natbib}

\usepackage{style}
 
\usepackage[final]{neurips_2022}

\usepackage[utf8]{inputenc} %
\usepackage[T1]{fontenc}    %
\usepackage{hyperref}       %
\usepackage{url}            %
\usepackage{booktabs}       %
\usepackage{amsfonts}       %
\usepackage{nicefrac}       %
\usepackage{microtype}      %
\usepackage{xcolor}         %
\usepackage{comment}
\usepackage{wrapfig}
\usepackage{caption}
\usepackage{titletoc}

\author{%
  Christina Baek$^1$ \quad Yiding Jiang$^1$ \quad Aditi Raghunathan$^1$ \quad Zico Kolter$^{1,2}$\\
  $^1$Carnegie Mellon University, $^2$Bosch Center for AI \\ 
  \texttt{\{kbaek, yidingji, raditi, zkolter\}@cs.cmu.edu} 
} 

\begin{document}
\newcommand{\cb}[1]{{\color{blue} [Christina: #1]}}
\newcommand{\yj}[1]{{\color{purple} [Yiding: #1]}}
\newcommand{\ar}[1]{{\color{red} [AR: #1]}}
\newcommand{\zk}[1]{{\color{brown} [Zico: #1]}}

\title{Agreement-on-the-Line: Predicting the Performance of Neural Networks under Distribution Shift}
\maketitle

\begin{abstract}
Recently, \citet{Miller1} showed that a model's in-distribution (ID) accuracy has a strong linear correlation with its out-of-distribution (OOD) accuracy on several OOD benchmarks --- a phenomenon they dubbed ``accuracy-on-the-line''.  While a useful tool for model selection (i.e., the models with better ID accuracy are likely to have better OOD accuracy), this fact does not help estimate the \emph{actual} OOD performance of models without access to a labeled OOD validation set. In this paper, we show a similar but surprising phenomenon also holds for the \emph{agreement} between pairs of neural network classifiers: whenever accuracy-on-the-line holds, we observe that the OOD agreement between the predictions of \emph{any} two pairs of neural networks (with potentially different architectures) also observes a strong linear correlation with their ID agreement. Furthermore, we observe that the slope and bias of OOD vs. ID agreement closely matches that of OOD vs. ID accuracy. This phenomenon, which we call \textit{agreement-on-the-line}, has important practical applications: without any labeled data, we can \emph{predict the OOD accuracy of classifiers}, since OOD agreement can be estimated with just unlabeled data. Our prediction algorithm outperforms previous methods both in shifts where agreement-on-the-line holds and, surprisingly, when accuracy is not on the line. This phenomenon also provides new insights into deep neural networks: unlike accuracy-on-the-line, agreement-on-the-line appears to only hold for neural network classifiers.
\end{abstract}

\section{Introduction}
Machine learning operates well when models observe and make decisions on data coming from the same distribution as the training data. Yet in the real world, this assumption rarely holds. Environments are never fully controlled. Robots interact with their surroundings, effectively changing what they see in the future. Self-driving cars face constant distribution shift when driving to new cities under changing weather conditions. Models trained on clinical data from one hospital face challenges when deployed for a different hospital with different subpopulations. Under these premises, practitioners constantly face the problem of estimating a model's performance on new data distributions (\textit{out-of-distribution}, or OOD) that are related to but different from the data distribution that the model was trained on (\textit{in-distribution}, or ID). Depending on the distribution shift, models may sometimes break catastrophically under new conditions, or may only suffer a small degradation in performance. Differentiating between such cases is crucial in practice.

Assessing OOD performance is difficult because in reality, labeled OOD data is often very costly to obtain. On the other hand, \textit{unlabeled} OOD data is much easier to obtain. A natural question is whether we can leverage \emph{unlabeled} OOD data for estimating the OOD performance. This paradigm of using unlabeled data to predict the OOD generalization performance has received much attention recently \cite{garg2022leveraging, chen2021detecting, yu2022predicting,dengzheng21, Deng21, Chen21, Guillory21}. Although many different metrics have been proposed, their success varies widely depending on the shift and the ID performance of the model. While it is impossible for a method to always work with no assumptions \cite{garg2022leveraging}, a major obstacle to using these methods is that there is currently no understanding of when they work or a recipe to detect when their predictions might be unreliable. 

In a separate but related line of work, \citet{Miller1} demonstrated that in a wide variety of common OOD prediction benchmarks such as CIFAR-10.1~\cite{recht_cifar10.1andimagenetv2}, ImageNetV2~\cite{recht_cifar10.1andimagenetv2}, CIFAR-10C~\cite{hendrycks_cifar10c}, fMoW-\textsc{wilds}~\cite{christie2018functional}, there exists an almost perfect positive linear correlation between the ID test vs. OOD accuracy of models. When this phenomenon, called \textit{accuracy-on-the-line}, occurs, improving performance on the ID test data directly leads to improvements on OOD performance. Furthermore, if we have access to the slope and bias of this correlation, predicting OOD accuracy becomes  straightforward. Unfortunately, accuracy-on-the-line is not a universal phenomenon. In some datasets, such as Camelyon17-\textsc{wilds}~\cite{bandi2018detection}, models with the same ID test performance have OOD performance that varied largely.  Thus, while the accuracy-on-the-line phenomenon is interesting, its practical use is somewhat limited since determining whether accuracy-on-the-line holds requires labeled OOD data in the first place.

\begin{figure}[t]
    \centering
    \includegraphics[scale=0.32]{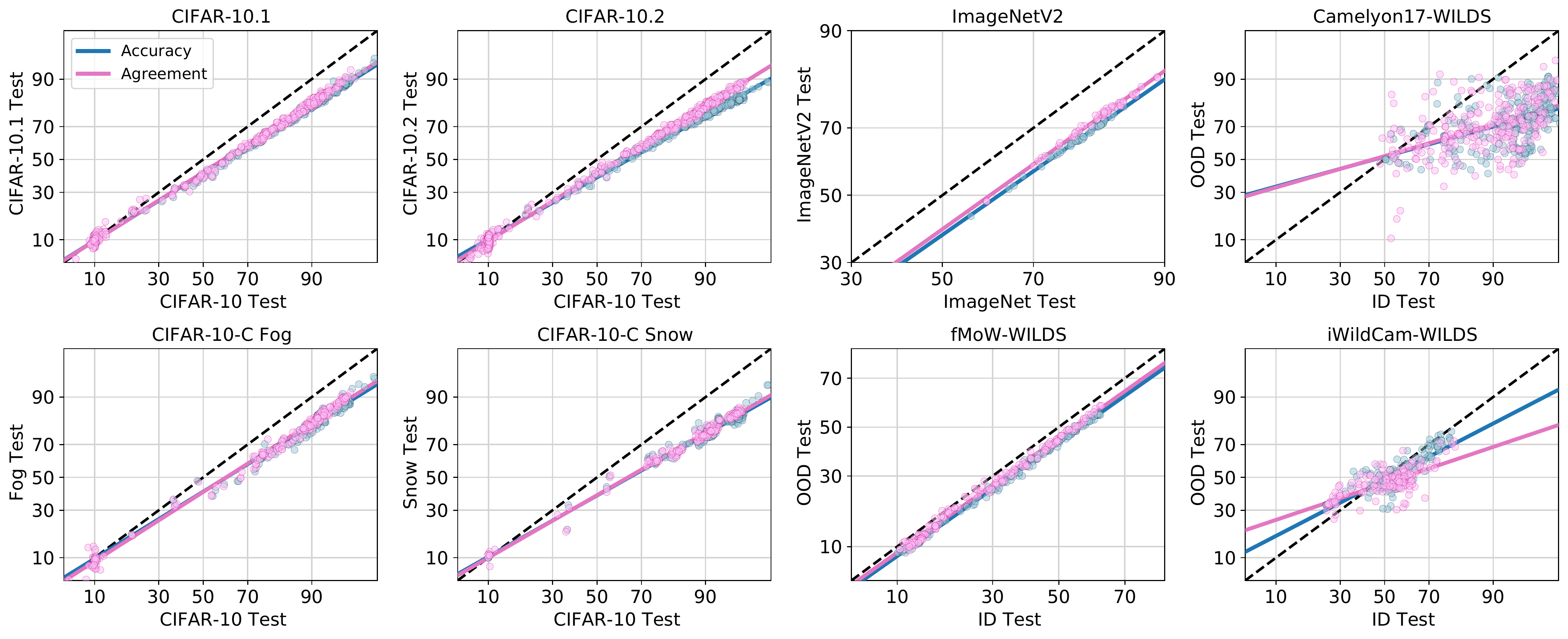}
    \caption{When ID vs. OOD accuracy is linearly correlated, the ID vs. OOD agreement is also linearly correlated. Additionally, when ID vs. OOD accuracy is not linearly correlated, agreement is also not linearly correlated. Each blue point in the scatter plot represents the accuracy of a single model. Each pink point represents the agreement between a pair of models. To avoid cluttering the figure, given $n$ models of interest, we only plot $n$ random pairs. The axes are probit scaled as described in Section \protect{\ref{subsection:experimental_setup}}.}
    \label{fig:linearly_correlated}
\end{figure}

In this work, we begin by observing an analogous phenomenon based upon \emph{agreement} rather than accuracy.  Specifically, if we consider pairs of neural network of classifiers, and look at the agreement of their predictions (the proportion of cases where they make the same prediction, which requires no labeled data to compute), we find that there \emph{also} often exists a strong linear correlation between ID vs. OOD agreement.  We call this phenomenon \textit{agreement-on-the-line}.  Importantly, however, this phenomenon appears to be tightly coupled with accuracy-on-the-line: when agreement-on-the-line holds, accuracy-on-the-line also holds; and when agreement-on-the-line \emph{does not} hold, neither does accuracy-on-the-line.  Furthermore, when these properties hold, the linear correlations of both accuracy-on-the-line and agreement-on-the-line appear to have roughly \emph{the same slope and bias}.  Interestingly, unlike accuracy-on-the-line, which appears to be a general phenomenon, agreement-on-the-line, especially the fact that the slope and bias of the linear correlation agree across accuracy and agreement, appears to occur only for neural networks.  Indeed, the phenomenon is quite unintuitive, given that there is no a-priori reason to believe that agreement and accuracy would be connected in such a manner; nonetheless, we find this phenomenon occurs repeatedly across multiple datasets and vastly different neural network architectures.

This phenomenon is of immediate practical interest.  Since agreement-on-the-line can be validated \emph{without} any labeled OOD data, we can use it as a proxy to assess whether accuracy-on-the-line holds, and thus whether ID accuracy is a reasonable OOD model selection criteria. Furthermore, since the slope and bias of the agreement-on-the-line fit can \emph{also} be estimated without labeled OOD data, (for the cases where agreement-on-the-line holds) we can use this approach to derive a simple algorithm for estimating the OOD generalization of classifiers, without \emph{any} access to labeled OOD data.  The approach outperforms competing methods and predicts OOD test error with a mean absolute estimation error of $\leq 2\%$ on datasets where agreement-on-the-line holds.  On datasets where agreement-on-the-line does not hold, the method as expected does not perform as well, but surprisingly \emph{still} outperforms competing methods in terms of predicting OOD performance.

To summarize, our contributions are as follows:
\begin{enumerate}
    \item We discover and empirically analyze the agreement-on-the-line phenomenon: that ID vs. OOD \emph{agreement} for pairs of neural network classifiers lies on a line precisely when the corresponding ID vs. OOD \emph{accuracy} also lies on a line.  Furthermore, the slope and bias of these two lines are approximately equal.
    
    \item Leveraging this phenomenon, we develop a simple method for estimating the OOD performance of classifiers without \emph{any} access to labeled OOD data (and by observing whether agreement-on-the-line holds, the method also provides a ``sanity check'' that these estimates are reasonable).  The proposed method outperforms all competing baselines for this task.
\end{enumerate}

\section{Related Works}
\paragraph{Accuracy-on-the-line} Before \citet{Miller1}, several other works proposing new benchmarks for performance evaluation ~\citep{miller2020effect, recht_cifar10.1andimagenetv2, roelofs2019meta, yadav2019cold} have also observed a strong linear correlation between ID and OOD performance of models. Theoretical analysis of this phenomenon is limited. \citet{Miller1} constructs distribution shifts where accuracy-on-the-line can be observed assuming the data is Gaussian, however they do not incorporate any assumptions about the classifier. Another work by \citet{mania2020classifier} analyzes the phenomenon under the assumption that given two models, it is unlikely that the lower accuracy model classifies a data point correctly while the higher accuracy model classifies it incorrectly. Such model similarity has been observed by~\cite{mania2019model, geirhos2021partial}. Recent works have also observed some nonlinear correlations~\cite{liang2022nonlinear} and negative correlations in performance~\cite{teney2022id, kaplun2022deconstructing}. In this work, we focus on the results from \citet{Miller1} and leave these nuances for future study.

\paragraph{Estimating ID generalization via agreement.}  
Departing from conventional approaches based on uniform convergence~\cite{neyshabur17exploring, dziugaite2017computing, bartlett2017spectrally, nagarajan2019deterministic}, several recent works~\citep{negrea2020, zhou2020, garg2021, jiang2022assessing} propose different approaches for estimating generalization error. In particular, this work is closely related to~\citet{jiang2022assessing}, which shows that the disagreement between two models trained with different random seeds closely tracks the ID generalization error of the models, if the ensemble of the models are well-calibrated. Predicting ID generalization via disagreement has also previously been proposed by \citet{madani} and \citet{nakkiran2021distributional}. Our method also uses disagreement for estimating performance but, unlike these works, we focus on OOD generalization, and, more importantly, we do not require calibration or models with the same architecture. 

\paragraph{OOD generalization.} The problem of characterizing generalization in the OOD setting is even more challenging than the ID setting. \citet{ben2006analysis} provides one of the first uniform-convergence-based bounds for \textit{domain adaptation}, a related but harder framework of improving the OOD performance of models given unlabeled OOD data and labeled ID data. Several works~\citep{mansour2009domain, cortes2010learning, kuzborskij2013stability} build on this approach and extend it to other learning scenarios. Most of these works attempt to bound the difference between ID and OOD performance via a certain notion of closeness between the original distribution and shifted distribution (e.g., the total variation distance and the $\mc{H} \Delta \mc{H}$ divergence which is related to agreement), and build on the uniform-convergence framework~\citep{redko2020survey}. As pointed out by \citet{Miller1}, these approaches provide upper bounds on the OOD performance that grows looser as the distribution shift becomes larger, and the bounds do not capture the precise trends observed in practice. Predicting the actual OOD performance using unlabeled data has gained interest in the past decade. These methods can roughly be divided into three categories:

\paragraph{1. Placing assumptions on the distribution shift.}\citet{JMLR:v11:donmez10a} assume knowledge of the marginal of the shifted label distribution $P(y)$ and show that OOD accuracy can be predicted if the shifted distribution satisfies several properties. \citet{steinhardt2016risk} work under the assumption that the data $x$ can be separated into ``views'' that are conditionally independent given label $y$. \citet{Chen21} assume prior knowledge about the shift and use an importance weighting procedure. 

\paragraph{2. Placing assumptions on the classifiers.} Given multiple classifiers of interest, \citet{platanios_logic2, platanios_logic1} form logical constraints based on assumptions about the hypothesis distribution to identify individual classifier's error. On the other hand, \citet{pmlr-v38-jaffe15} relates accuracy to the  classifiers’ covariance matrix  under the assumption that classifiers make independent errors and do better than random.

\paragraph{3. Empirically measuring the distribution shift.}
A group of works \cite{elsahar-galle-2019-annotate, schelter_typical, dengzheng21, Deng21} train a regression model over metrics that measure the severity of the distribution shift. Inspired by the observation that the maximum softmax probability (or confidence) for OOD points is typically lower \cite{DBLP:conf/iclr/HendrycksG17, hendrycks_cifar10c},  \citet{Guillory21} and more recently  \citet{garg2022leveraging} utilize model confidence to predict accuracy. \citet{Chuang0J20} uses agreement with a set of domain-invariant predictors as a proxy for the unknown, true target labels. This method was extended upon by \citet{chen2021detecting} which improves the predictors by self-training. \citet{yu2022predicting} observed that the distance between the model of interest $f$ and a reference model trained on the pseudolabels of $f$ showed strong linear correlation with OOD accuracy.

Though a large number of methods have been proposed, for the large majority, it is not well-understood when they will work. Intuitively, no method will work on all shifts without additional assumptions~\citep{garg2022leveraging}. But is there some \emph{simple general structure} to shifts in the real world that allows us to reliably predict OOD accuracy? Even if such a structure is not universal, can we easily \emph{inspect} if this structure holds? What is a plausible assumption we can make about the OOD \emph{behaviour of classifiers}? The novelty and significance of our work comes from trying to better understand and address these questions, specifically for neural networks. In this work, we observe a phenomenon related to, but stronger than accuracy-on-the-line that allows us to reliably predict the OOD accuracy of neural networks.

\begin{table}[t]
\small
    \centering
    \begin{tabular}{c ccc   ccc c}
        \hline
         \textbf{Dataset} &  \multicolumn{3}{c}{\textbf{Accuracy}} & \multicolumn{3}{c}{\textbf{Agreement}} & \textbf{Confidence Interval}\\ 
         & Slope & Bias & $R^2$  & Slope & Bias & $R^2$ & \\ 
         \cmidrule(lr){2-2}\cmidrule(lr){3-3}\cmidrule(lr){4-4} \cmidrule(lr){5-5}\cmidrule(lr){6-6}\cmidrule(lr){7-7} \cmidrule(lr){8-8}
         CIFAR-10.1v6 &	0.842 &	-0.216 & 0.999 & 0.857 & -0.205 & 	0.997 & (-0.046, 0.017)\\
        CIFAR-10.2 &	0.768 &	-0.287 & 0.999 & 0.839 & -0.226 & 0.996 & (-0.120, -0.030)\\
        ImageNetv2 & 0.946 & -0.309 & 0.997	 & 0.972 & -0.274 &	0.993 & (-0.0720, 0.061)\\ 
        CIFAR-10C-Fog & 0.834	& -0.228	& 0.995	 & 0.870 & -0.239 & 0.996 & (-0.077, 0.053)\\    
        CIFAR-10C-Snow & 0.762 & -0.289 & 0.974	 & 0.766 & -0.266 & 0.974 & (-0.067, 0.047)\\ 
        fMoW-\textsc{wilds} & 0.952	& -0.163	& 0.998	& 0.954 & -0.121 & 0.995 & (-0.042, 0.030)\\ 
        \midrule
        Camelyon17-\textsc{wilds} & 0.373 & 0.046 & 0.263 & 0.381 & 0.075 & 0.226 & -  \\
        iWildCam-\textsc{wilds} & 0.700 & -0.037 & 0.738	& 0.411 & -0.094 & 0.424& -  \\
        \bottomrule
    \end{tabular}
    \vspace{3mm}
    \caption{Slope, bias, and coefficients of determination (R$^2$) values of linear correlations between ID vs. OOD accuracy and ID vs. OOD agreement. The slope/bias of these linear correlations match when the $R^2$ value is high (i.e. strong linear correlation). We also look at the 95$\%$ confidence interval of the difference in slopes between the agreement line and accuracy line for the datasets where we observe strong correlation. For all datasets excluding CIFAR-10.2, we observe that the slope difference is not statistically significant. See Section \protect{~\ref{subsec:observations}} for more details.}
    \label{tab:linearly_correlated}
    \vspace{-1mm}
\end{table}

\section{The agreement-on-the-line phenomenon}
\label{sec:phenomena}
\subsection{Notation and setup}
\label{sec:notation}

Let $\mc{H}$ denote a set of neural networks trained on $(X_{\msf{train}}, \mb{y}_{\msf{train}}) = \{(x_i,y_i)\}_{i=1}^{m_{\msf{train}}}$ sampled from $\mc{D}_{\msf{ID}}$. 

Given any pair of models $h, h' \in \mc{H}$, for a distribution $\mc{D}$, the expected accuracy and agreement are defined as:
{\small
\begin{align}
    \msf{Acc}(h) &= \E_{x,y \sim \mc{D}}[\mathbbm{1}\{h(x) = y\}], \quad
    \msf{Agr}(h, h') = \E_{x \sim \mc{D}}[\mathbbm{1}\{h(x) = h'(x)\}].
\end{align}}
We assume access to a labeled validation set $(X_{\msf{val}}, \mb{y}_{\msf{val}}) = \{(x_i,y_i)\}_{i=1}^{m_{\msf{Val}}}$ sampled from $\mc{D}_{\msf{ID}}$ that allows us to estimate the ID accuracy $\widehat{\msf{Acc}}_{\msf{ID}}(h)$ as the sample average of $\mathbbm{1}\{h(x) = y\}$ over the validation set. We do not assume access to a labeled OOD validation set, as this is often impractical to obtain, and thereby cannot directly estimate $\widehat{\msf{Acc}}_{\msf{OOD}}(h)$ in a similar manner. 

Agreement, on the other hand, only requires access to unlabeled data. We assume access to \emph{unlabeled} samples $X_{\msf{OOD}} = \{x_i\}_{i=1}^{m_{\msf{OOD}}}$ from the shifted distribution of interest $\mc{D}_{\msf{OOD}}$. Hence, we can estimate both the ID and OOD agreement as follows:
{\small
\begin{align}
\label{eq:sampleaverage}
\widehat{\msf{Agr}}_{\msf{ID}}(h, h') = \frac{1}{m_{\msf{val}}} \sum_{x \in X_{\msf{val}}} \mathbbm{1}\{h(x) = h'(x)\}, ~~\widehat{\msf{Agr}}_{\msf{OOD}}(h, h') = \frac{1}{m_{\msf{OOD}}} \sum_{x \in X_{\msf{OOD}}} \mathbbm{1}\{h(x) = h'(x)\}
\end{align}}

\subsection{Experimental setup}
\label{subsection:experimental_setup}
We study the ID vs. OOD accuracy and agreement between pairs of models across more than 20 common OOD benchmarks and hundreds of independently trained neural networks. 

\paragraph{Datasets.}
We present results on 8 dataset shifts in the main paper, and include results for other distribution shifts in the Appendix~\ref{app:more_dataset}. These 8 datasets span:
\begin{enumerate}
    \item Dataset reproductions: CIFAR-10.1 \cite{recht_cifar10.1andimagenetv2}, CIFAR-10.2 \cite{lu_cifar10.2} reproductions of CIFAR-10 \cite{cifar10} and ImageNetV2 \cite{recht_cifar10.1andimagenetv2} reproduction of ImageNet \cite{deng2009imagenet} 
    \item Synthetic corruptions: CIFAR-10C Fog and CIFAR-10C Snow \cite{hendrycks_cifar10c}
    \item Real-world shifts from~\citep{koh2021wilds}: satellite images (fMoW-\textsc{wilds}), images from camera traps in the wildlife (iWildCam-\textsc{wilds} \cite{beery2020iwildcam}), and images of cancer tissue (Camelyon17-\textsc{wilds} \cite{bandi2018detection})
\end{enumerate} 
Appendix~\ref{app:more_dataset} includes results on CINIC-10 \cite{cinic10}, STL-10 \cite{stl10}, and other \textsc{wilds} and CIFAR-10C benchmarks. We also investigate an analogous phenomenon for the F1-score used to assess the reading comprehension performance of language models on the Amazon-SQuAD benchmark \cite{miller2020effect}. 

\paragraph{Models.}
For ImageNetV2, we evaluate 50 ImageNet pretrained models from the timm \cite{rw2019timm} package. On the remaining 7 shifts, we evaluate on all independently trained models in the testbed created and utilized by~\citep{Miller1} consisting of $\geq 150$ models for each shift. The evaluated models span a variety of convolutional neural networks (e.g. ResNet \cite{resnet}, DenseNet \cite{densenet}, EfficientNet \cite{tan2019efficientnet}, VGG \cite{liu2015vgg}) as well as various Vision Transformers \cite{dosovitskiy2020vit}. All architectures and models are listed in the Appendix~\ref{app:model}. 

\paragraph{Probit scaling.}
\citet{Miller1} report their results after probit scaling $\Phi^{-1}(\cdot)$ \footnote{The probit transform is the inverse of the cumulative density function of the standard Gaussian distribution.} the ID vs. OOD accuracies due to a better linear fit. We apply the same probit transform to both accuracy and agreement in our experiments.

\subsection{Observations} \label{subsec:observations}We empirically observe a peculiar phenomenon in deep neural networks, which we call \emph{agreement-on-the-line} characterized by the following three properties:
 \begin{description}
     \item[Prop(i)]When ID vs. OOD accuracy observes a strong linear correlation ($\geq 0.95$ $R^2$ values), we see that ID vs. OOD \emph{agreement is also strongly linearly correlated}. 
     \item[Prop(ii)]When both accuracy and agreement observe strong linear correlation, we see that these linear correlations have almost the \emph{same slope and bias}. 
     \item[Prop(iii)]When the linear correlation of ID vs. OOD accuracy is weak ($\leq 0.75\,\,R^2$ values), the linear correlation between ID and OOD agreement is similarly weak. \footnote{The $R^2$ thresholds were only chosen to discretize the strength of the linear correlations as strong or weak for the 8 shifts. As shown in Appendix~\ref{app:more_dataset}, the phenomenon actually follows a gradient i.e. when the $R^2$ value is higher, the slope/bias of ID vs. OOD accuracy and ID vs. OOD agreement become closer to each other.}
 \end{description} 
 We show the agreement-on-the-line phenomenon on 8 datasets in Figure \ref{fig:linearly_correlated} and Table \ref{tab:linearly_correlated}. On CIFAR-10.1, CIFAR-10.2, ImageNetV2, CIFAR-10C Fog/Snow, and fMoW-\textsc{wilds}, we find that both ID vs. OOD accuracy and agreement observe strong linear correlations, and the linear fits have the same slope and bias (Prop(i), Prop(ii)). On the other hand, on the datasets Camelyon17-\textsc{wilds} and iWildCam-\textsc{wilds} where accuracy is not linearly correlated, agreement is also not linearly correlated (Prop(iii)).

To ensure that the differences in the slopes is not statistically significant, we construct the following hypothesis test. For each dataset, we randomly sample 1000 subsets of 10 models from the model testbed and compute the corresponding difference in the slope of the linear fits: $\mathsf{Slope_{Acc}} - \mathsf{Slope_{Agr}}$. In Table~\ref{tab:linearly_correlated}, we look at the 95$\%$ confidence interval of the distribution of these slope differences for the 6 datasets where we observe a strong correlation. For all datasets with the exception of CIFAR10.2, the null hypothesis (i.e. difference in slope is 0) lies within the 95$\%$ confidence interval thus cannot be rejected.

\begin{figure}
  \begin{minipage}[c]{0.65 \textwidth}
    \includegraphics[width=\textwidth]{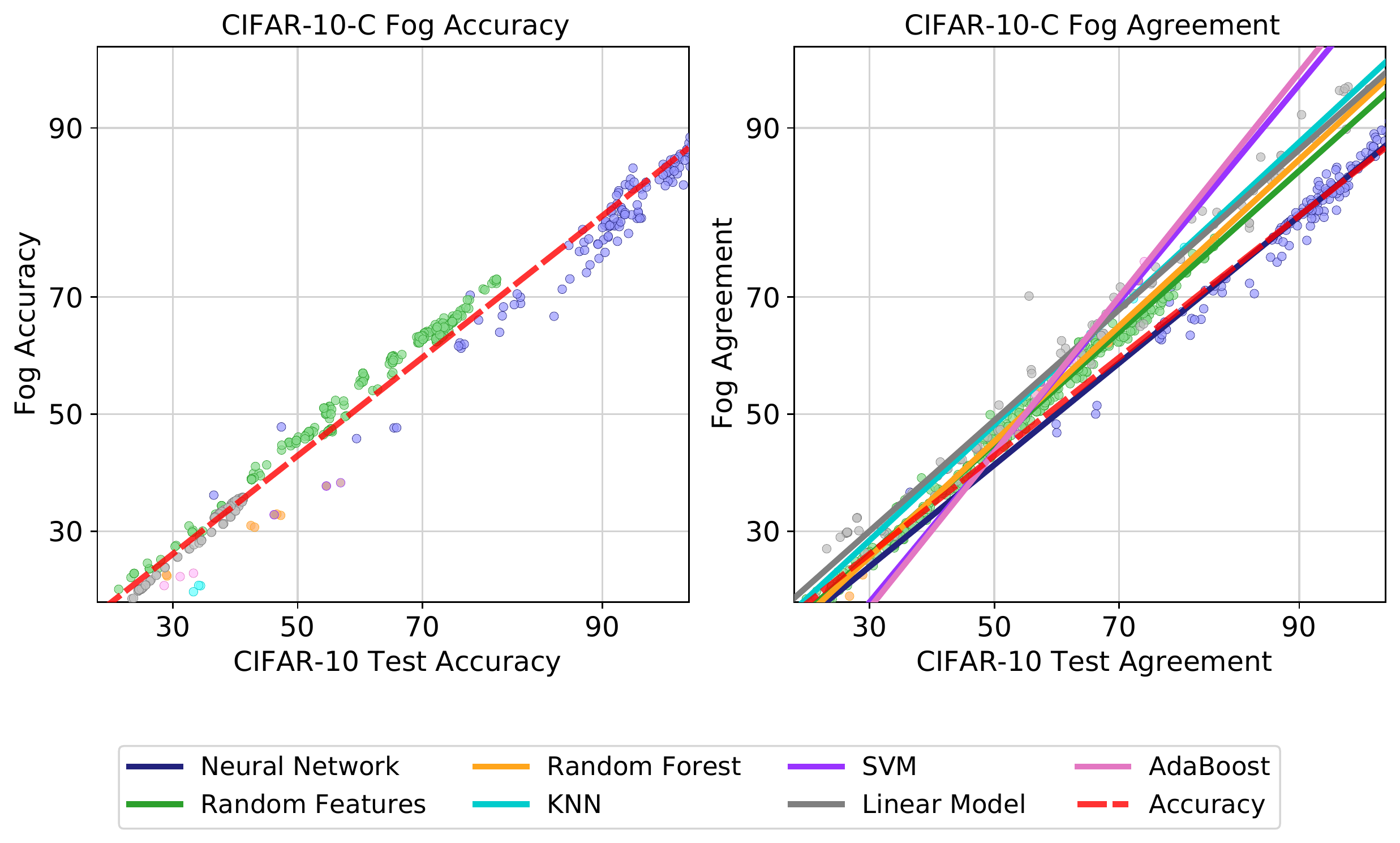}
  \end{minipage}\hfill
  \begin{minipage}[c]{0.3\textwidth}
    \caption{ We observe whether the agreement-on-the-line phenomenon happens across various model classes on the CIFAR-10 Fog dataset. As shown on the left, the ID vs. OOD accuracy of all model classes lie on the same line. We plot ID vs. OOD agreement between pairs of models from the same model class and observe that only the linear correlation between ID vs. OOD agreement of neural networks match that of ID vs. OOD accuracy (in red).} 
    \label{fig:nn_only}
  \end{minipage}
\end{figure}

\subsection{What makes agreement-on-the-line interesting?}
First, \emph{agreement can be estimated with just unlabeled data}. Hence, the agreement-on-the-line phenomenon has important practical implications for both \textit{checking} whether the distribution shift observes accuracy-on-the-line and \textit{predicting} the actual value of OOD accuracy without any OOD labels. We present a method to estimate OOD error using this phenomenon in Section~\ref{sec:method}.

Second, agreement-on-the-line does \emph{not} directly follow from accuracy-on-the-line. Prior work has observed that expected ID accuracy often equals ID agreement over pairs of models with the same architecture, trained on the same dataset but with different random seeds~\citep{jiang2022assessing}. Agreement-on-the-line goes beyond these results in two ways: (i) agreement between models with \emph{different architectures} (Fig. \ref{fig:nn_only}) and (ii) agreement between different checkpoints on the \emph{same training run} (Fig. \ref{fig:traj}) is also on the ID vs. OOD agreement line. These ID/OOD agreements \emph{do not equal} the expected ID/OOD accuracy. Indeed, understanding why agreement-on-the-line holds requires going beyond the theoretical conditions presented in the prior work~\citep{jiang2022assessing} which do not hold for this expanded set of models. Furthermore, the phenomenon of matching slopes is \emph{not} due to the models performing well ID and OOD. Trivially, the agreement between a model and a perfect classifier is the accuracy of the model. However, in our experiments, even pairs of bad models display this phenomenon. See Appendix~\ref{app:calibration} for further discussion.

Finally, we emphasize that there is something special about neural networks that makes the ID vs. OOD agreement trend identical to the ID vs. OOD accuracy trend. This is unlike accuracy-on-the-line that holds across a wide range of models including neural networks and classical approaches. Figure \ref{fig:nn_only} shows CIFAR-10 Test vs. CIFAR-10C Fog accuracy and agreement of linear models (e.g. logistic and ridge regression) and various non-linear models (e.g. Kernel SVM \cite{cortes1995svm}, k-Nearest Neighbors, Random Forests \cite{randomforest}, Random Features \cite{randomfeatures}, AdaBoost \cite{adaboost}). See plots for other datasets in Appendix~\ref{app:more_dataset}. We look at agreement between pairs of models from the same model family. While Prop(i) seems to hold for several other model families on several shifts, Prop(ii) only holds for neural networks, i.e. the slope and bias of the agreement line \textit{do not match} the slope and bias of the accuracy line for other model families.

\section{A method for estimating OOD accuracy}
\label{sec:method}
In this section, we describe how the phenomenon of agreement-on-the-line (described in Section~\ref{sec:phenomena}) offers a simple practical method to perform model selection and estimate accuracy under distribution shifts. Recall from Section~\ref{sec:notation} that we have labeled ID validation data $(X_{\msf{val}}, \mb{y}_{\msf{val}})$  and unlabeled OOD data $X_{\msf{OOD}}$. 

\paragraph{Model selection.} Without OOD labeled data, can we determine which model is likely to achieve the best OOD performance? When accuracy-on-the-line holds and ID vs. OOD accuracy is linearly correlated, we can simply pick the model with highest ID accuracy. In practice, how does one determine if accuracy-on-the-line holds without labeled OOD data? By Prop(i) and Prop(iii), agreement-on-the-line implies accuracy-on-the-line. Hence, we simply need to check if ID and OOD agreement (which can be estimated as in~\eqref{eq:sampleaverage}) are linearly correlated, in order to know if our model selection criterion based on ID accuracy is valid. 

\paragraph{OOD error prediction.} Agreement-on-the-line allows us to go beyond model selection and actually \emph{predict OOD accuracy}. Intuitively, we can estimate the slope and bias of the agreement line with just unlabeled data. By Prop(ii), they match the slope and bias of the accuracy line and hence, we can estimate the OOD accuracy by linearly transforming the ID accuracy (with the appropriate probit scaling). We formalize this intuition below and provide an algorithm for OOD accuracy estimation in Algorithm~\ref{alg:ALine}. Implementation of our method is available at \url{https://github.com/kebaek/agreement-on-the-line}.

Recall (Section~\ref{sec:notation}) that given $n$ distinct models of interest $\mc{H} = \{h_i\}_{i=1}^n$, we can estimate $\msf{Acc}_{\msf{ID}}(h)$, $\msf{Agr}_{\msf{ID}}(h, h')$ and $\msf{Agr}_{\msf{OOD}}(h, h')$ as sample averages over ID labeled validation data and OOD unlabeled data for all $h, h' \in \mc{H}$. We now describe an estimator $\widehat{\msf{Acc}}_{\msf{OOD}}(h)$ for the OOD accuracy of a model $h \in \mc{H}$. 

From agreement-on-the-line, we know that when ID vs. OOD agreement lies on a line for all $h, h' \in \mathcal{H} \times \mathcal{H}$, ID vs. OOD accuracy for all $h \in \mathcal{H}$ would approximately also lie on the same line:
\begin{align}
\label{eq:acc_corr}
    \Phi^{-1}(\msf{{Acc}_{OOD}}(h)) = a \cdot \Phi^{-1}(\msf{{Acc}_{ID}}(h)) + b 
    \Leftrightarrow \Phi^{-1}(\msf{{Agr}_{OOD}}(h, h'))= a \cdot \Phi^{-1}(\msf{{Agr_{ID}}}(h, h')) + b
\end{align}
We estimate the slope and bias of the linear fit by performing linear regression after applying a probit transform on the agreements as follows.
\begin{align}
\label{eq:linreg}
    \hat{a}, \hat{b} = \arg \min_{a,b \in \R} \sum_{i \neq j} (\Phi^{-1}(\msf{\widehat{Agr}_{OOD}}(h_i, h_j)) - a \cdot \Phi^{-1}(\msf{\widehat{Agr}_{ID}}(h_i, h_j)) - b)^2
\end{align}
For each model $h \in \mathcal{H}$, given its ID validation accuracy, one could simply plug the estimated slope $\hat{a}$ and bias $\hat{b}$ from~\eqref{eq:linreg}, and $\widehat{\msf{Acc}}_{\msf{ID}}(h)$ (sample average over validation set) into~\eqref{eq:acc_corr} to get an estimate of the model's OOD accuracy. We call this \textit{simple} algorithm ALine-S.

Notice that ALine-S does not directly use the OOD agreement estimates concerning the model of interest---we only use agreements indirectly via the estimates $\hat{a}$ and $\hat{b}$. We find that a better estimator can be obtained by \textit{directly} using the model's OOD agreement estimates via simple algebra as follows. 

First, note that for any pair of models $h, h' \in \mathcal{H}$, it directly follows from~\eqref{eq:acc_corr} that
\begin{align}
\label{eq:pair}
    \frac{\Phi^{-1}(\msf{{Acc}_{OOD}}(h)) + \Phi^{-1}(\msf{{Acc}_{OOD}}(h'))}{2} = a \cdot \frac{\Phi^{-1}(\msf{{Acc}_{ID}}(h)) + \Phi^{-1}(\msf{{Acc}_{ID}}(h'))}{2} + b 
\end{align}
By substituting $b = \Phi^{-1}(\msf{{Agr}_{OOD}}(h, h')) - a \cdot \Phi^{-1}(\msf{{Agr}_{ID}}(h, h'))$ into~\eqref{eq:pair}, we can get that average OOD accuracy of any pair of models $h, h' \in \mathcal{H}$ is  
\begin{equation}
\begin{split}
\label{eq:prediction}
    &\frac{1}{2} \underbrace{\Phi^{-1}(\msf{{Acc}_{OOD}}(h))}_{\text{unknown}}+\frac{1}{2} \underbrace{\Phi^{-1}(\msf{{Acc}_{OOD}}(h'))}_{\text{unknown}} \\
    =&\,\, \underbrace{\Phi^{-1}(\msf{{Agr}_{OOD}}(h, h')) + a \cdot \left(\frac{\Phi^{-1}(\msf{{Acc}_{ID}}(h)) + \Phi^{-1}(\msf{{Acc}_{ID}}(h'))}{2} - \Phi^{-1}(\msf{{Agr}_{ID}}(h, h'))\right)}_{\text{known (can estimate via sample average over $X_\msf{OOD}$ and $(X_{\msf{val}}, y_{\msf{val}})$)}}.
\end{split}
\end{equation}
We can plug in estimates of the terms on the right hand side ($\hat{a}$ from linear regression~\eqref{eq:linreg}) and the rest from sample averages. 
In this way, we can construct a system of linear equations of the form~\eqref{eq:prediction} involving ``unknown'' estimates of the probit transformed OOD accuracy of models and other ``known'' quantities. We solve the system via linear regression to obtain the unknown estimates. We call this procedure ALine-D, and it is described more explicitly in Algorithm \ref{alg:ALine}. Note that there must be at least 3 models in the set of interest $\mc{H}$ for the system of linear equations in \eqref{eq:prediction} to have a unique solution. 

\begin{algorithm}
\caption{ALine-D: Predicting OOD Accuracy}\label{alg:ALine}
\begin{algorithmic}[1]
\State \textbf{Input:} $m_{\mathrm{ID}}$ validation samples $(X_{\mathrm{ID-val}},   \textbf{y}_{\mathrm{ID-val}})$, $m_{\mathrm{OOD}}$ unlabeled samples $X_{\mathrm{OOD}}$, a set containing $n$ models of interest $\mathcal{H}$
\State Get $\mathsf{\widehat{Acc}_{ID}}(h_i) \ \forall i \in [n]$
\State Get $\mathsf{\widehat{Agr}_{ID}}(h_i, h_j)$ and $\mathsf{\widehat{Agr}_{OOD}}(h_i, h_j)$ for all pairs of models $i \neq j$
\State Get $\hat{a}, \hat{b} = \arg \min_{a,b \in \R} \sum_{i\neq j} (\Phi^{-1}(\msf{\widehat{Agr}_{OOD}}(h_i, h_j)) - a \cdot \Phi^{-1}(\msf{\widehat{Agr}_{ID}}(h_i, h_j)) - b)^2$
\State Initialize $A \in \R^{\frac{n(n-1)}{2} \times n}$, $\mb{b} \in \R^{\frac{n(n-1)}{2}}$
\State $ i = 0$
\For{ $h_j, h_{k} \in \mathcal{H}$}
    \State $A_{ij} = \frac{1}{2}, A_{ik} = \frac{1}{2}$, $A_{i\ell} = 0 \ \forall l \notin \{j, k\}$
    \State $\mb b_i = \Phi^{-1}(\msf{\widehat{Agr}_{OOD}}(h_j, h_k)) + \hat{a} \cdot \left(\frac{\Phi^{-1}(\msf{\widehat{Acc}_{ID}}(h_j) + \Phi^{-1}(\msf{\widehat{Acc}_{ID}}(h_k)))}{2} - \Phi^{-1}(\msf{\widehat{Agr_{ID}}}(h_j, h_k))\right)$
    \State $i = i + 1$
\EndFor

\State Get $\mb{w}^* = \arg \min_{\mb w \in \R^n} \norm{A \mb{w} - \mb{b}}{2}^2$
\State \Return $\Phi(w_i^*) \,\,\, \forall i \in [n]$
\end{algorithmic}
\end{algorithm}

\section{Experiments}
\label{sec:experiments}

\paragraph{Datasets and models.} We evaluate our methods, the simple plug in of slope/bias estimate ALine-S and the more involved ALine-D, on the same models and datasets from Section \ref{sec:phenomena} and two additional datasets CIFAR-10C-Saturate and RxRx1-\textsc{wilds} (See Appendix \ref{app:more_dataset} and \ref{app:model} for details on these datasets). Specifically, we look at CIFAR-10.1, CIFAR-10.2, ImageNetV2, CIFAR-10C, fMoW-\textsc{wilds}, and RxRx1-\textsc{wilds}, where we observe a strong correlation.  We also look at the performance of models on datasets where we do not see a strong linear correlation, specifically Camelyon-\textsc{wilds} and iWildCam-\textsc{wilds}. 

\paragraph{Baseline methods.} We choose 4 existing unlabeled estimation methods for comparison: Average Threshold Confidence (ATC) by \citet{garg2022leveraging},  DOC-Feat in \citet{Guillory21}, Average Confidence (AC) in \cite{DBLP:conf/iclr/HendrycksG17}, and naive Agreement \cite{madani, nakkiran2021distributional, jiang2022assessing}. All of these methods, like ALine, are based on the softmax outputs of the model. See Appendix~\ref{app:baselines} for more details about previous methods.

We implement the version of ATC that performed best in the paper, i.e. with negative entropy as the score function and temperature scaling to calibrate the models in-distribution. Although DOC was deemed the best method in \citet{Guillory21}, we use DOC-Feat since DOC requires information from multiple OOD datasets. For ATC, DOC, and AC, consistent with the experimental design in \citet{garg2021}, we report the best number achieved between before versus after temperature scaling. We also compare with the more recent, ProjNorm by~\citet{yu2022predicting} which showed stronger linear correlation with OOD accuracy than Rotation \cite{Deng21} and ATC \cite{garg2022leveraging}. We compare with this method separately in Section~\ref{sec:other_method}, as they do not provide a way to directly estimate the OOD accuracy.

\begin{table}[b]
    \centering
    \small
    \begin{tabular}{c cccccc}
    \hline
        \textbf{Dataset} & \textbf{ALine-D}$^*$ & \textbf{ALine-S}$^*$ & \textbf{ATC} & \textbf{AC} & \textbf{DOC} & \textbf{Agreement}\\ \cmidrule(lr){1-1}
         \cmidrule(lr){2-2}\cmidrule(lr){3-3}\cmidrule(lr){4-4} \cmidrule(lr){5-5}\cmidrule(lr){6-6}\cmidrule(lr){7-7} 
          CIFAR-10.1 & \textbf{1.11} & 1.17 & 1.21 & 4.51 & 3.87 & 5.98 \\
          CIFAR-10.2 & \textbf{3.93} & \textbf{3.93} & 4.35 & 8.23 & 7.64 & 5.42\\
          ImageNetV2 & 2.06 & 2.08  & \textbf{1.14} & 66.2 & 11.50 & 6.70\\
         CIFAR-10C-Fog & \textbf{1.45} & 1.75 & 1.78 & 4.47 & 3.93 & 3.47\\
        CIFAR-10C-Snow & 1.32 & 1.97 & \textbf{1.31} & 5.94 & 5.49 & 2.57\\
        CIFAR-10C-Saturate & \textbf{0.41} & 0.77 &  0.69 & 2.03 & 1.51 & 4.14\\
        fMoW-\textsc{wilds} & \textbf{1.30} & 1.44 & 1.53 & 2.89 & 2.60 & 8.99\\ 
        RxRx1-\textsc{wilds} & \textbf{0.27} & 0.52 & 2.97 & 2.46 & 0.75 & 8.69 \\
        \midrule Camelyon17-\textsc{wilds} & \textbf{5.47} & 8.31 & 11.93 & 13.30 & 13.57 &  6.79\\ 
        iWildCam-\textsc{wilds} & 4.95 & 6.01 & 12.12 & \textbf{4.46}  & 5.02 & 7.35 \\
        \bottomrule
    \end{tabular}
    \vspace{3mm}
    \caption{Mean Absolute Error (MAE) of the OOD accuracy predictions with $\%$ as units.  ALine-D outperforms other methods on both shifts where we do and do not see accuracy-on-the-line. $^*$ denotes our methods.}
    \label{tab:mae}
\end{table}

\subsection{Main results: comparison to other methods.} 
\label{sec:other_method}
\begin{figure}[t]
    \centering
    \includegraphics[scale=0.4]{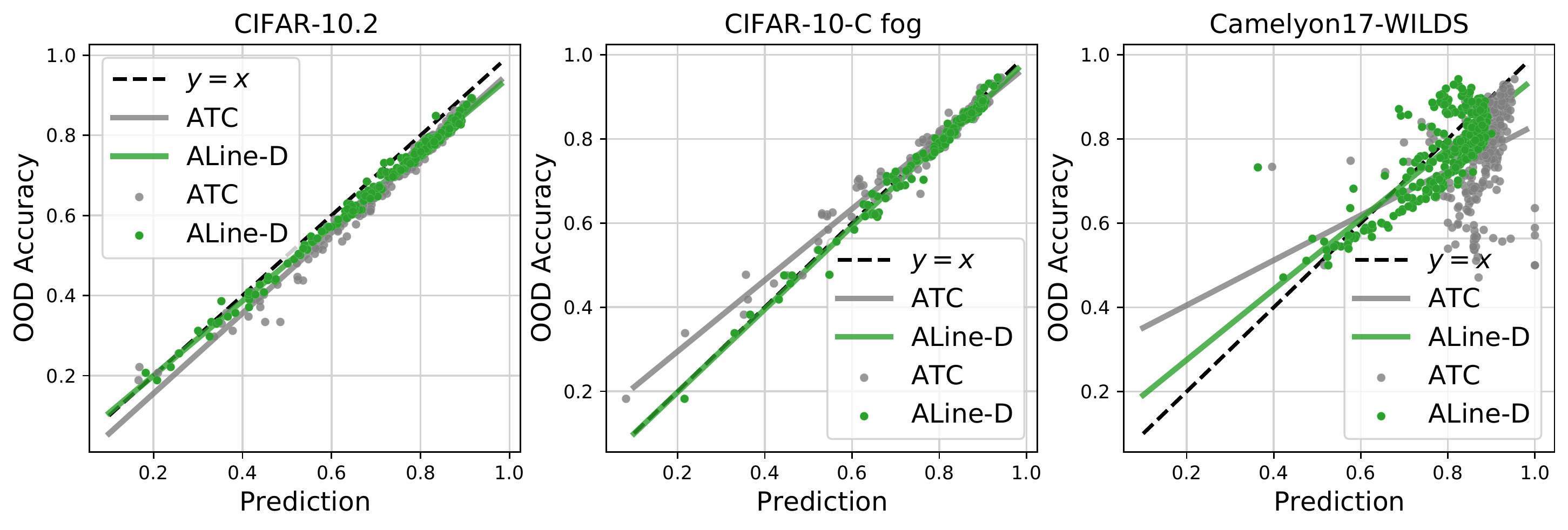}
    \caption{Prediction vs. OOD accuracy. We observe the scatter plot of prediction vs. OOD accuracy of ALine-D and ATC, the second best performing method from Table \protect{\ref{tab:mae}}. We observe that our linear fit is closer to the diagonal, as ATC underperforms on models that have low OOD accuracy. }
    \label{fig:comparison}
    \vspace{-2mm}
\end{figure}

In Table \ref{tab:mae}, we observe that ALine-D generally outperforms other methods on datasets where agreement-on-the-line holds. On ImageNet to ImageNetV2 and CIFAR-10 to CIFAR-10C-Snow, ATC performs marginally better. As can be seen in Figure \ref{fig:comparison}, ATC generally cannot accurately predict the model's OOD performance for models that do not perform very well. This is consistent with experimental results in \cite{garg2022leveraging} and \cite{yu2022predicting}.  On the other hand, ALine performs equally well on ``bad'' models and ``good'' models. In some sense, given a collection of models where we are interested in the performance of each, ATC, AC, DOC-Feat, and Agreement only utilize information from the model of interest, whereas ALine utilizes the collective information from all models for each individual prediction. 

As expected, on datasets where we do not observe a linear correlation between ID and OOD agreement (and accuracy), ALine does not perform very well, with a mean absolute estimation error of around $5\%$. Interestingly, the other methods also do not perform very well on these datasets. No method successfully predicts the OOD accuracy for every distribution shift. The advantage of ALine is that there is a concrete way to verify when the method will successfully predict the OOD accuracy (i.e. check whether agreement is on the line). Other prediction methods do not have any way of characterizing when they will be successful. Finally, we note that ALine-D actually surpasses previous methods even when accuracy-on-the-line does not hold, suggesting that the algorithm has additional beneficial properties that require further study.

\subsection{Correlation analysis}
\begin{wraptable}{r}{0.5\textwidth}
\small
    \centering
    \begin{tabular}{c cc   cc}
        \hline
         \textbf{Dataset} &  \multicolumn{2}{c}{\textbf{ALine-D}} & \multicolumn{2}{c}{\textbf{ProjNorm}}\\ 
         & $\rho$ & $R^2$  & $\rho$ & $R^2$ \\ 
         \cmidrule(lr){2-2}\cmidrule(lr){3-3}\cmidrule(lr){4-4} \cmidrule(lr){5-5} 
         CIFAR-10C & \textbf{0.995} & \textbf{0.974} & 0.98 & 0.973 \\
        \bottomrule
    \end{tabular}
    \caption{Correlation analysis. We compare the coefficients of determination (R$^2$) and rank correlations ($\rho$) between ALine-D and ProjNorm, a metric shown to have stronger correlation than ATC and Rotation.}
    \label{tab:projnorm}
\end{wraptable}

Rather than predicting OOD accuracy, it could be useful to have a metric that just strongly correlates with the OOD accuracy, if the application simply requires an understanding of relative performance such as model selection. Recently, \citet{yu2022predicting} proposed ProjNorm, a measurement they show has a very strong linear correlation with OOD accuracy, moreso than other recent methods including  Rotation \cite{Deng21} and ATC \cite{garg2022leveraging}. To compare Aline-D with ProjNorm, we replicate the CIFAR-10C study in \citet{yu2022predicting}, where they train a ResNet18 model and predict its performance across all corruptions and severity levels of CIFAR-10C (See their Table 1 in \cite{yu2022predicting}). Since ALine-D is an algorithm that requires a set of models for prediction, we use the 29 pretrained models from the CIFAR-10 testbed of \citet{Miller1}, as the other models in the set. We look at the linear correlation of the estimates of OOD accuracy and the true accuracy across all corruptions and find that ALine-D achieves \emph{stronger} correlation than ProjNorm (Table \ref{tab:projnorm}). See Appendix~\ref{app:experiment} for more experimental details.

\subsection{Estimating performance along a training trajectory}
\begin{figure}[t]
  \begin{center}
  \includegraphics[width=0.45\textwidth]{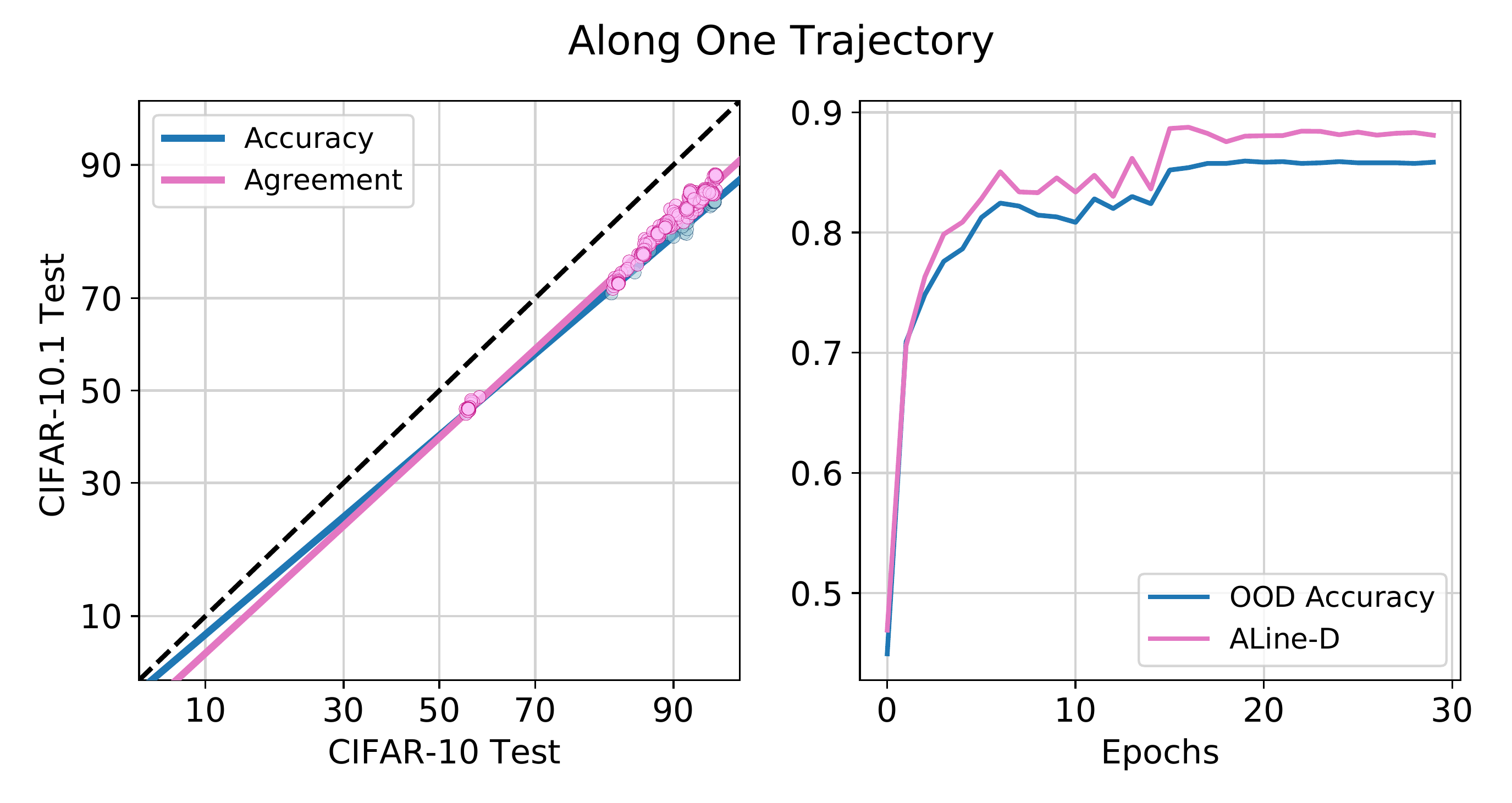}
    \includegraphics[width=0.45\textwidth]{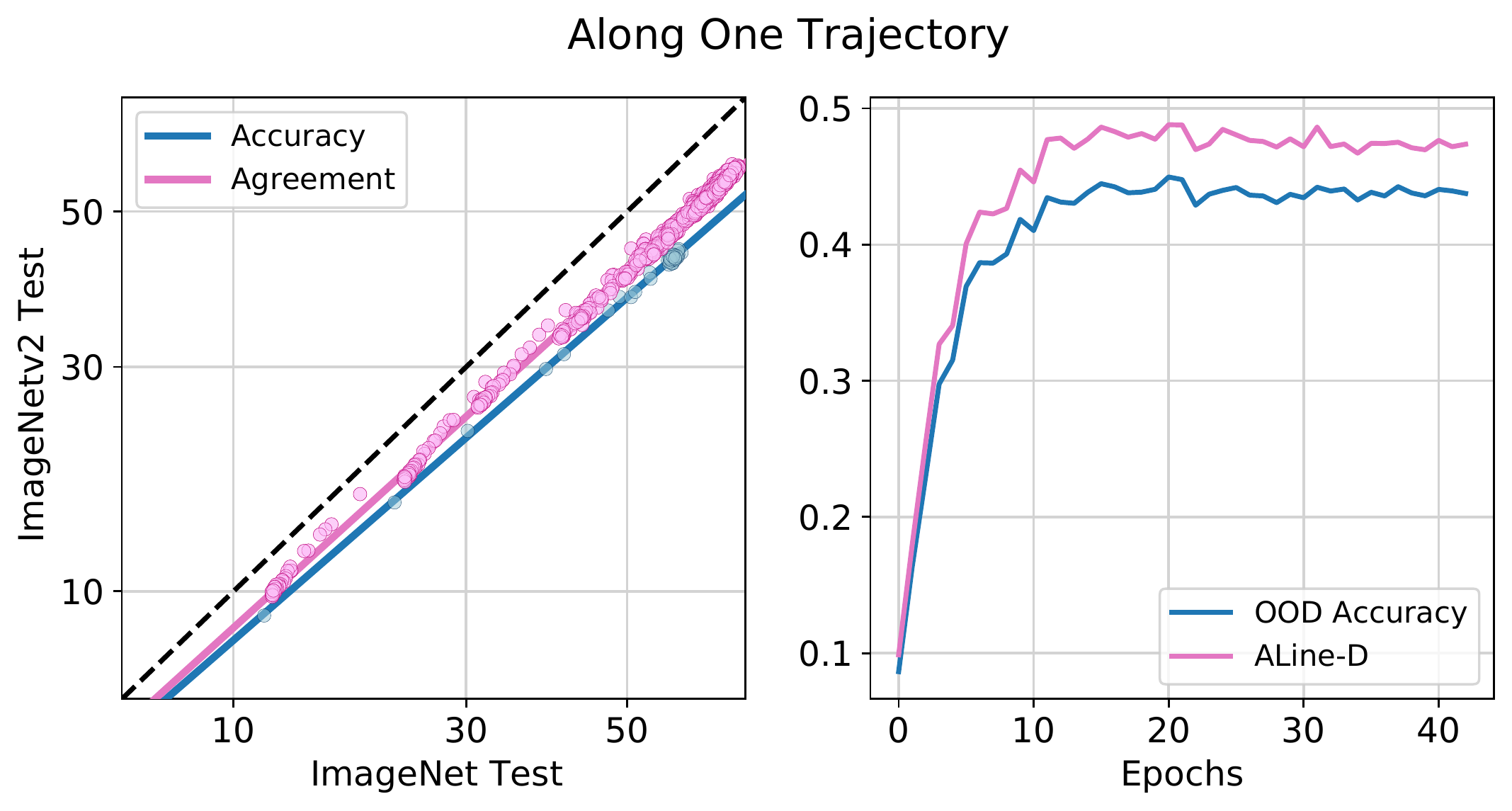}
  \end{center}
  \caption{ALine-D tracks OOD accuracy across training epochs with a MAE of $2.19\%$ for CIFAR10.1 and $3.61\%$ for ImageNetv2. }
  \label{fig:traj}
  \vspace{-1mm}
\end{figure}

We assess whether ALine can be utilized even in situations where the practioner only cares about the performance of a few models. In such situations, one could efficiently gather many models by training a single model and saving checkpoints along the way. We analyze whether our phenomenon is helpful for predicting such highly correlated hypotheses, instead of independently trained models. In Figure \ref{fig:traj}, we collect the ID and OOD test predictions every 5 epochs across CIFAR-10 training of a single ResNet18 model and every epoch across ImageNet training of a ResNet50 model. We see that even the agreement between every pair of checkpoints of a model across training is enough to get a good linear fit that matches the slope and bias of ID vs. OOD accuracy. By applying ALine-D to these checkpoints, we get a very good estimate of the OOD performance of the model across training epochs though slightly worse than for a collection of independently trained models. This suggests that given a model of interest, ALine does not require practitioners to train a large number of models, but just train one and save its predictions across training iterations. We do a more careful ablation study in Appendix~\ref{app:ablation}, looking at the number of models required for close estimates of accuracy.

\section{Conclusion}
The contributions of this work are two-fold. First, we observe the agreement-on-the-line phenomenon, and show that it correlates strongly with accuracy-on-the-line over a range of datasets and models.  We also highlight that certain aspects of this phenomenon, namely the fact that the slope and bias of the linear fit is largely the \emph{same} across agreement and accuracy, are specific to neural networks, and thus fundamentally seem connected to these classes of models.  Second, using this empirical phenomenon, we propose a surprisingly simple but effective method for predicting OOD accuracy of classifiers, while only having access to unlabeled data from the new domain (and one that can be ``sanity checked'' via testing whether agreement-on-the line holds).  Our method outperforms existing state-of-the-art approaches to this problem.
Importantly, we do \emph{not} claim that this phenomenon is universal, but we found it to be true across an extensive range of neural networks and OOD benchmarks that we experimented on. In addition to its practical relevance, this observation itself reveals something very interesting about the way neural networks learn, which we leave for future study. 

\begin{ack}
We thank Rohan Taori and Saurabh Garg for valuable discussions regarding the model testbeds and temperature scaling used in this work, respectively. Christina Baek was supported by a Presidential Fellowship sponsored by Carnegie Mellon University. Yiding Jiang was supported by funding from the Bosch Center for Artificial Intelligence. Aditi Raghunathan was supported by an Open Philanthropy AI Fellowship.
\end{ack}

\bibliographystyle{plainnat}
\bibliography{reference.bib}

\appendix
\section{Baseline Methods}
\label{app:baselines}
In Section~\ref{sec:experiments}, we compare ALine-S/D to ATC, DOC-Feat, AC, and Agreement. We detail these methods below. Given model $h$ (the output after softmax) and unlabeled OOD data, the algorithms are

\paragraph{Average Threshold Confidence (ATC)} 
\begin{align}
    \mathrm{ATC}(h) = \frac{1}{|X_{OOD}|} \sum_{x \in X_{OOD}} \mathbf{1}\{s(h(x)) > t\}
\end{align}

where $s(h(x)) = \sum_j h_j(x) \log(h_j(x))$ and $t$ satisfies
\begin{align}
    \frac{1}{|X_{val}|} \sum_{x \in X_{val}} \mathbf{1}\{s(h(x)) < t\} = 1 -  \mathsf{\widehat{Acc}_{ID}}(h)
\end{align}

\paragraph{DOC-Feat}
\begin{align}
    \mathrm{DOC}(h) = \mathsf{\widehat{Acc}_{ID}(h)} + \frac{1}{|X_{Val}|} \sum_{x \in X_{Val}} \max_k h_k(x) - \frac{1}{|X_{OOD}|} \sum_{x \in X_{OOD}} \max_k h_k(x)
\end{align}
\paragraph{Average Confidence (AC)} Given model $h$ and unlabeled OOD data, 
\begin{align}
    \mathrm{AC}(h_\theta) = \frac{1}{|X_{OOD}|} \sum_{x \in X_{OOD}} \max_k h_k(x)
\end{align}

\paragraph{Agreement} Given a pair of models $h, h'$, Agreement simply predicts their average OOD accuracy to be the agreement between the models $\mathsf{\widehat{Agr}_{OOD}}(h, h')$.

\newpage
\section{Theoretical Analysis from Miller et al.}
\label{sec:appendix}
To get a better understanding of the agreement-on-the-line phenomenon, we replicate a theoretical experiments in \citet{Miller1} using the same set of neural networks from their testbed. Specifically, we look at CIFAR-10 with different added Gaussian noise (their Figure 4).

\subsection{Matching gaussian noise}
\citet{Miller1} conduct a theoretical analysis on a toy gaussian mixture model to better understand the accuracy-on-the-line phenomena. From their analysis, they predict that accuracy-on-the-line occurs if the covariances of the ID and OOD data are the same up to some constant scaling factor. Inspired by this, they show that accuracy-on-the-line holds stronger on CIFAR-10 data corrupted with gaussian noise that matches the covariance of CIFAR-10 test data versus isotropic gaussian noise. Interestingly, even for this simple setting, we similarly observe that the ID vs OOD agreement trend is stronger on covariance matched gaussian noise.

\begin{figure}[H]
    \centering
    \includegraphics[scale=0.2]{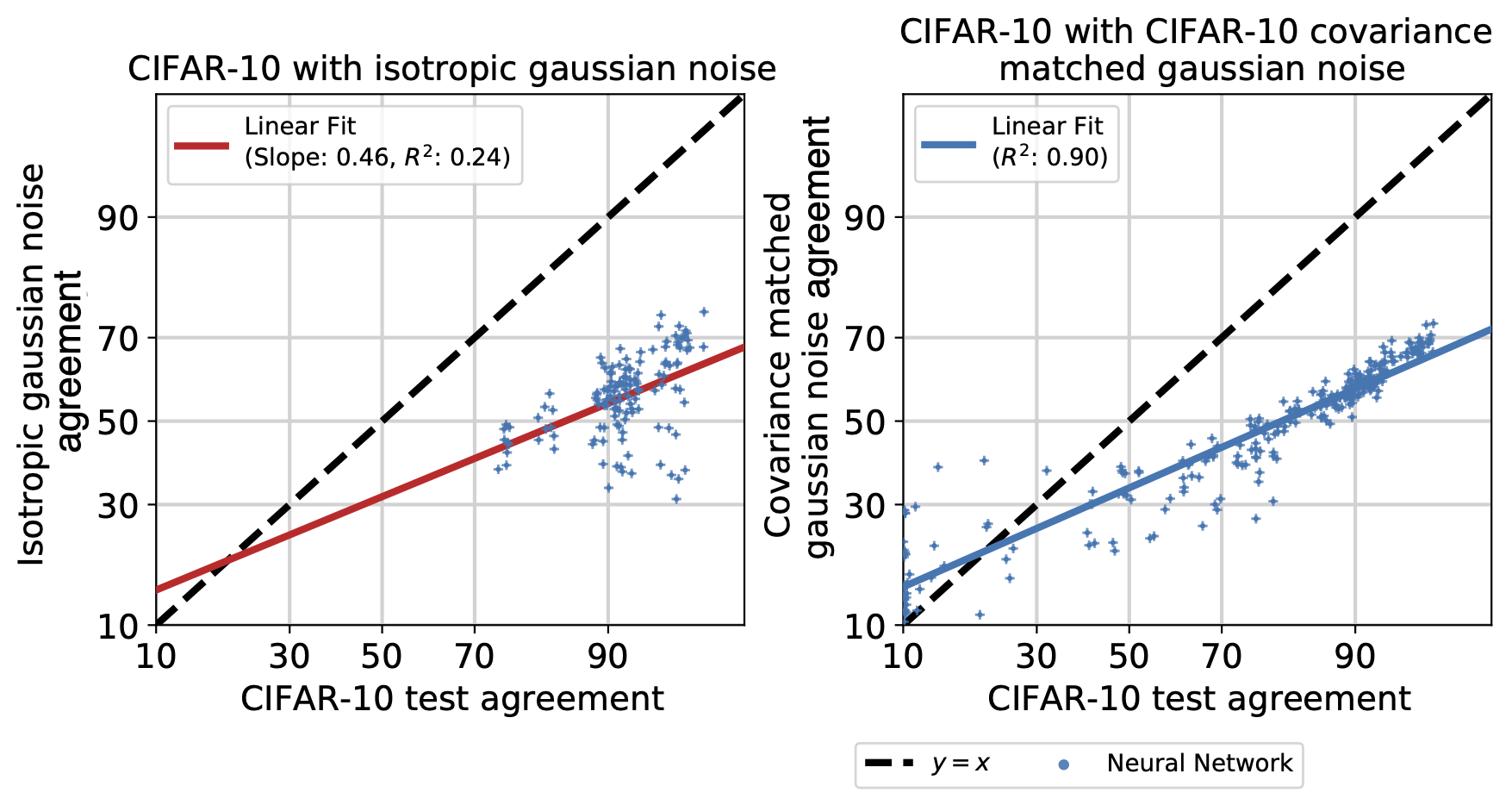}
    \caption{We look at ID vs OOD agreement of CIFAR-10 with isotropic gaussian noise versus covariance matched gaussian noise. See Figure 4 in \protect{\citet{Miller1}}}
    \label{fig:matched_gaussian}
\end{figure}

\newpage
\section{Correlation results on more datasets}
\label{app:more_dataset}
To verify the agreement-on-the-line phenomenon, we compare the linear correlation between ID vs OOD accuracy and agreement across a variety of distribution shifts. Given a set of $n$ models, we plot the ID vs OOD accuracy of each model in the set. For agreement, we randomly pair each model with another model in the set, and plot the ID vs OOD agreement of these $n$ pairs. We provide $R^2$ values in the figure legends.

\paragraph{Datasets} In addition to the 8 datasets from the main body, we also observe the trend on other CIFAR10C corruptions \cite{hendrycks_cifar10c}. For each corruption, we evaluate the models over data from all 5 severity levels (both in Figure 1 of main body and appendix). 

We also look at shifting from CIFAR-10 to CINIC-10 \cite{cinic10} and CIFAR-10 to STL-10 \cite{stl10} which are shifts from changes in the image source. Specifically, the CINIC-10 test dataset contains both CIFAR-10 Test data and a selection of ImageNet images for CIFAR-10 class labels downsampled to $32 \times 32$. We only consider the downsampled ImageNet data as the OOD dataset. Similarly, STL-10 contains processed ImageNet images for CIFAR-10 class labels. Since STL-10 is an unsupervised learning dataset, we only utilize the labeled subset of STL-10 as the OOD dataset. Additionally, STL-10 only contains 9 of the 10 CIFAR-10 classes, so we restrict the dataset to just those 9 classes. 

Finally, we add results for a real-world shift from batch effects in images of cells in RxRx1-\textsc{wilds} and a reading comprehension dataset Amazon-SQuAD \cite{miller2020effect} which looks at the reading comprehension performance of models trained on paragraphs derived from Wikipedia articles on Amazon product reviews. 

\paragraph{Models} For ImageNetV2, we evaluate 49 ImageNet pretrained models from the timm \cite{rw2019timm} package. See their \href{https://github.com/rwightman/pytorch-image-models}{repository} for more details about the models. For RxRx1-\textsc{wilds} we trained 36 models of varying architecture and hyperparameters. Specifically, we vary weight decay between $[10^{-1}, 10^{-2}, 10^{-3},10^{-4},10^{-5}]$ and optimizers between SGD, Adam, and AdamW. See \ref{app:model} for architecture details. 

Finally, we utilize all independently trained models from the CIFAR10, iWildCam-\textsc{wilds}, fMoW-\textsc{wilds}, and Camelyon17-\textsc{wilds} testbeds created and utilized by~\citep{Miller1}. The hyperparameters used to train these models are explained in great detail in Appendix B.2 of \citet{Miller1}. The architectures of all models evaluated for the experiment are described in more detail in Appendix~\ref{app:model}.

\paragraph{Pretrained vs Not Pretrained} \citet{Miller1} showed that for some shifts, the ID vs OOD accuracy of ImageNet pretrained models follow a different linear trend than models trained from scratch in-distribution. In Figure 1 of the main body, we do not distinguish between pretrained and from scratch models as the trends for pretrained and from scratch models were the same for the 8 datasets we chose. Below, we divide results between pretrained and not pretrained models to be more precise. \footnote{The fMoW-\textsc{wilds} testbed also contains two models pretrained on CLIP} \footnote{For CINIC-10 and STL-10, we only look at from scratch models as the shifted images are derived from ImageNet.}

\subsection{Over randomly initialized models}
\begin{figure}[H]
    \centering %
\begin{subfigure}{0.3\textwidth}
  \includegraphics[width=\linewidth]{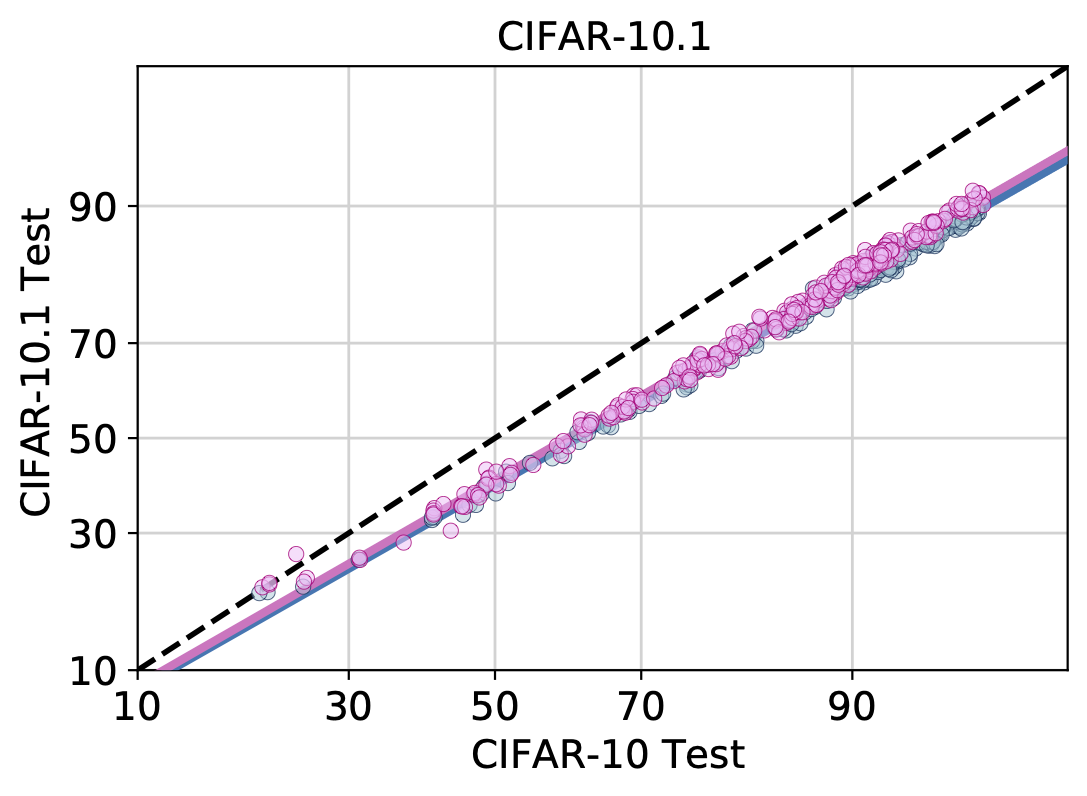}
  \caption{CIFAR10.1}
\end{subfigure}\hfil %
\begin{subfigure}{0.3\textwidth}
  \includegraphics[width=\linewidth]{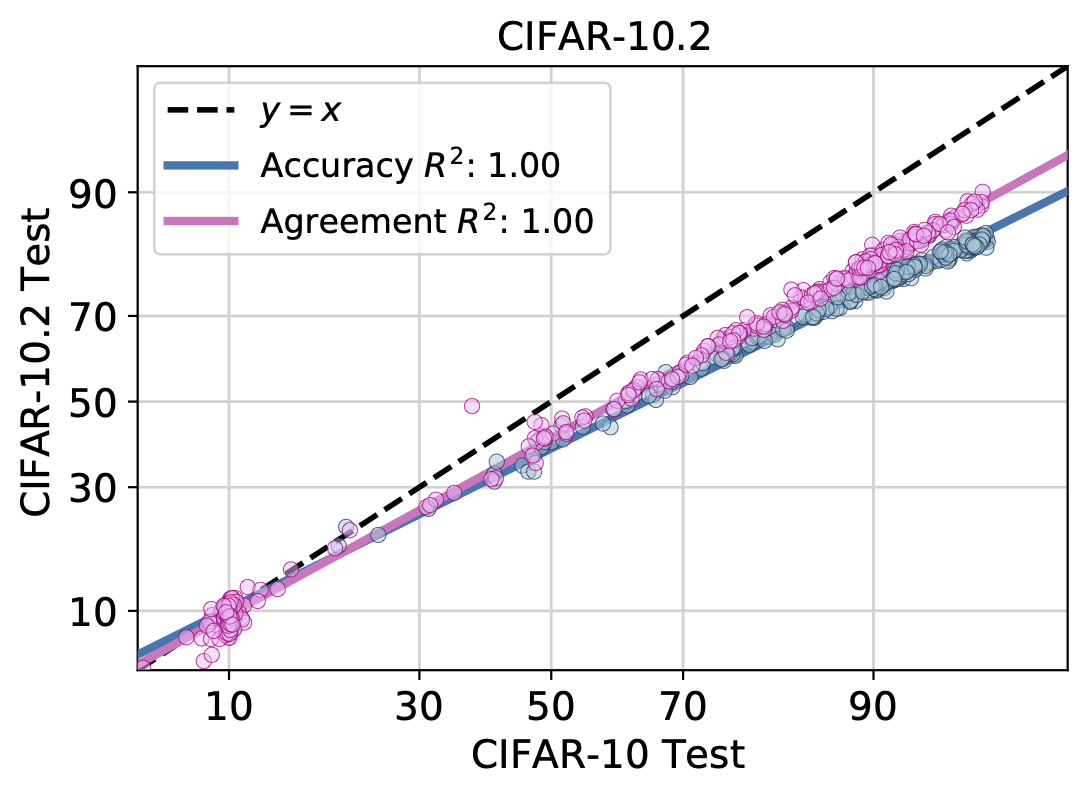}
  \caption{CIFAR10.2}
\end{subfigure}\hfil %
\begin{subfigure}{0.3\textwidth}
  \includegraphics[width=\linewidth]{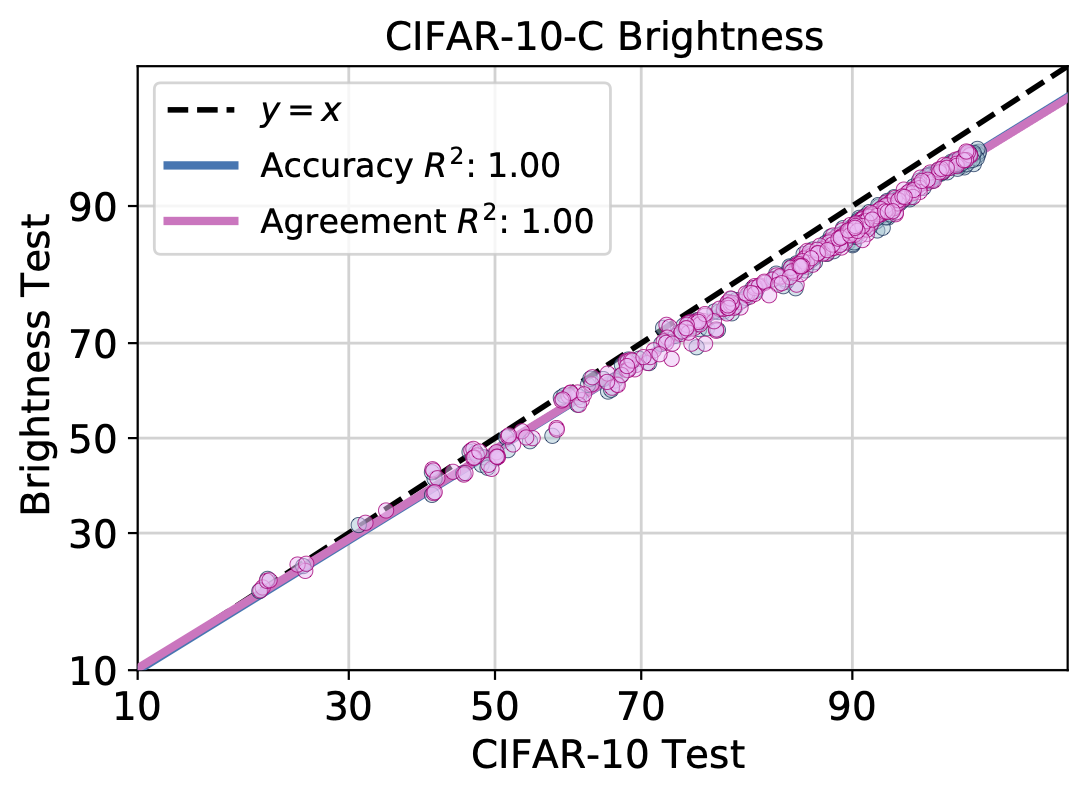}
  \caption{CIFAR10C Brightness}
\end{subfigure}
\medskip
\begin{subfigure}{0.3\textwidth}
  \includegraphics[width=\linewidth]{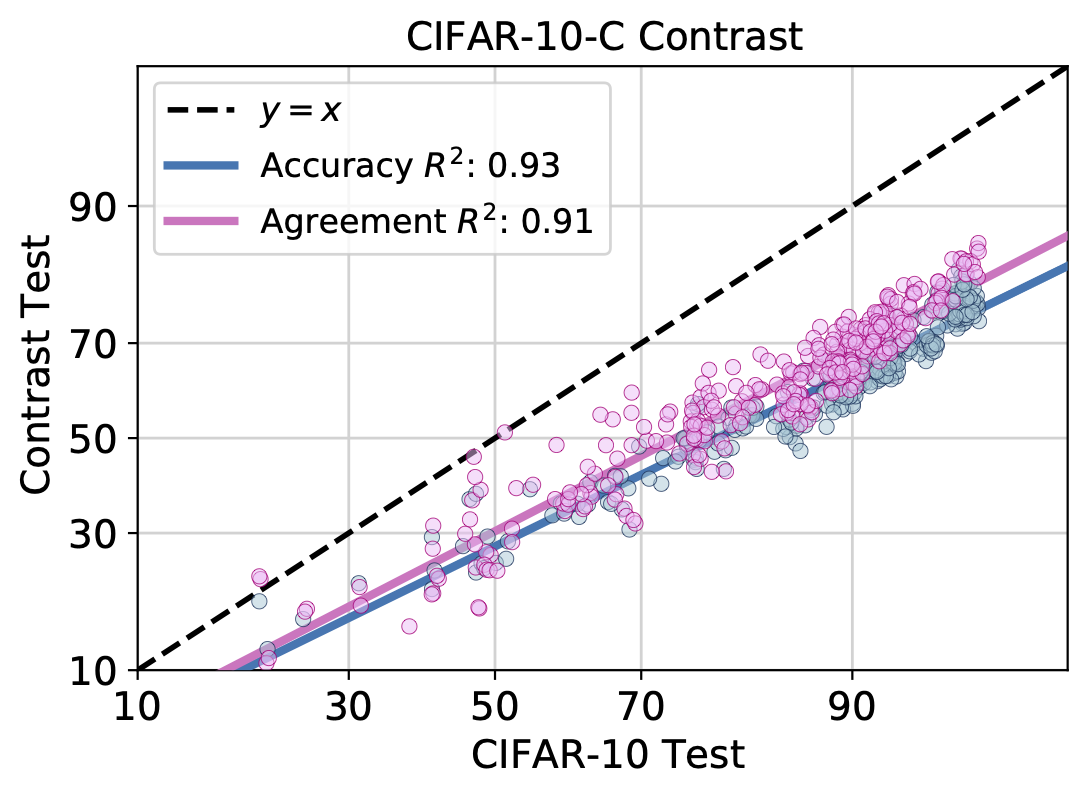}
  \caption{CIFAR10C Contrast}
\end{subfigure}\hfil %
\begin{subfigure}{0.3\textwidth}
  \includegraphics[width=\linewidth]{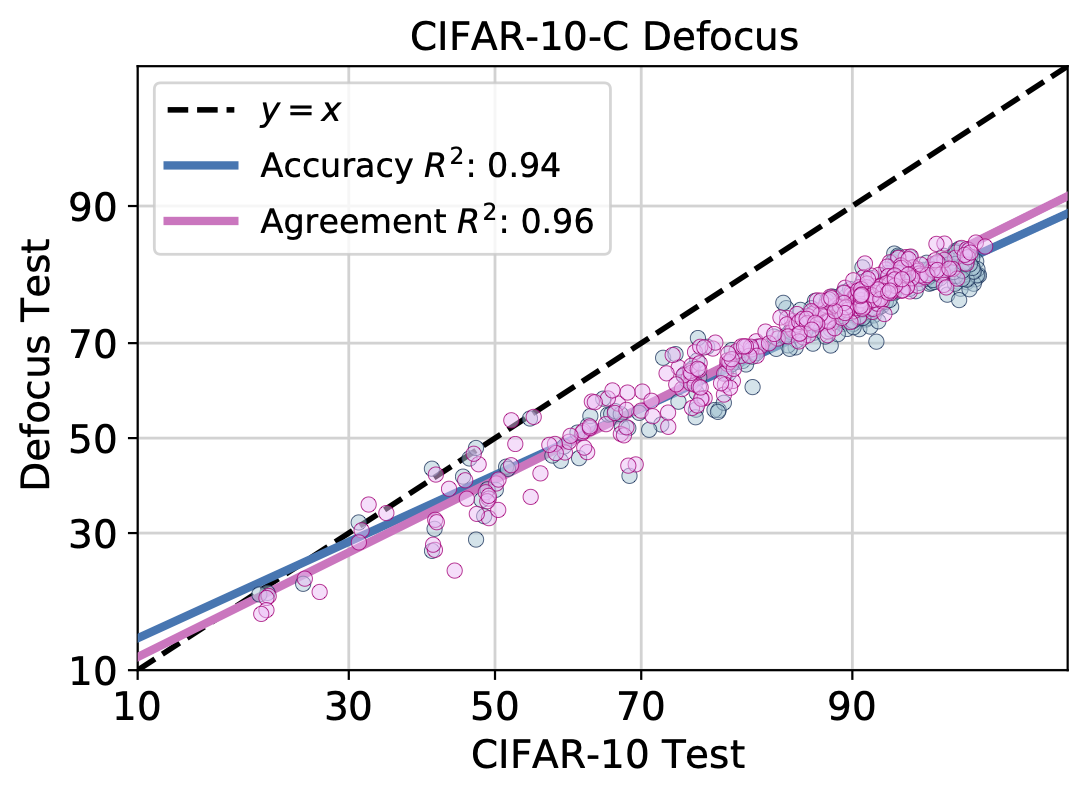}
  \caption{CIFAR10C Defocus Blur}
\end{subfigure}\hfil %
\begin{subfigure}{0.3\textwidth}
  \includegraphics[width=\linewidth]{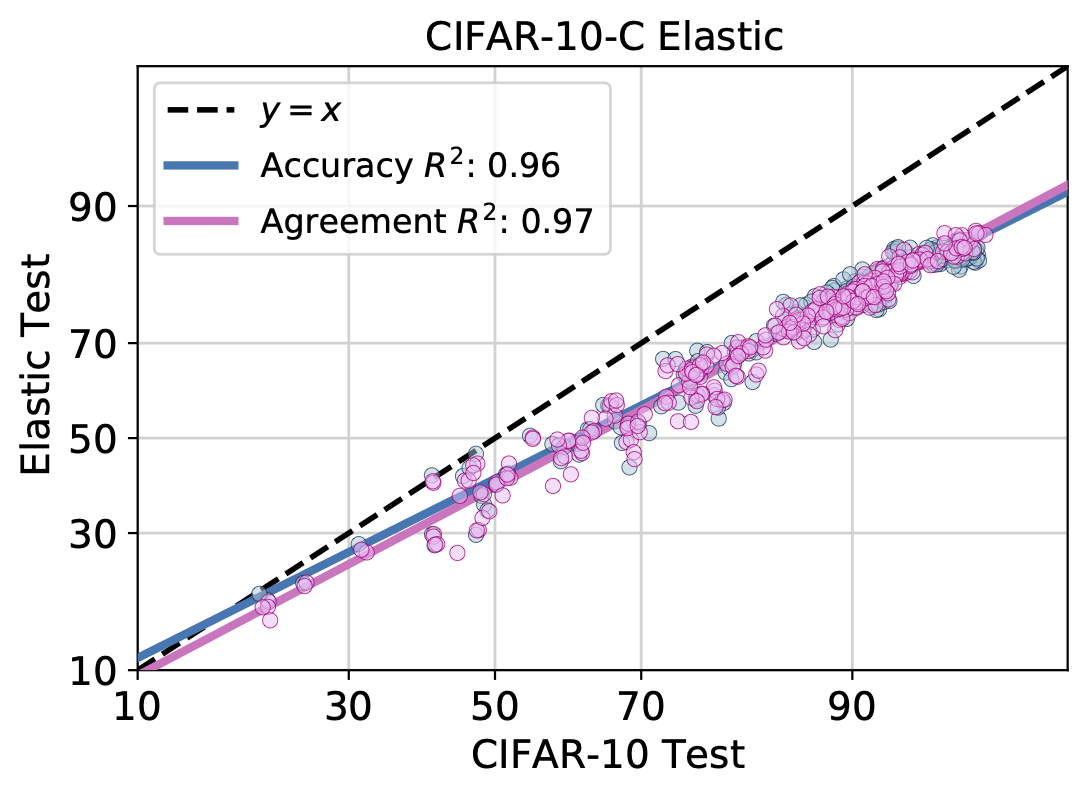}
  \caption{CIFAR10C Elastic Transform}
\end{subfigure}
\medskip
\begin{subfigure}{0.3\textwidth}
  \includegraphics[width=\linewidth]{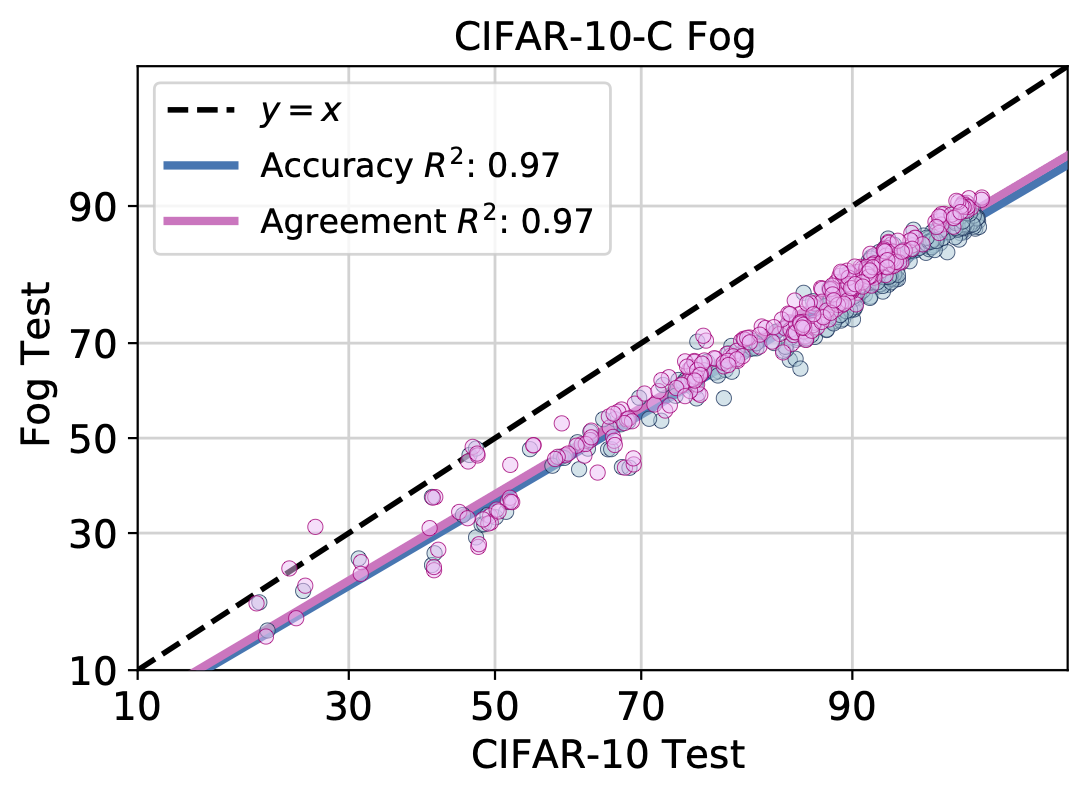}
  \caption{CIFAR10C Fog}
\end{subfigure}\hfil %
\begin{subfigure}{0.3\textwidth}
  \includegraphics[width=\linewidth]{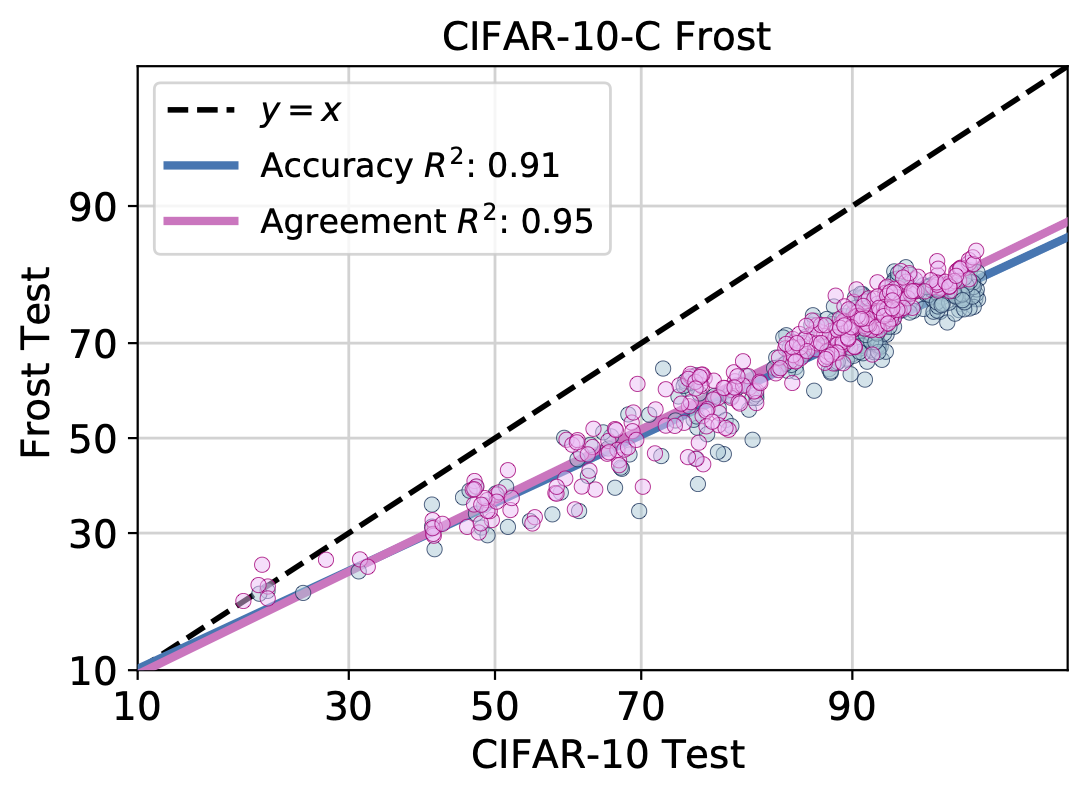}
  \caption{CIFAR10C Frost}
\end{subfigure}\hfil %
\begin{subfigure}{0.3\textwidth}
  \includegraphics[width=\linewidth]{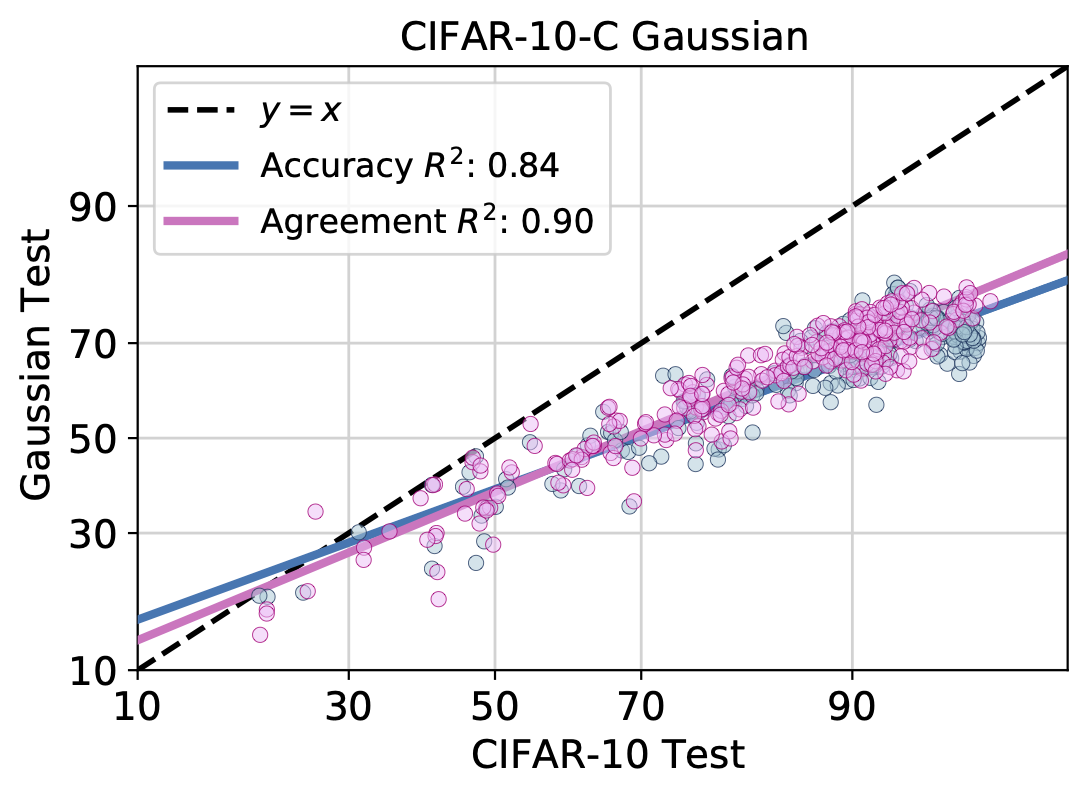}
  \caption{CIFAR10C Gaussian Blur}
\end{subfigure}
\medskip
\begin{subfigure}{0.3\textwidth}
  \includegraphics[width=\linewidth]{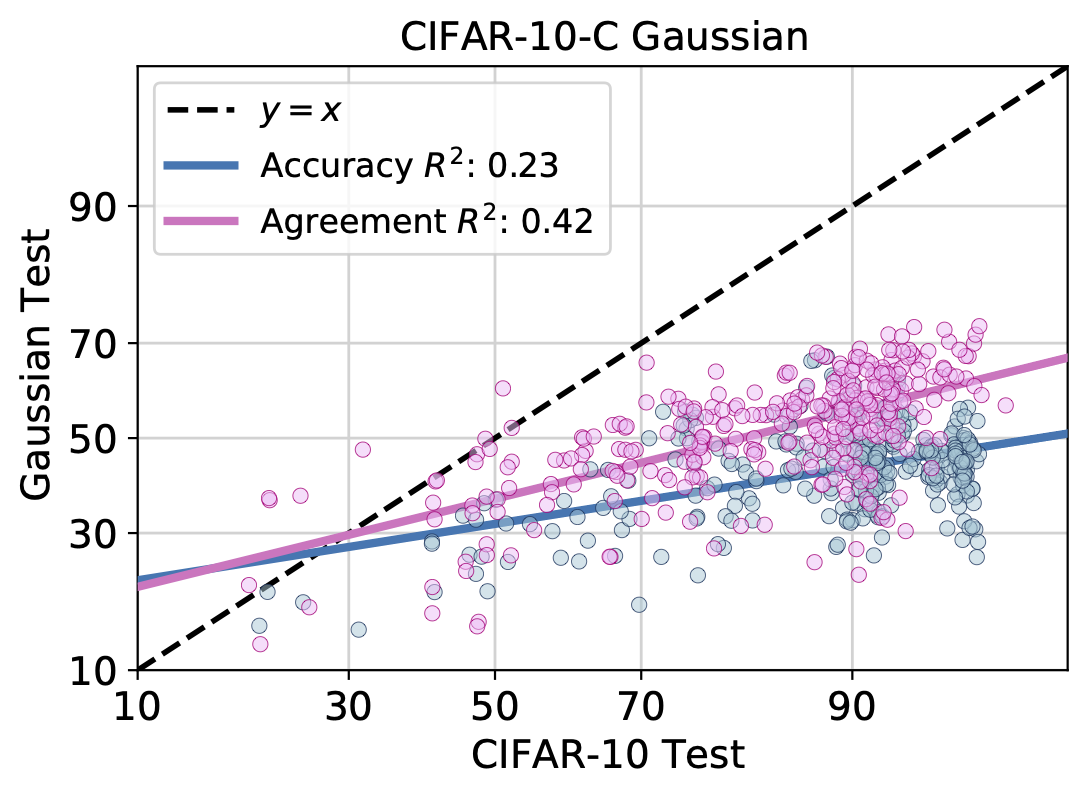}
  \caption{CIFAR10C Gaussian Noise}
\end{subfigure}\hfil %
\begin{subfigure}{0.3\textwidth}
  \includegraphics[width=\linewidth]{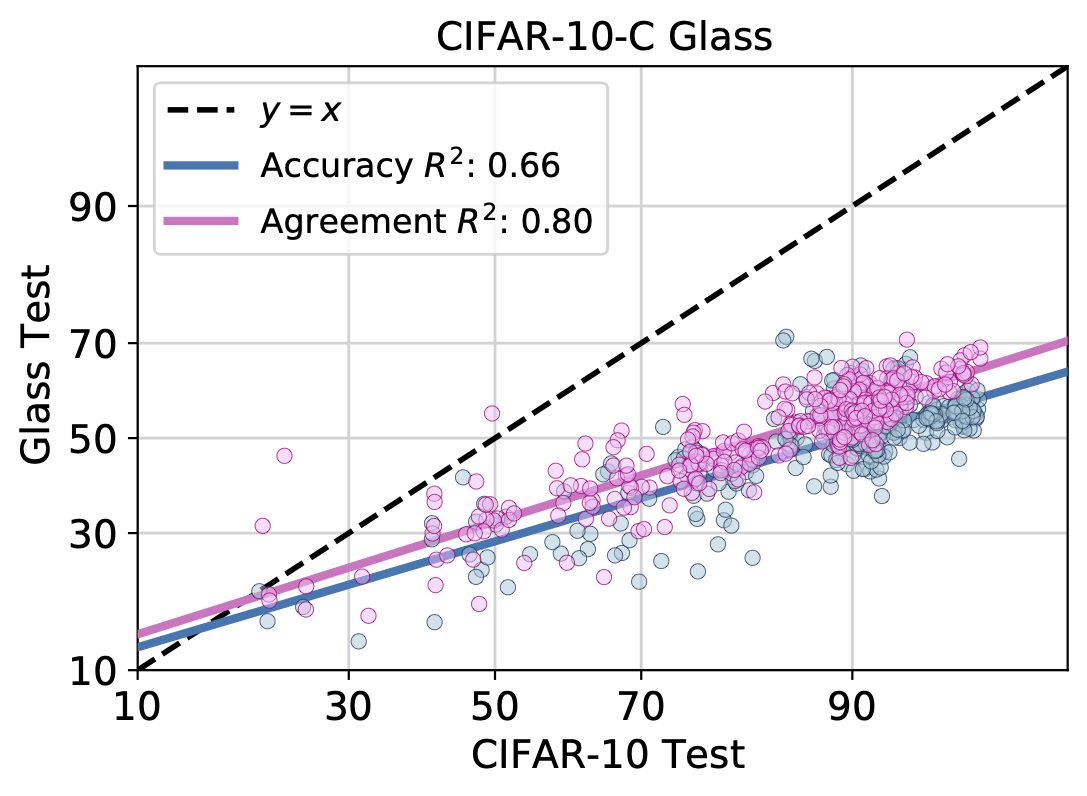}
  \caption{CIFAR10C Glass Blur}
\end{subfigure}\hfil %
\begin{subfigure}{0.3\textwidth}
  \includegraphics[width=\linewidth]{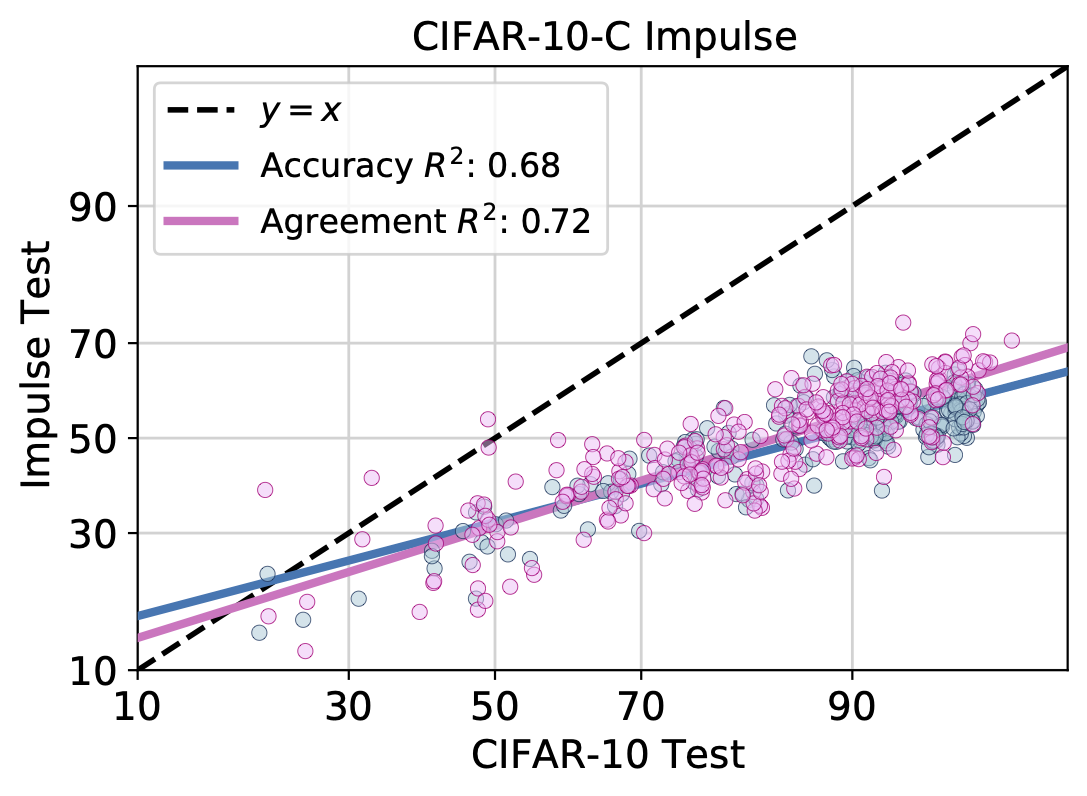}
  \caption{CIFAR10C Impulse Noise}
\end{subfigure}
\medskip
\begin{subfigure}{0.3\textwidth}
  \includegraphics[width=\linewidth]{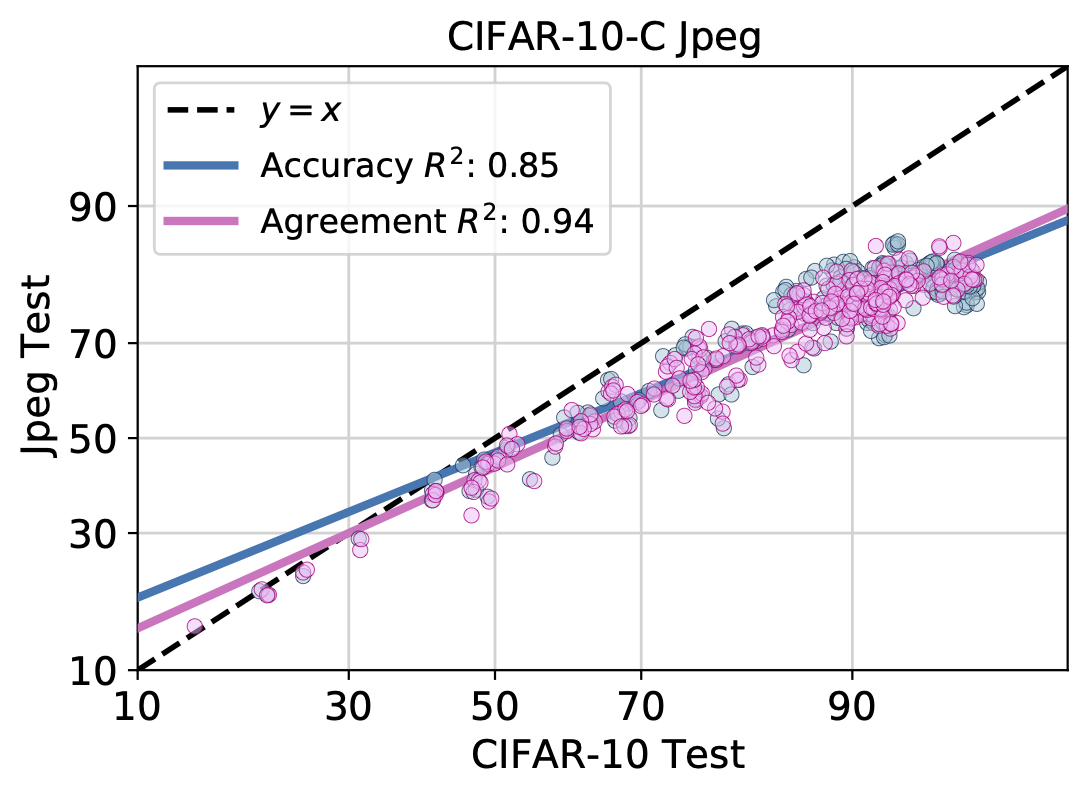}
  \caption{CIFAR10C JPEG Compression}
\end{subfigure}\hfil %
\begin{subfigure}{0.3\textwidth}
  \includegraphics[width=\linewidth]{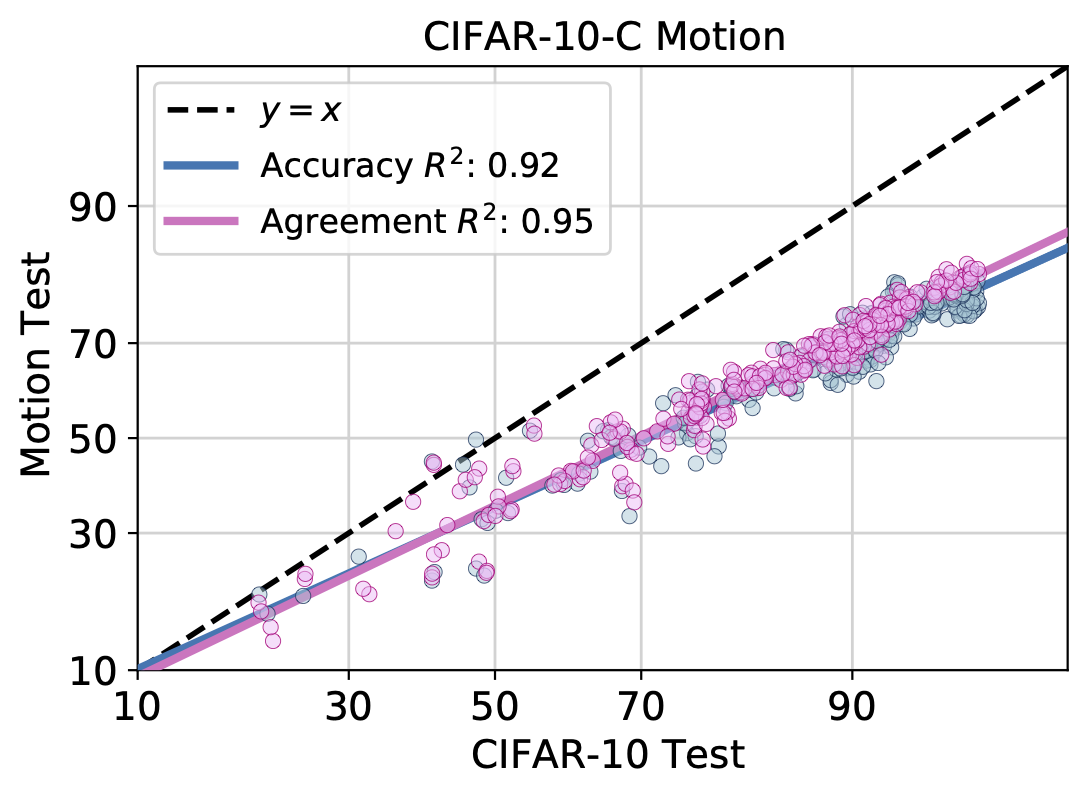}
  \caption{CIFAR10C Motion Blur}
\end{subfigure}\hfil %
\begin{subfigure}{0.3\textwidth}
  \includegraphics[width=\linewidth]{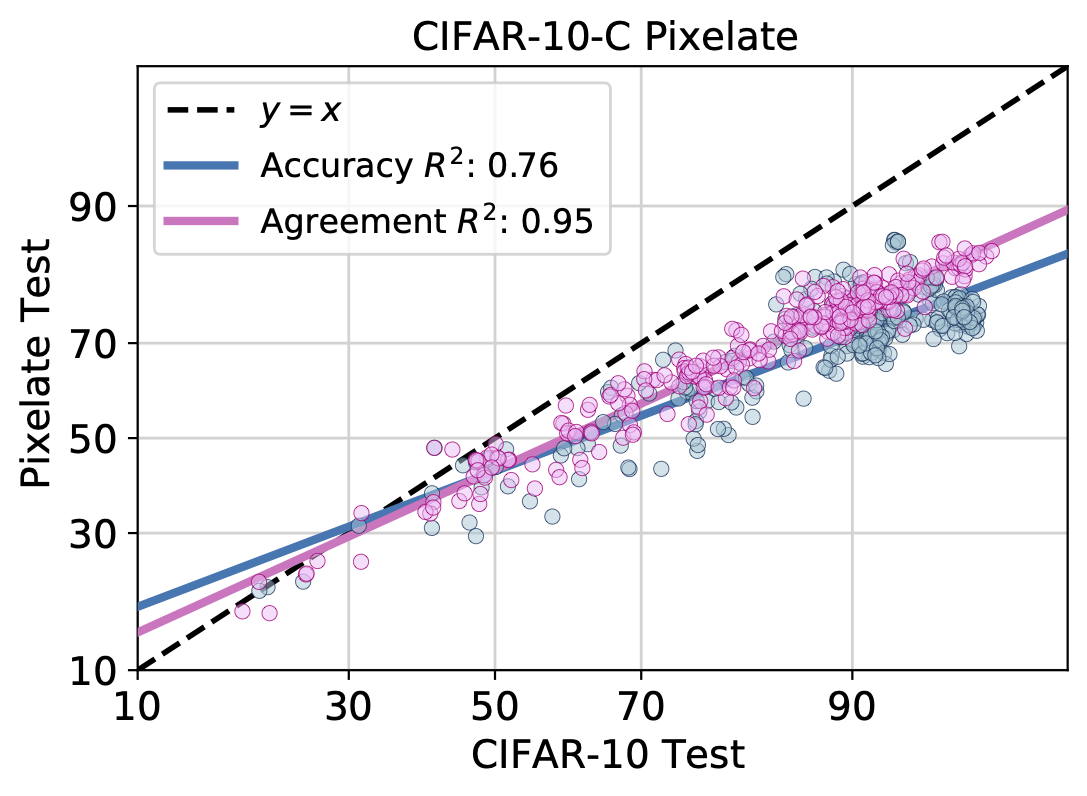}
  \caption{CIFAR10C Pixelate}
\end{subfigure}
\end{figure}
\begin{figure}[H]\ContinuedFloat
\centering
\medskip
\begin{subfigure}{0.3\textwidth}
  \includegraphics[width=\linewidth]{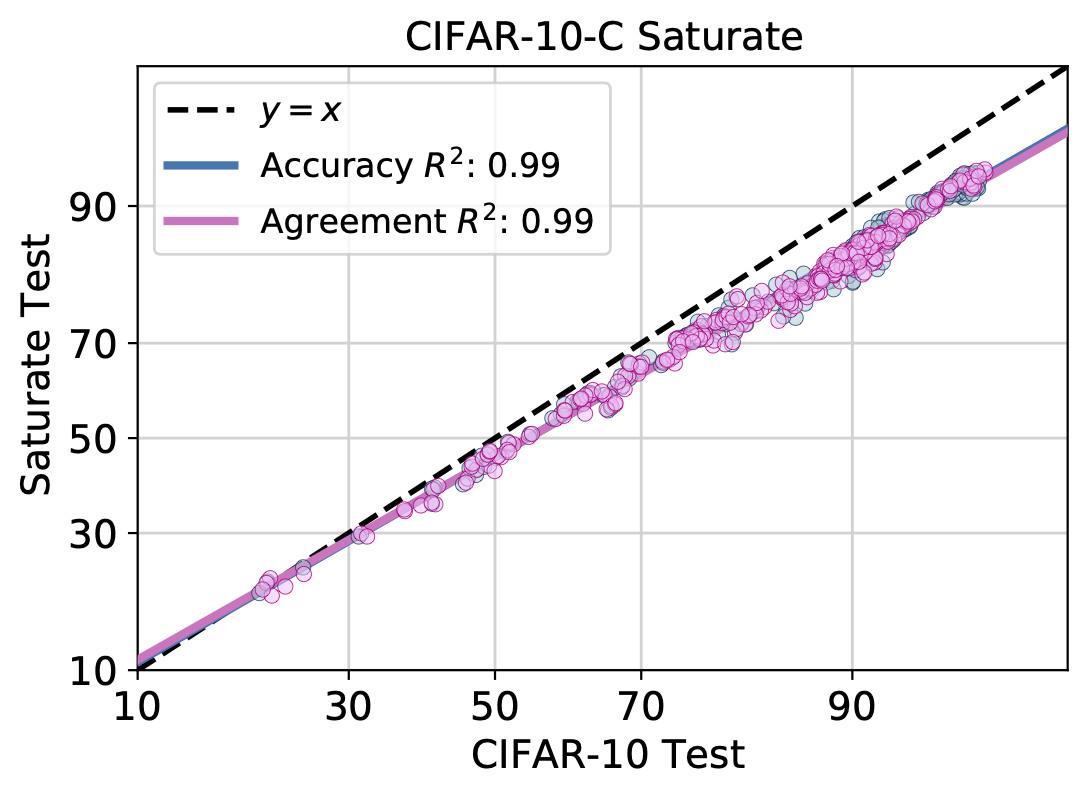}
  \caption{CIFAR10C Saturate}
\end{subfigure}\hfil %
\begin{subfigure}{0.3\textwidth}
  \includegraphics[width=\linewidth]{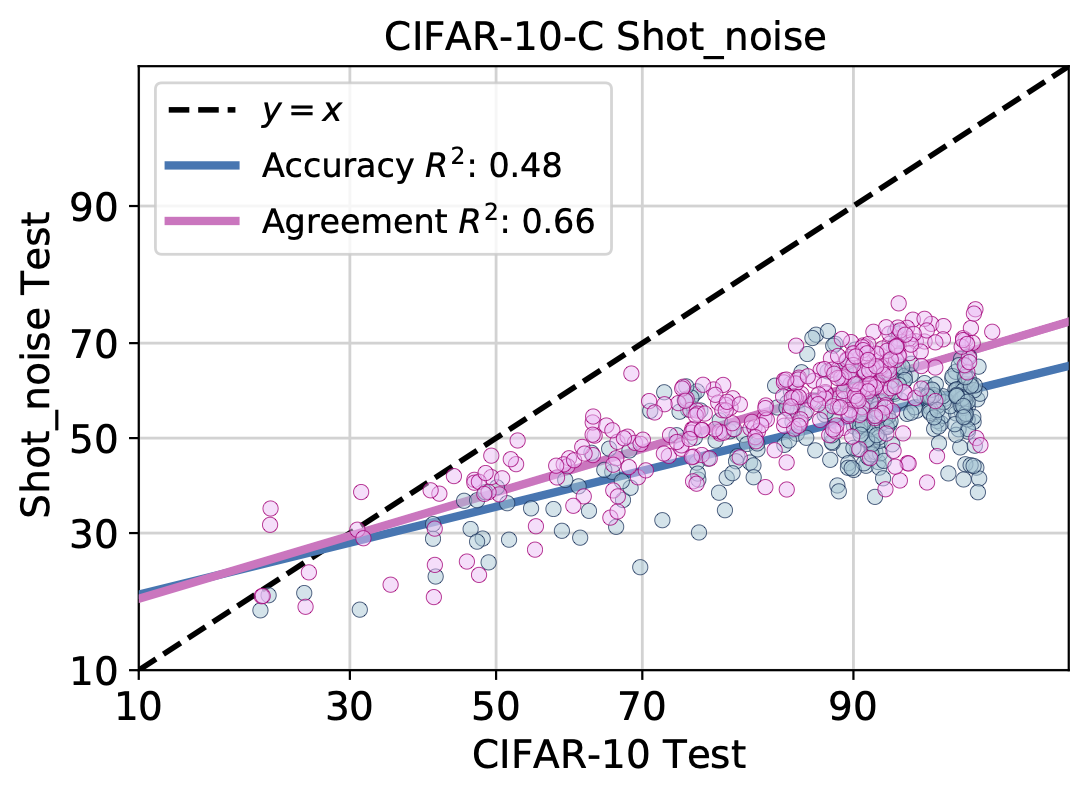}
  \caption{CIFAR10C Shot Noise}
\end{subfigure}\hfil %
\begin{subfigure}{0.3\textwidth}
  \includegraphics[width=\linewidth]{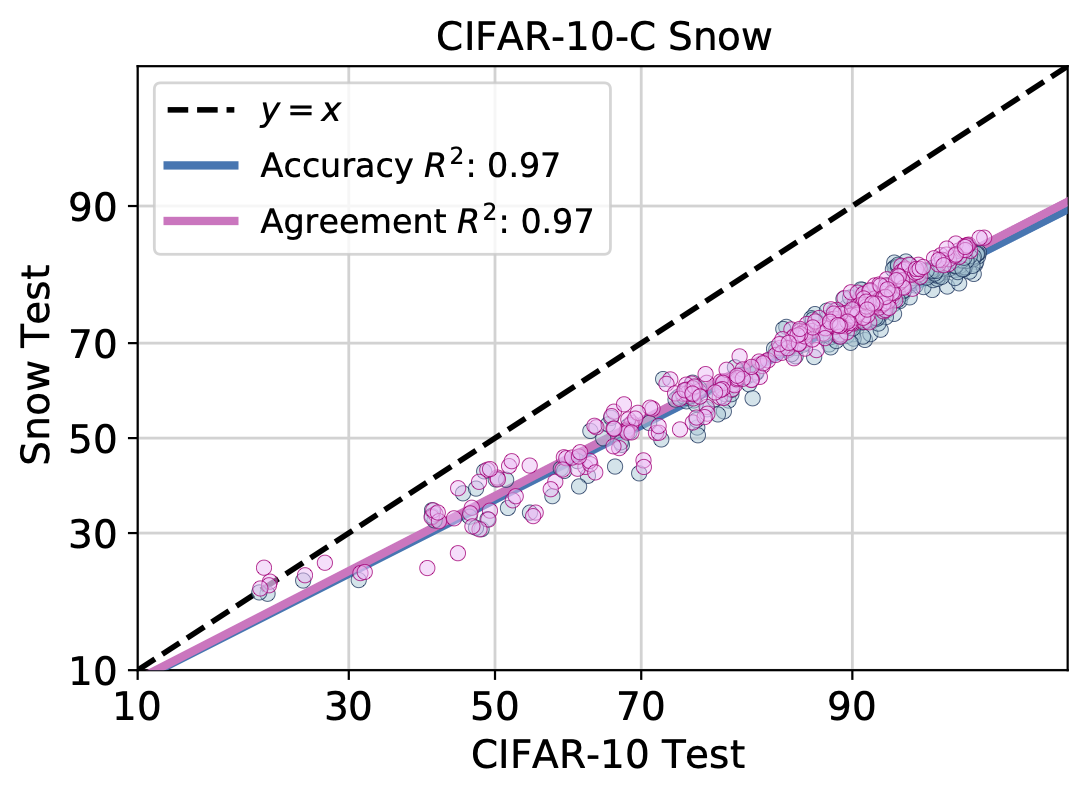}
  \caption{CIFAR10C Snow}
\end{subfigure}
\medskip
\begin{subfigure}{0.3\textwidth}
  \includegraphics[width=\linewidth]{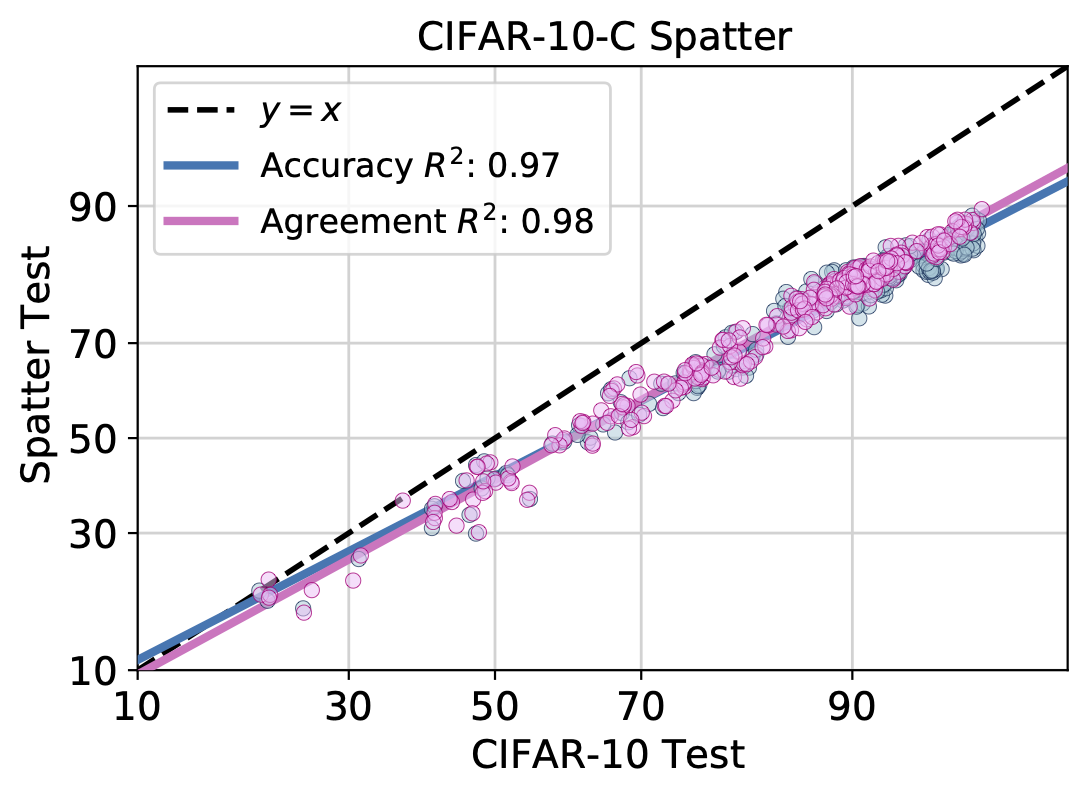}
  \caption{CIFAR10C Spatter}
\end{subfigure}\hfil %
\begin{subfigure}{0.3\textwidth}
  \includegraphics[width=\linewidth]{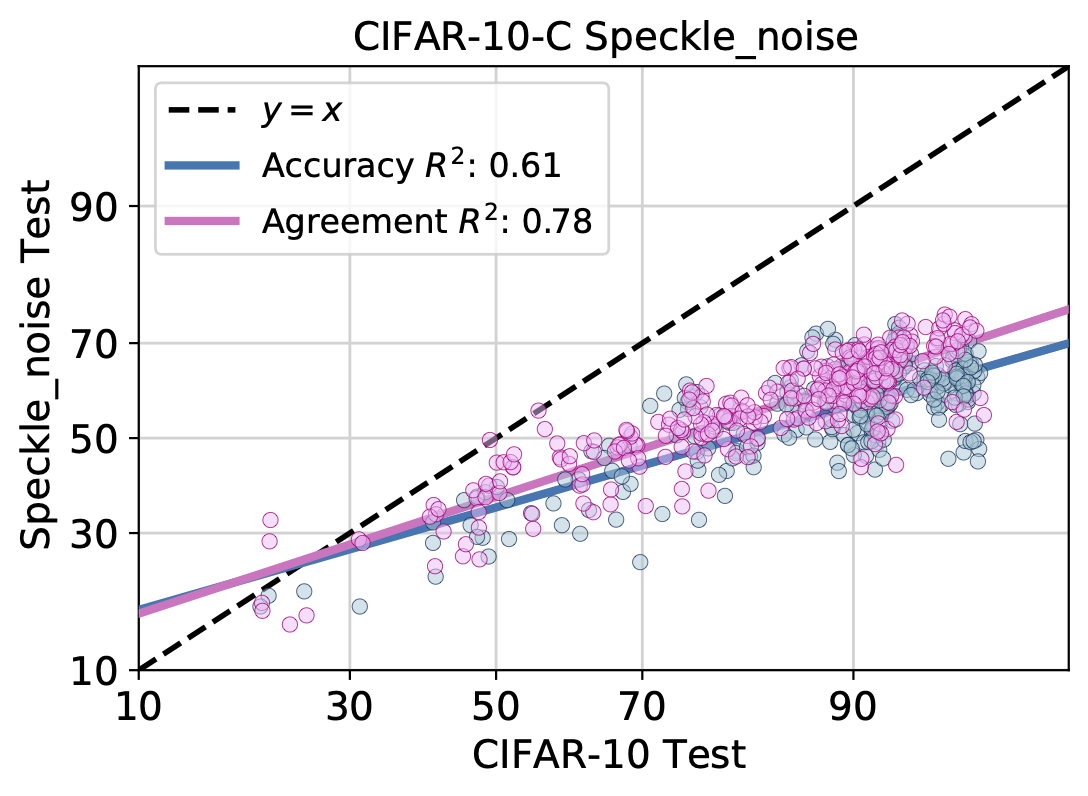}
  \caption{CIFAR10C Speckle Noise}
\end{subfigure}\hfil %
\begin{subfigure}{0.3\textwidth}
  \includegraphics[width=\linewidth]{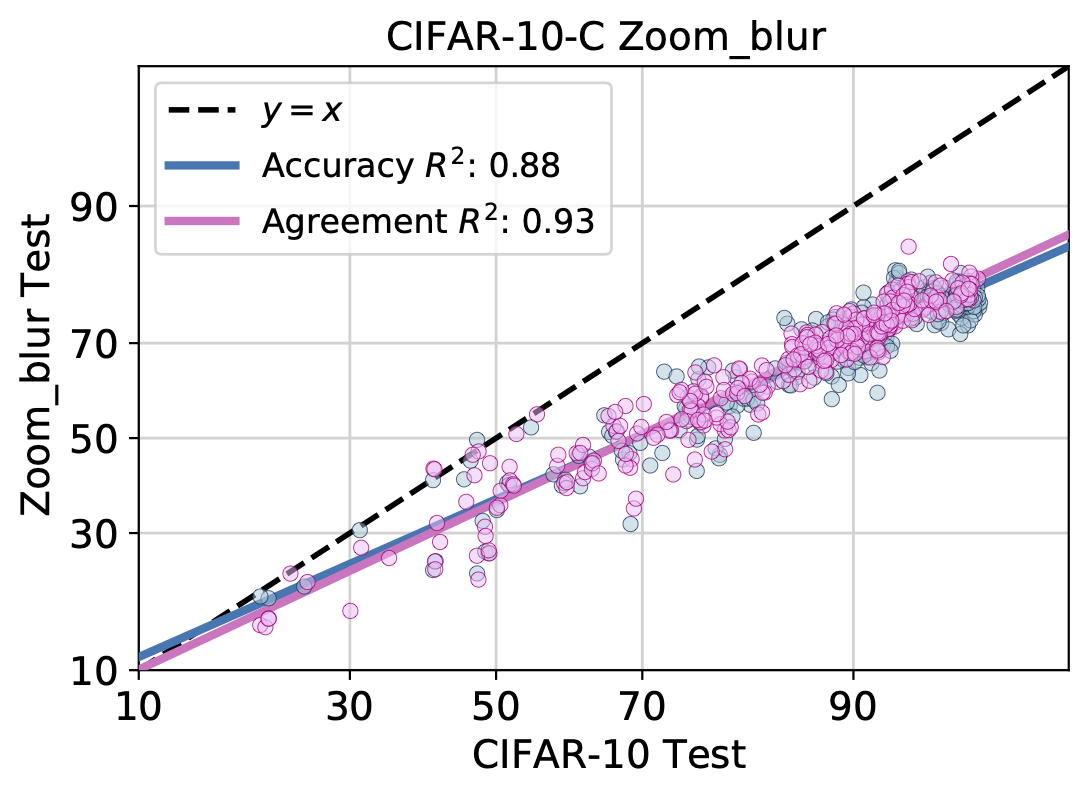}
  \caption{CIFAR10C Zoom Blur}
\end{subfigure}
\medskip
\begin{subfigure}{0.3\textwidth}
  \includegraphics[width=\linewidth]{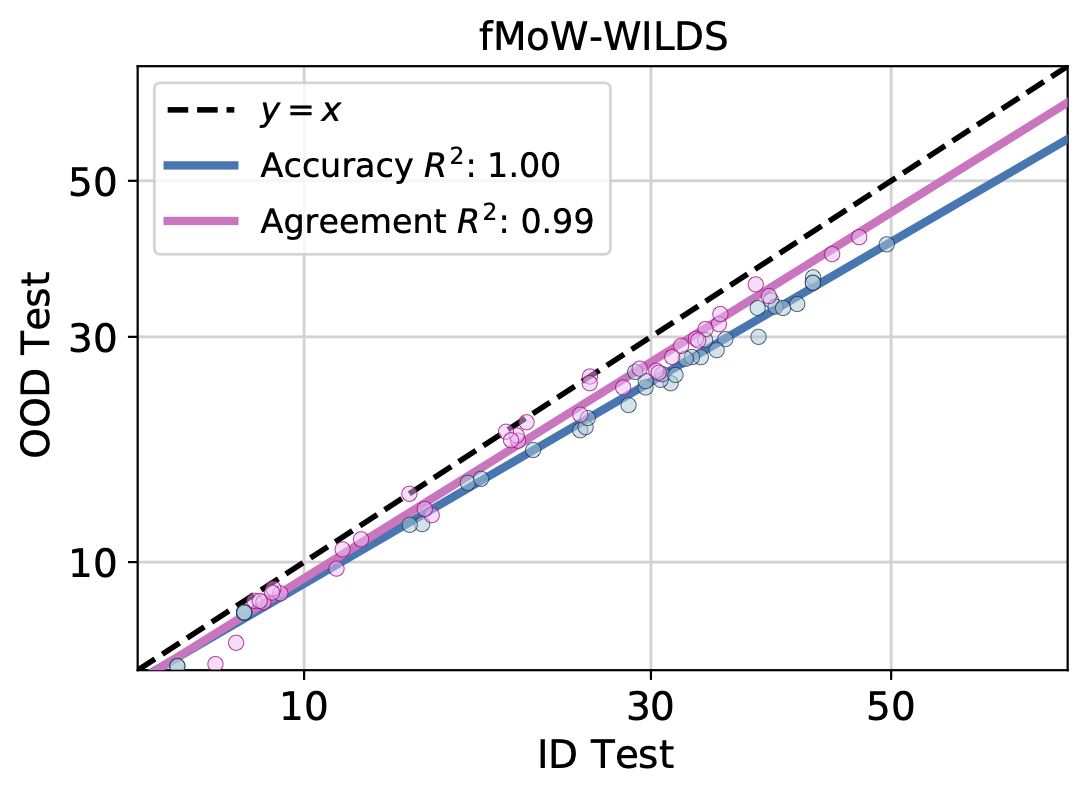}
  \caption{fMoW-\textsc{wilds}}
\end{subfigure}\hfil %
\begin{subfigure}{0.3\textwidth}
  \includegraphics[width=\linewidth]{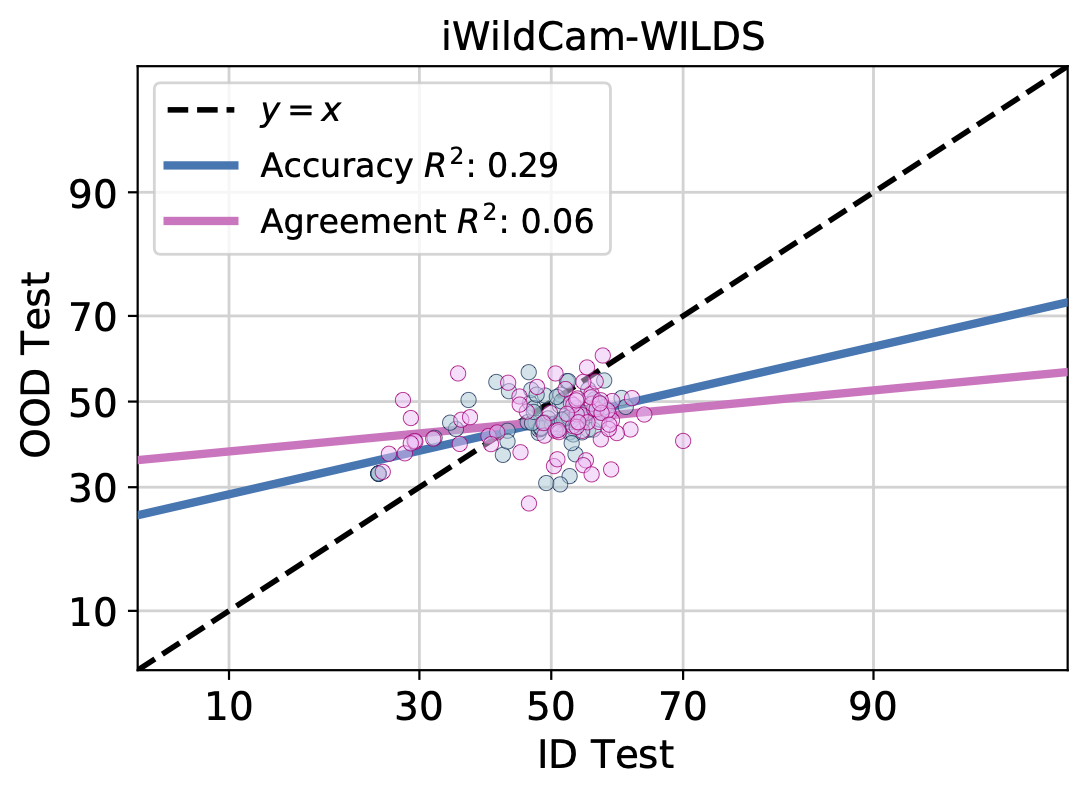}
  \caption{iWildCam-\textsc{wilds}}
\end{subfigure}\hfil %
\begin{subfigure}{0.3\textwidth}
  \includegraphics[width=\linewidth]{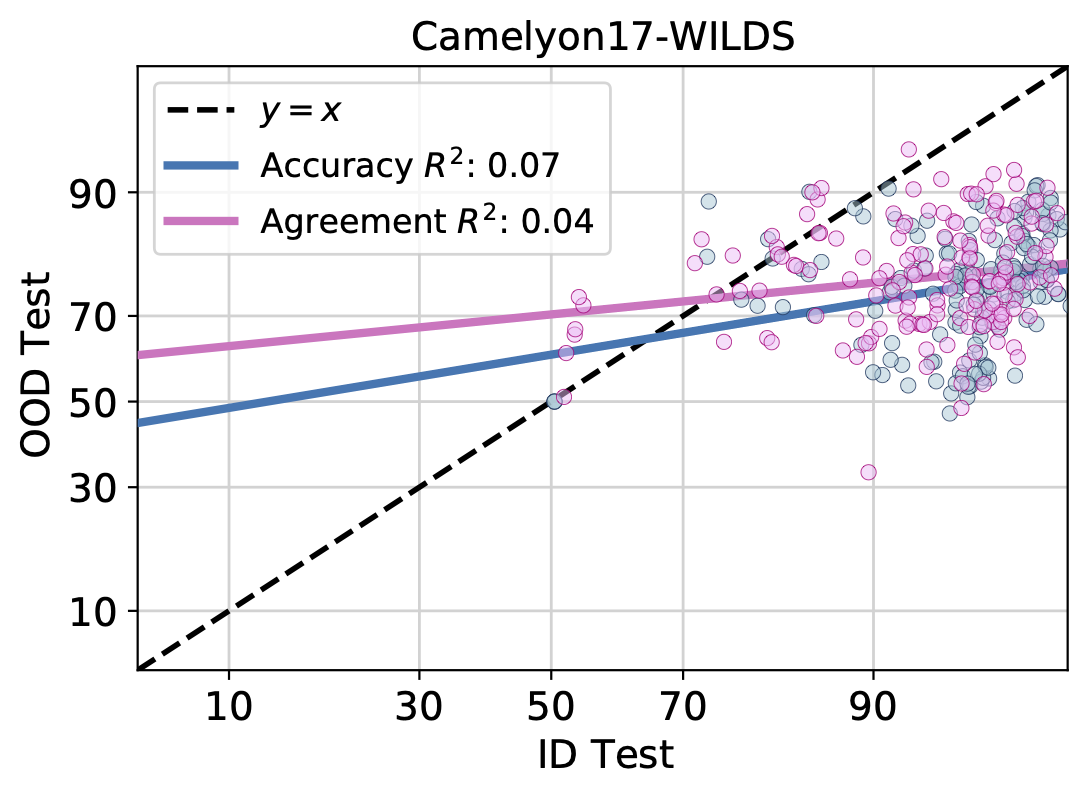}
  \caption{Camelyon17-\textsc{wilds}}
\end{subfigure}
\medskip
\begin{subfigure}{0.3\textwidth}
  \includegraphics[width=\linewidth]{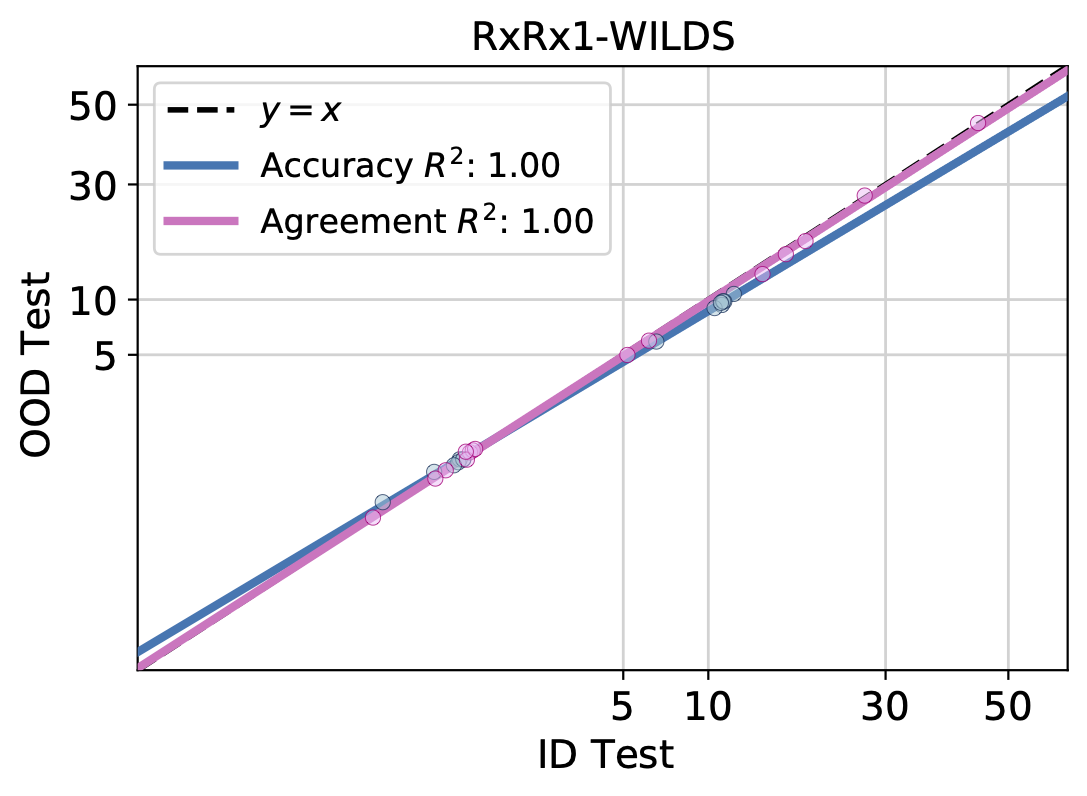}
  \caption{RxRx1-\textsc{wilds}}
\end{subfigure}
\begin{subfigure}{0.3\textwidth}
  \includegraphics[width=\linewidth]{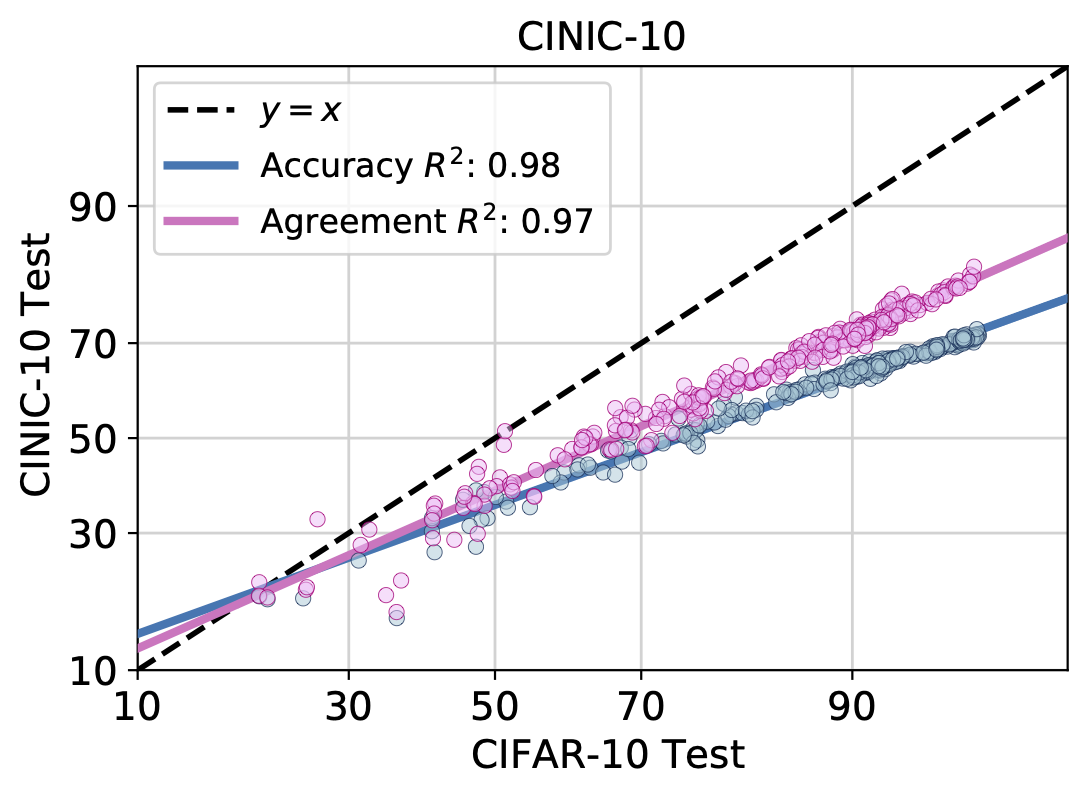}
  \caption{CINIC-10}
\end{subfigure}\hfil %
\begin{subfigure}{0.3\textwidth}
  \includegraphics[width=\linewidth]{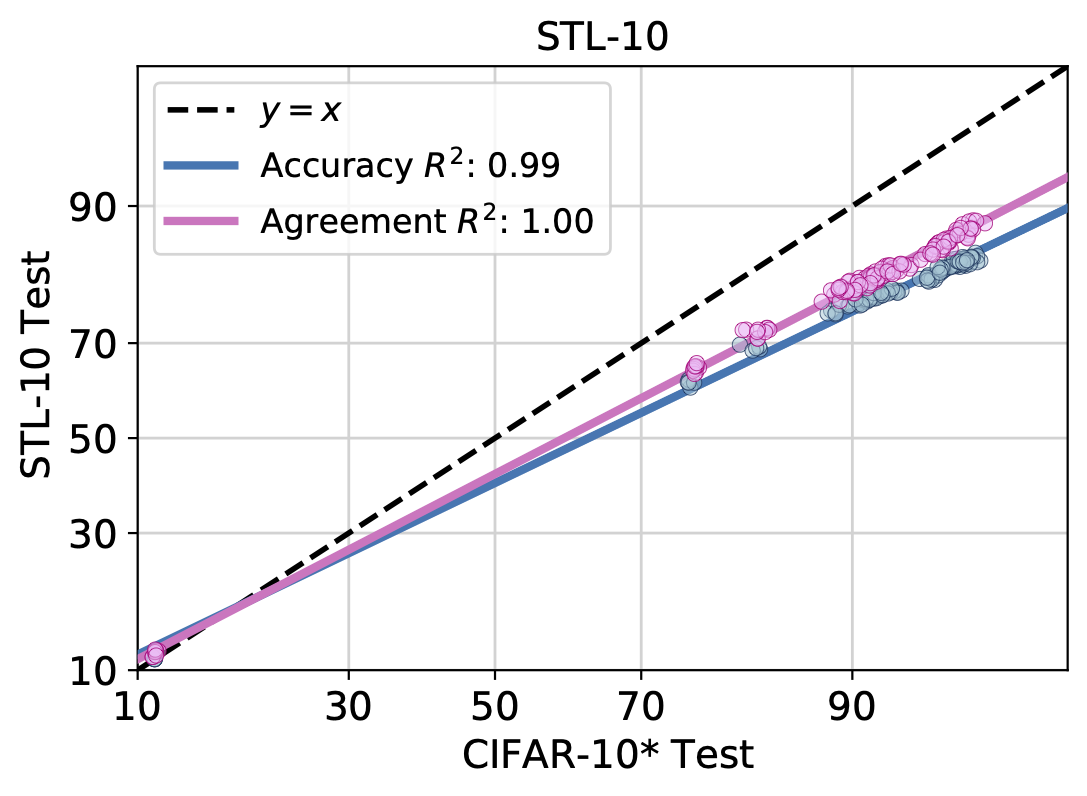}
  \caption{STL-10}
\end{subfigure}

\end{figure}

\newpage
\subsection{Over models pretrained on ImageNet}
\begin{figure}[H]
    \centering %
\begin{subfigure}{0.3\textwidth}
  \includegraphics[width=\linewidth]{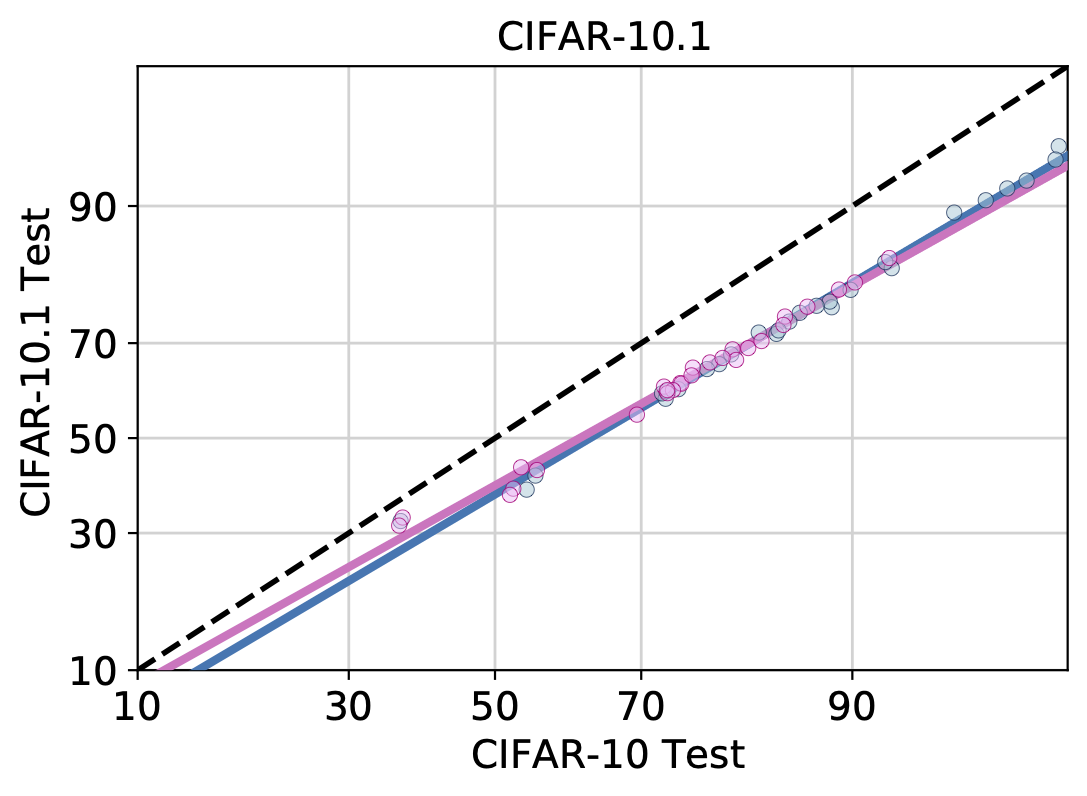}
  \caption{CIFAR10.1}
\end{subfigure}\hfil %
\begin{subfigure}{0.3\textwidth}
  \includegraphics[width=\linewidth]{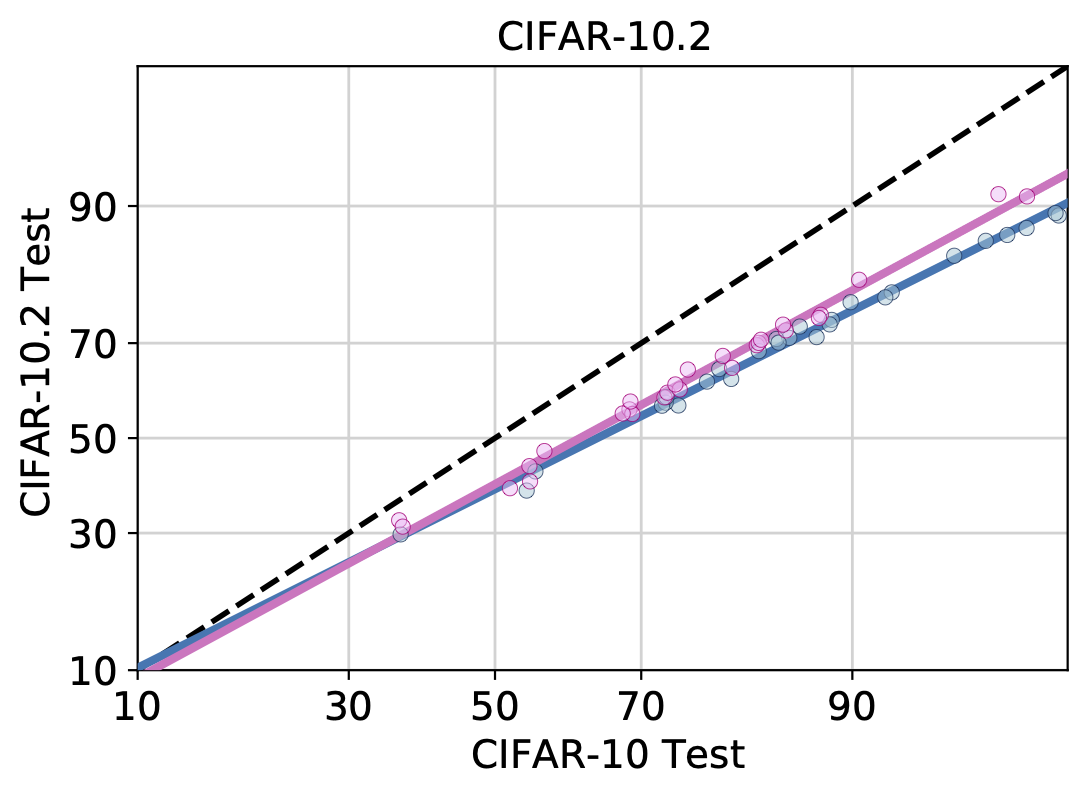}
  \caption{CIFAR10.2}
\end{subfigure}\hfil %
\begin{subfigure}{0.3\textwidth}
  \includegraphics[width=\linewidth]{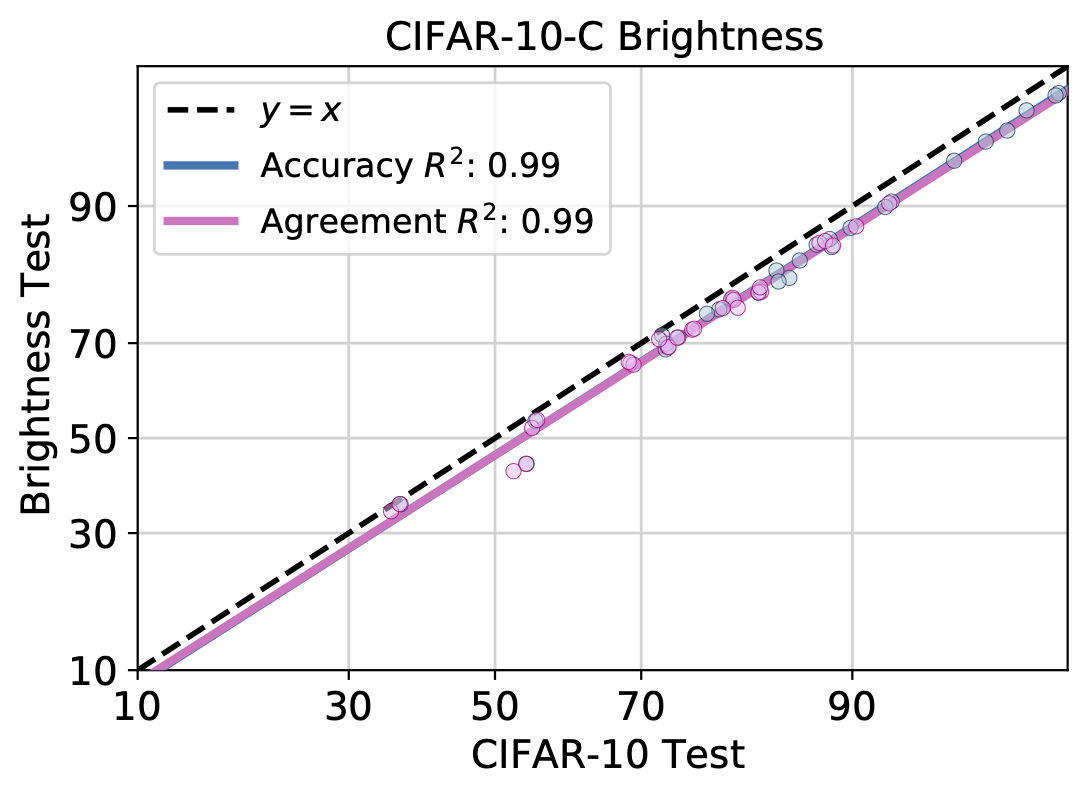}
  \caption{CIFAR10C Brightness}
\end{subfigure}
\medskip
\begin{subfigure}{0.3\textwidth}
  \includegraphics[width=\linewidth]{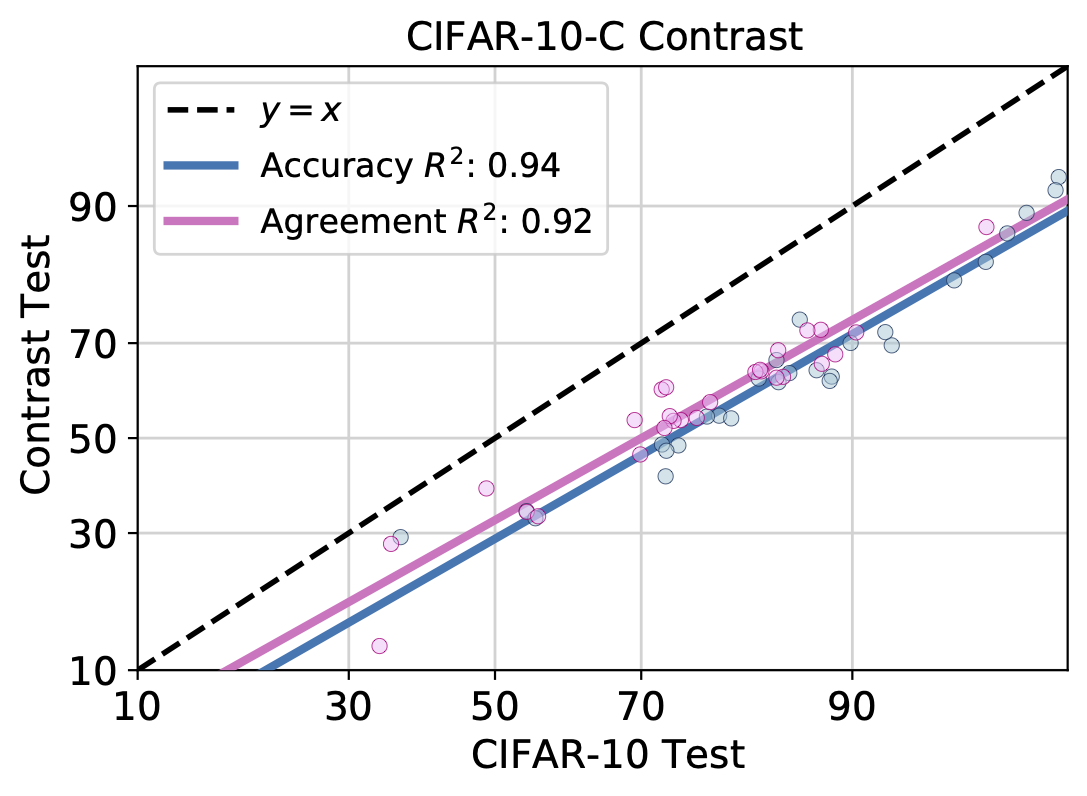}
  \caption{CIFAR10C Contrast}
\end{subfigure}\hfil %
\begin{subfigure}{0.3\textwidth}
  \includegraphics[width=\linewidth]{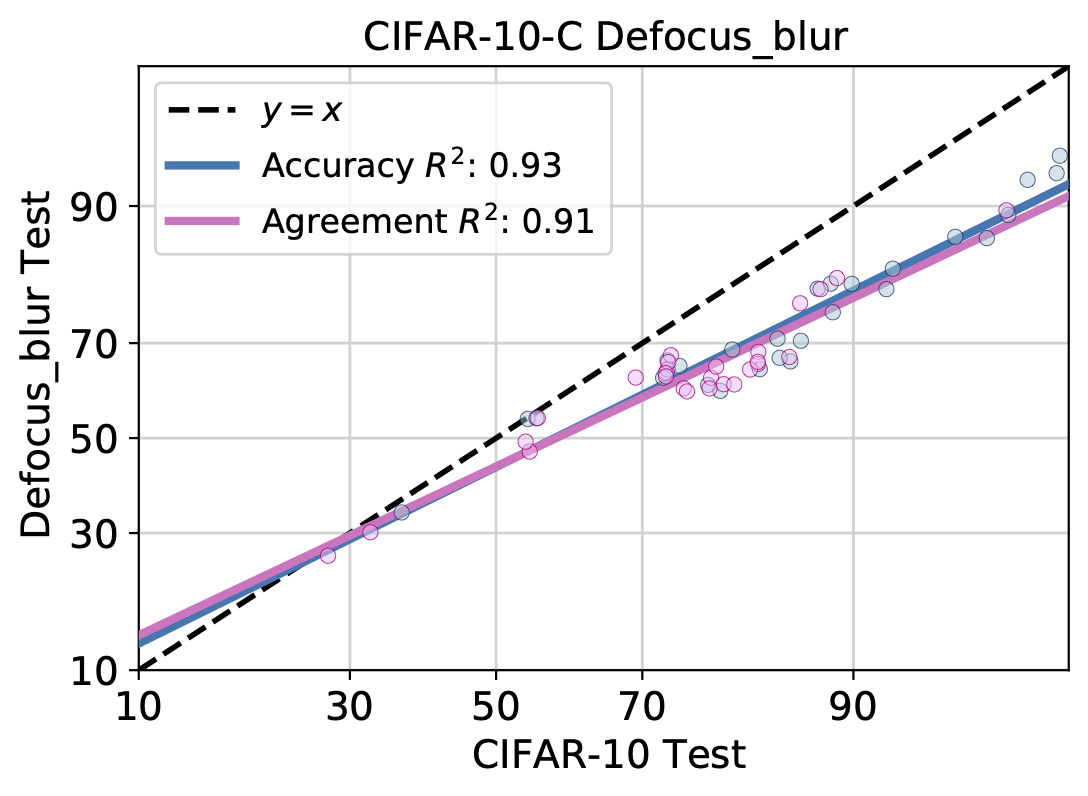}
  \caption{CIFAR10C Defocus Blur}
\end{subfigure}\hfil %
\begin{subfigure}{0.3\textwidth}
  \includegraphics[width=\linewidth]{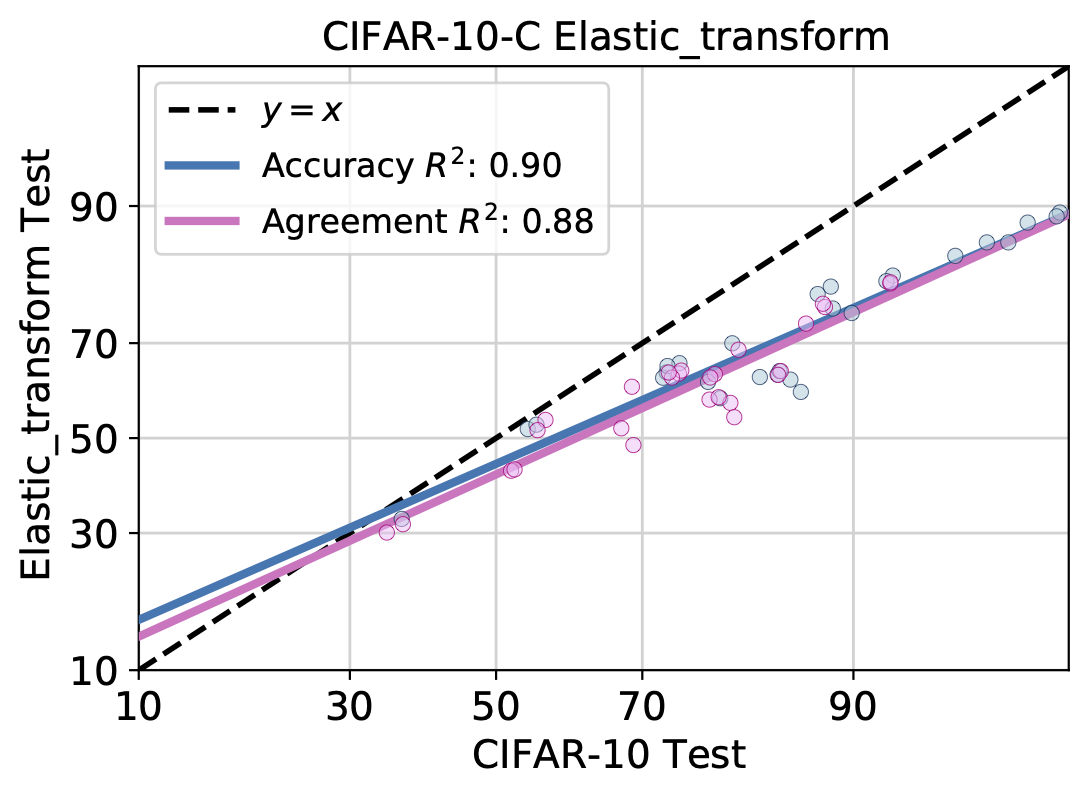}
  \caption{CIFAR10C Elastic Transform}
\end{subfigure}
\medskip
\begin{subfigure}{0.3\textwidth}
  \includegraphics[width=\linewidth]{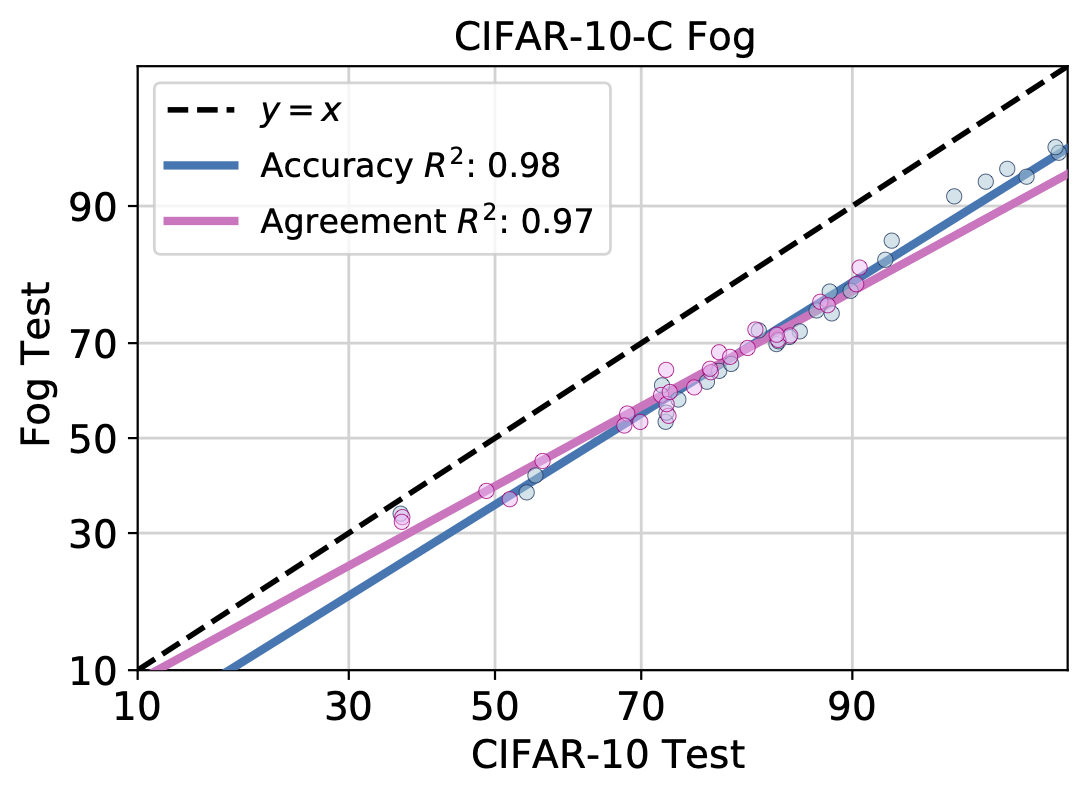}
  \caption{CIFAR10C Fog}
\end{subfigure}\hfil %
\begin{subfigure}{0.3\textwidth}
  \includegraphics[width=\linewidth]{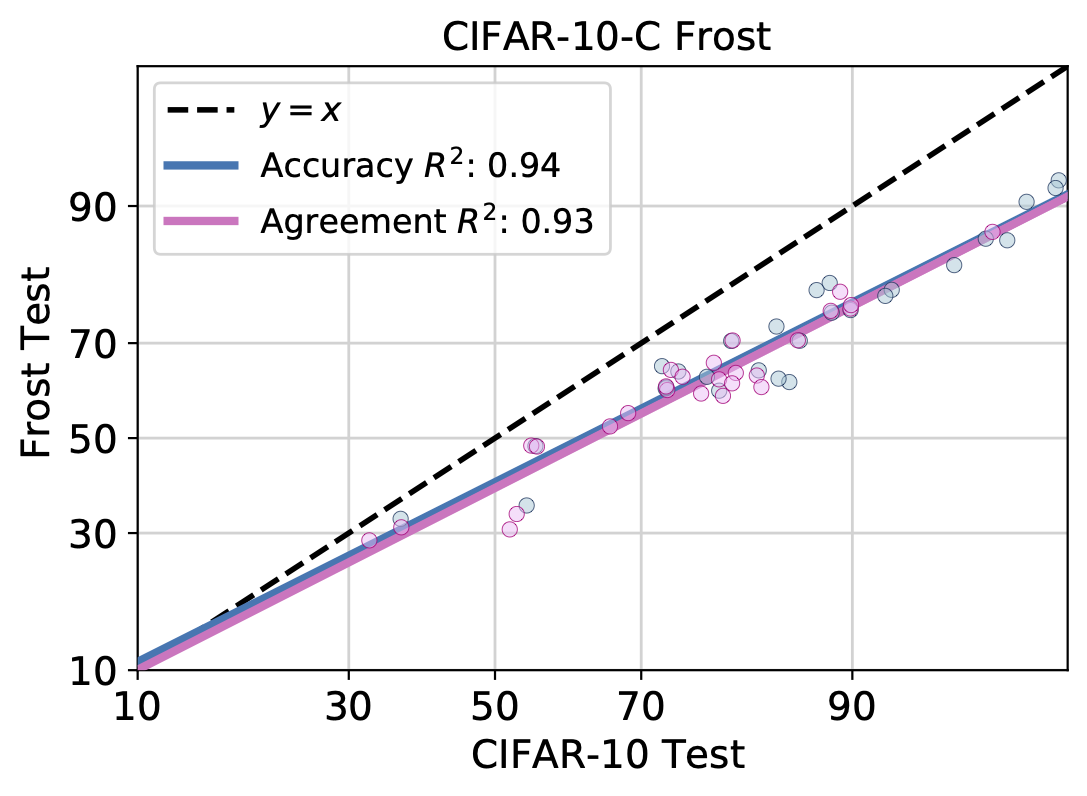}
  \caption{CIFAR10C Frost}
\end{subfigure}\hfil %
\begin{subfigure}{0.3\textwidth}
  \includegraphics[width=\linewidth]{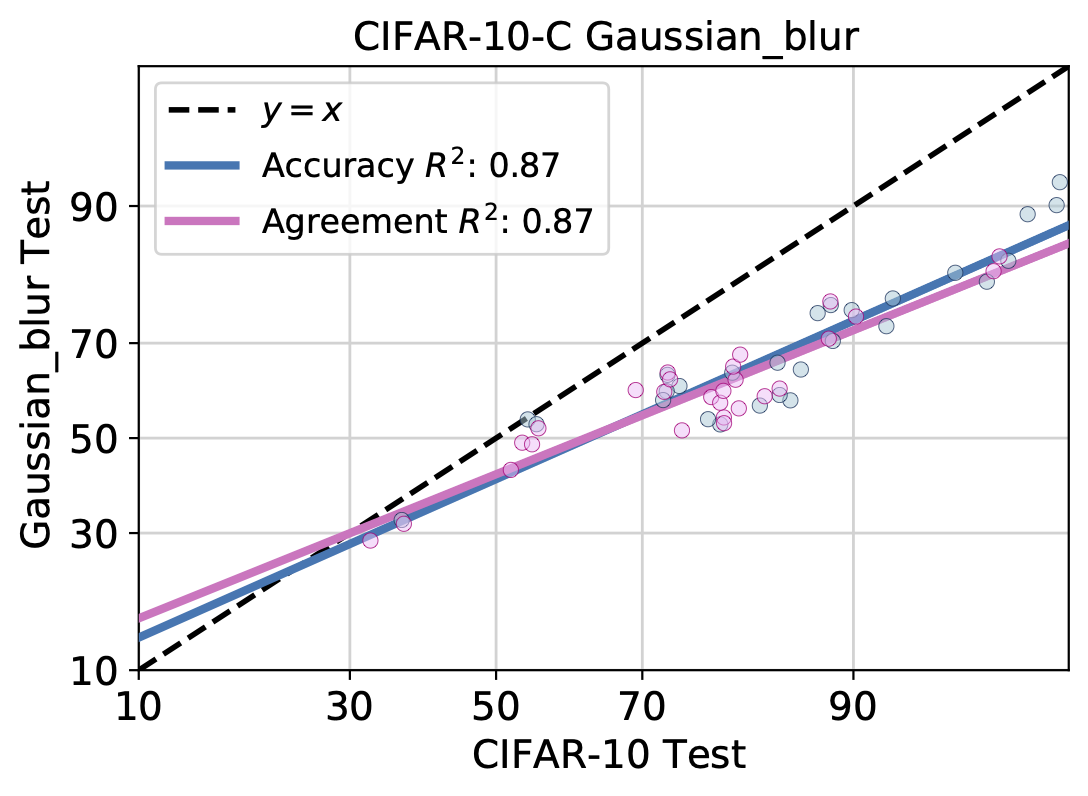}
  \caption{CIFAR10C Gaussian Blur}
\end{subfigure}
\medskip
\begin{subfigure}{0.3\textwidth}
  \includegraphics[width=\linewidth]{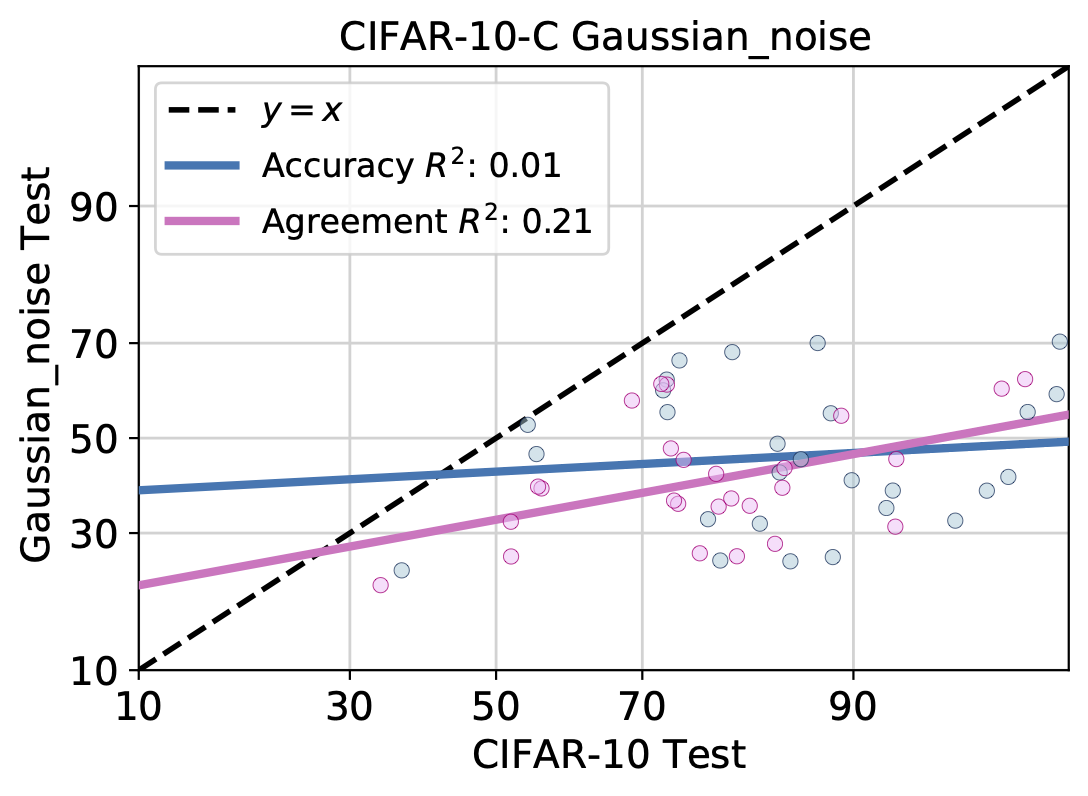}
  \caption{CIFAR10C Gaussian Noise}
\end{subfigure}\hfil %
\begin{subfigure}{0.3\textwidth}
  \includegraphics[width=\linewidth]{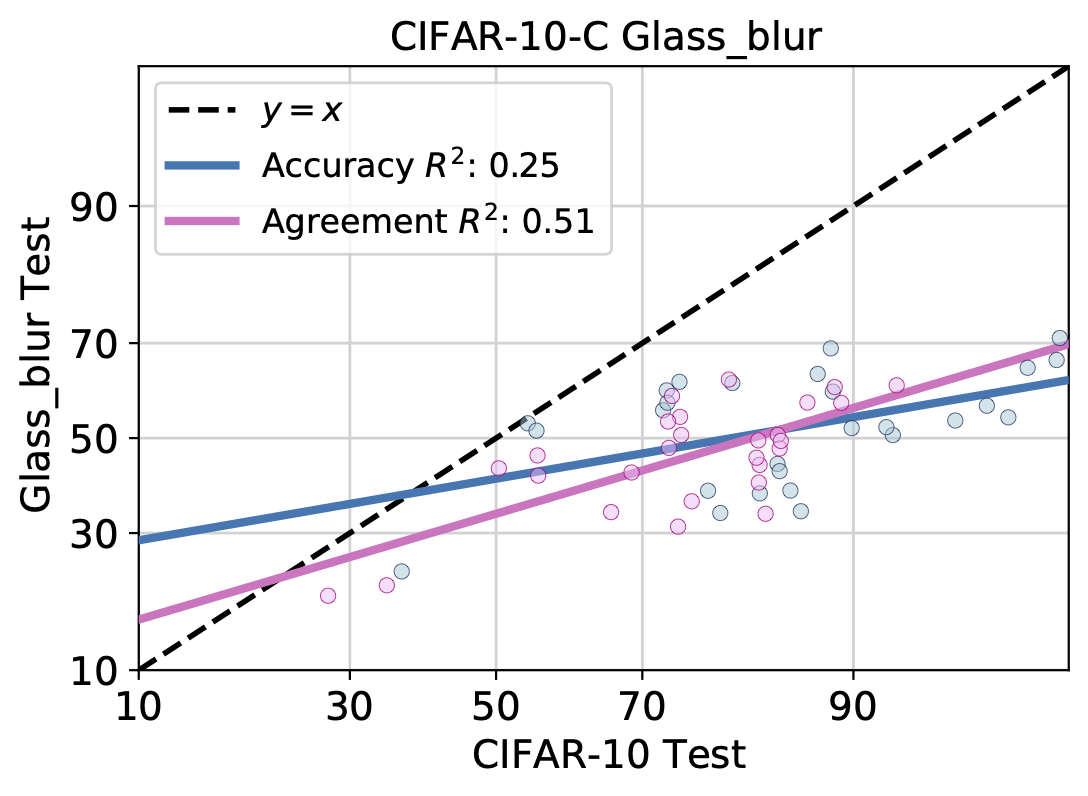}
  \caption{CIFAR10C Glass Blur}
\end{subfigure}\hfil %
\begin{subfigure}{0.3\textwidth}
  \includegraphics[width=\linewidth]{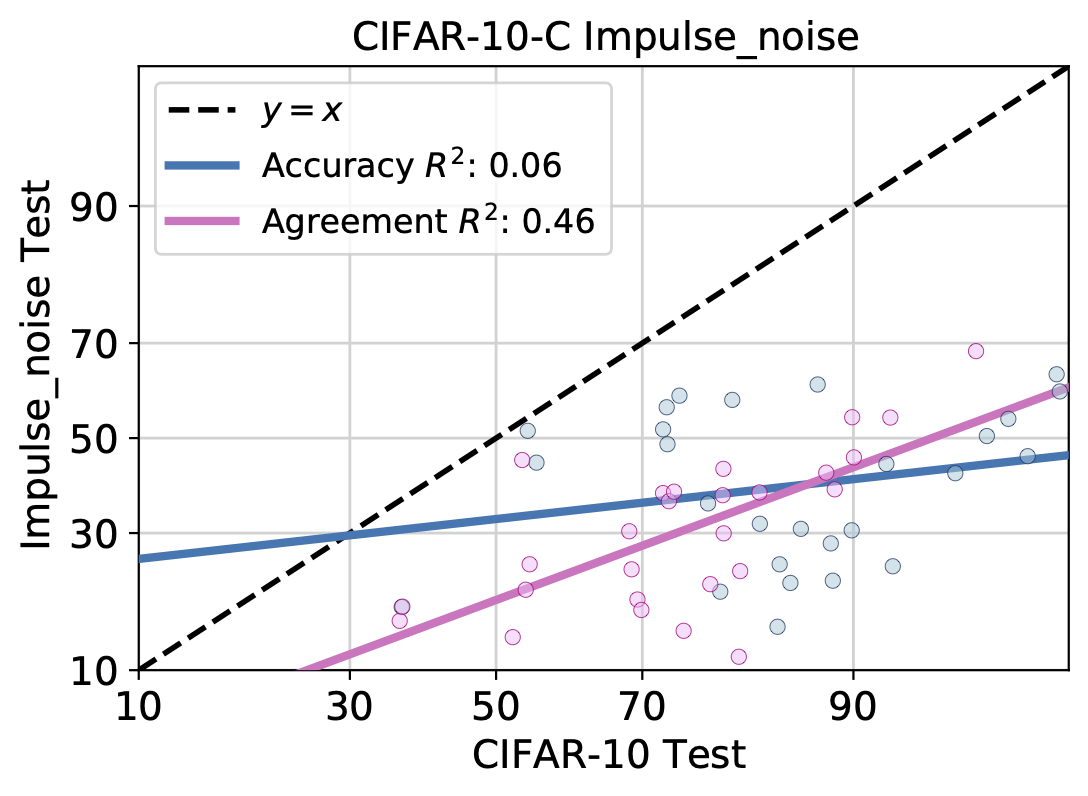}
  \caption{CIFAR10C Impulse Noise}
\end{subfigure}
\medskip
\begin{subfigure}{0.3\textwidth}
  \includegraphics[width=\linewidth]{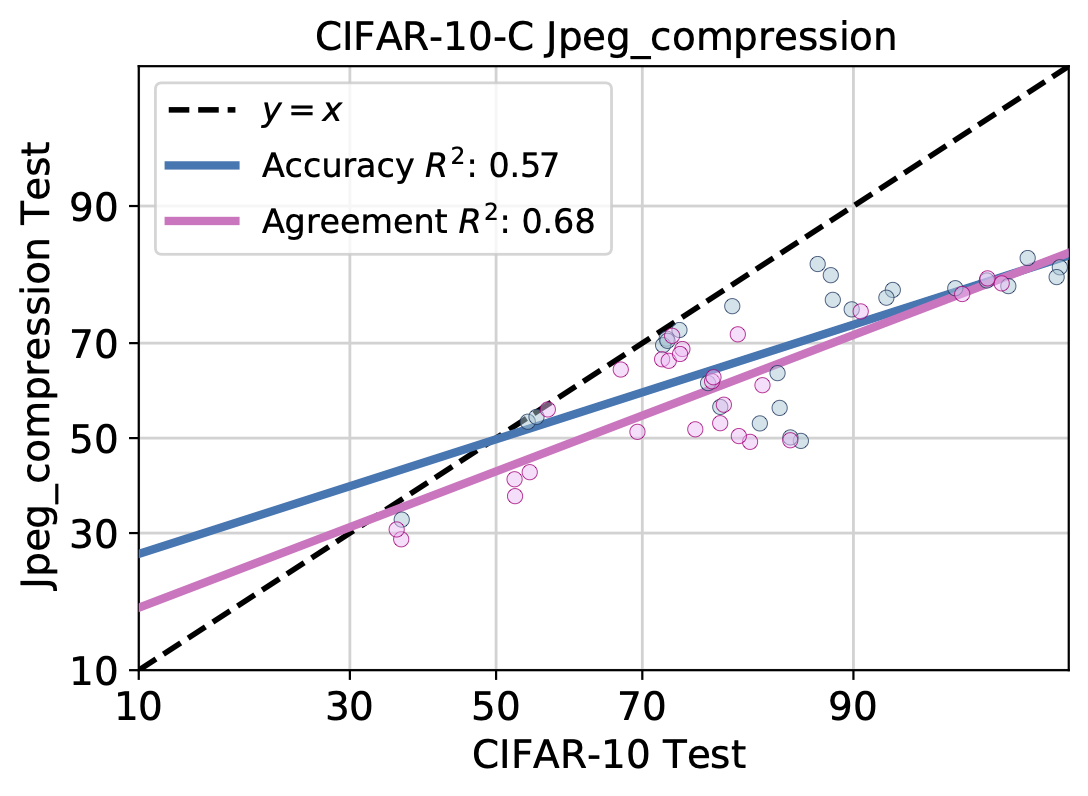}
  \caption{CIFAR10C JPEG Compression}
\end{subfigure}\hfil %
\begin{subfigure}{0.3\textwidth}
  \includegraphics[width=\linewidth]{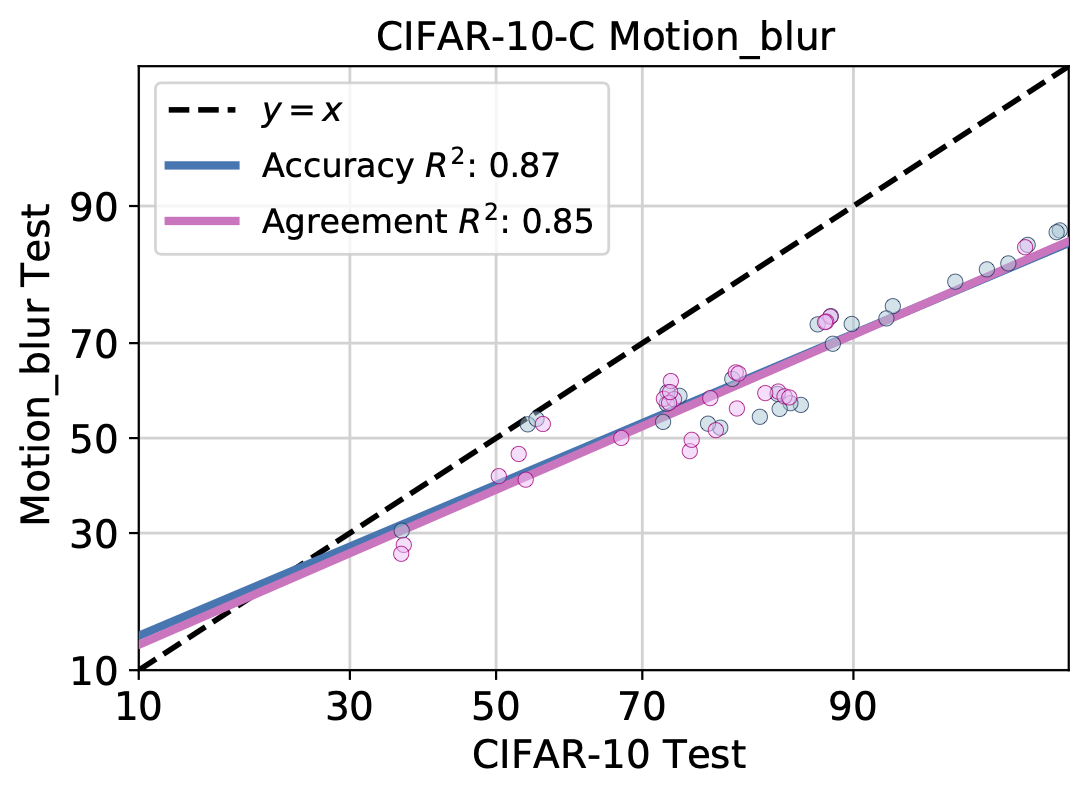}
  \caption{CIFAR10C Motion Blur}
\end{subfigure}\hfil %
\begin{subfigure}{0.3\textwidth}
  \includegraphics[width=\linewidth]{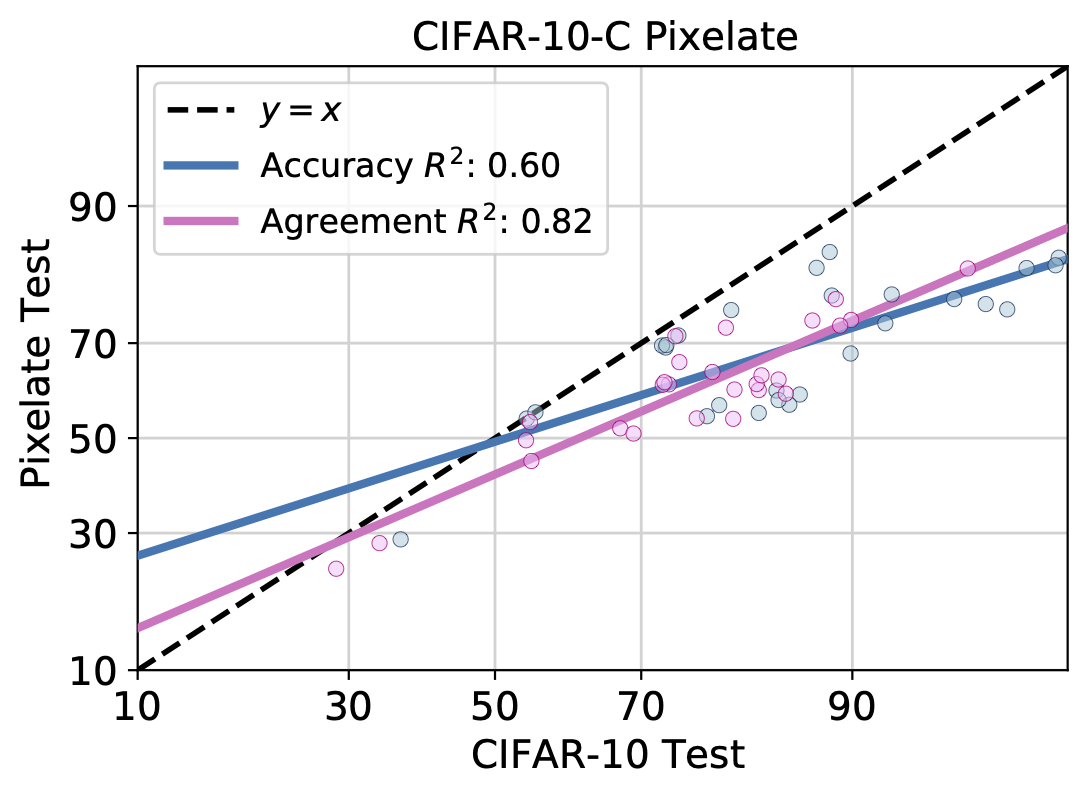}
  \caption{CIFAR10C Pixelate}
\end{subfigure}
\end{figure}

\begin{figure}[H]\ContinuedFloat
\centering
\medskip
\begin{subfigure}{0.3\textwidth}
  \includegraphics[width=\linewidth]{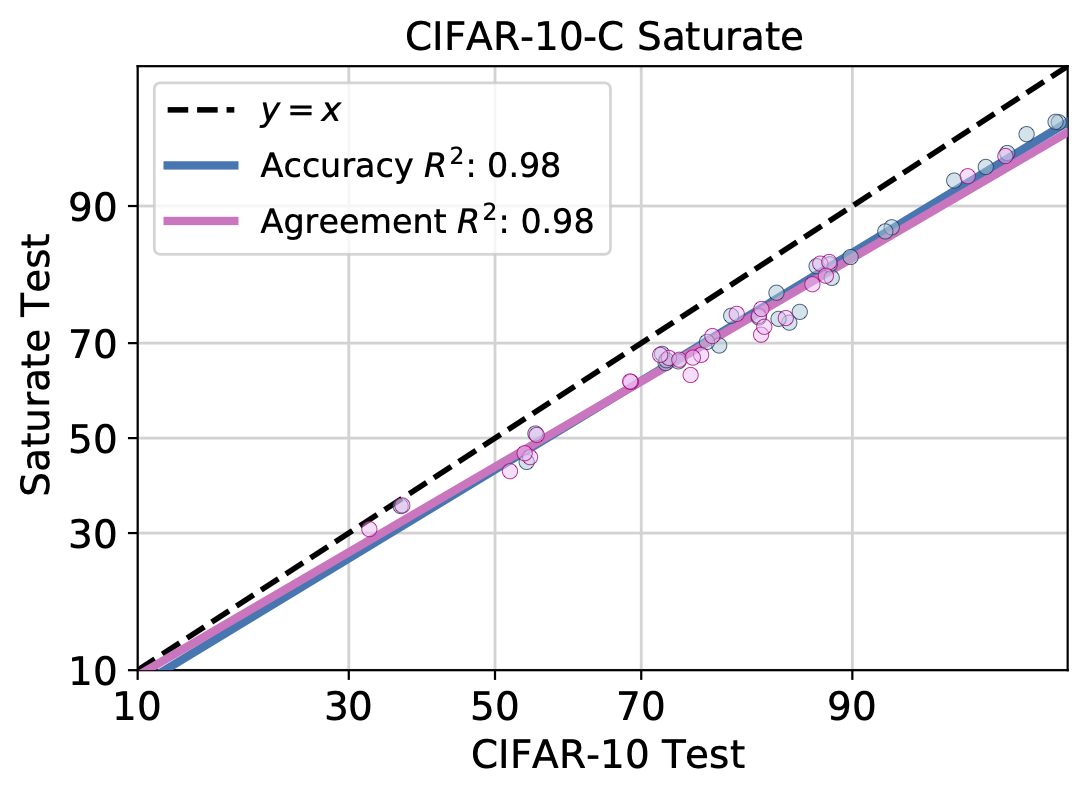}
  \caption{CIFAR10C Saturate}
\end{subfigure}\hfil %
\begin{subfigure}{0.3\textwidth}
  \includegraphics[width=\linewidth]{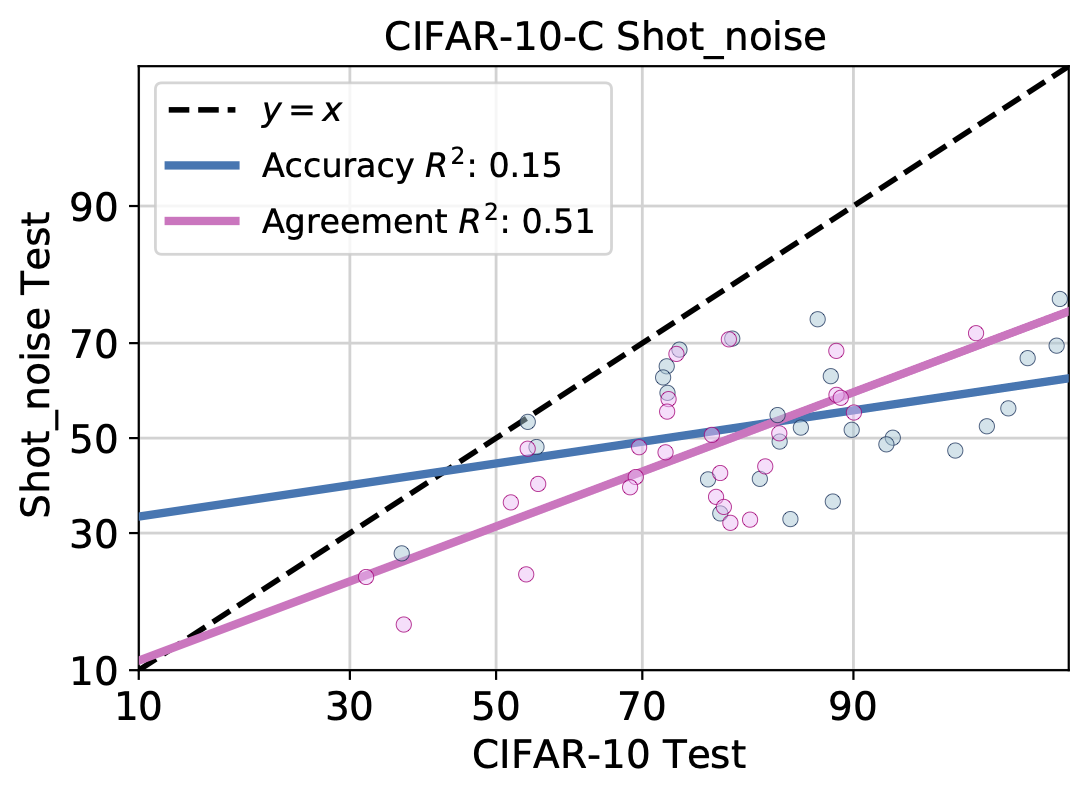}
  \caption{CIFAR10C Shot Noise}
\end{subfigure}\hfil %
\begin{subfigure}{0.3\textwidth}
  \includegraphics[width=\linewidth]{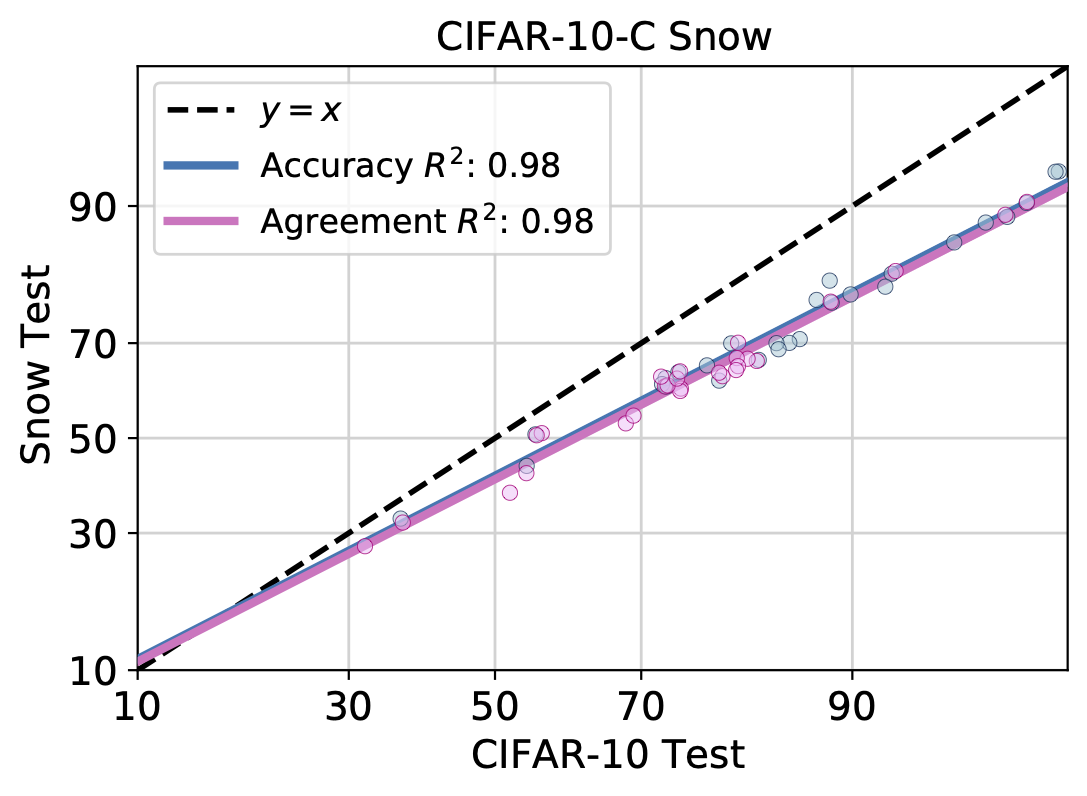}
  \caption{CIFAR10C Snow}
\end{subfigure}
\medskip
\begin{subfigure}{0.3\textwidth}
  \includegraphics[width=\linewidth]{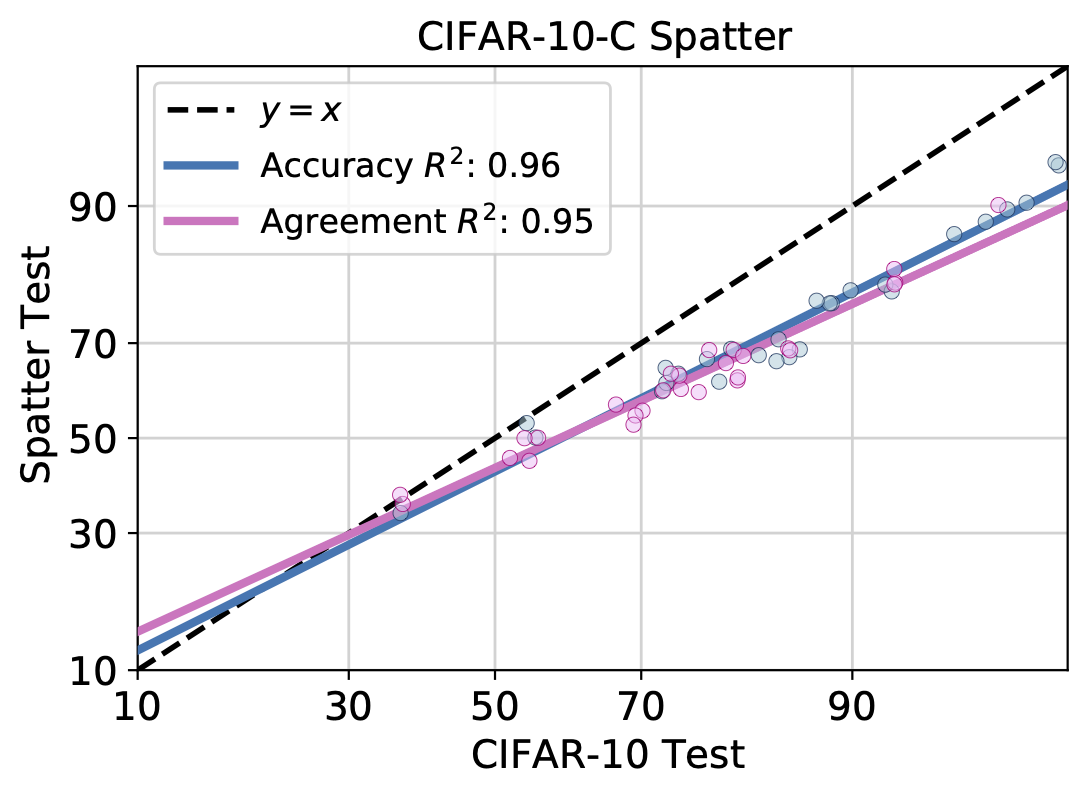}
  \caption{CIFAR10C Spatter}
\end{subfigure}\hfil %
\begin{subfigure}{0.3\textwidth}
  \includegraphics[width=\linewidth]{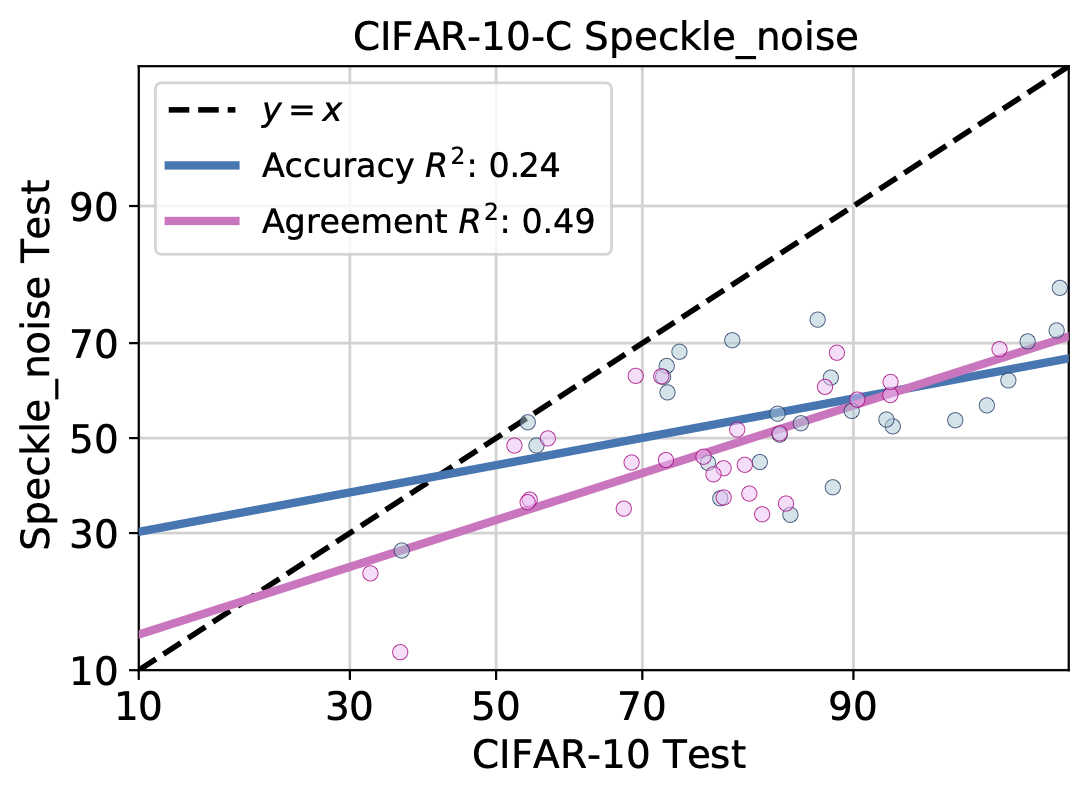}
  \caption{CIFAR10C Speckle Noise}
\end{subfigure}\hfil %
\begin{subfigure}{0.3\textwidth}
  \includegraphics[width=\linewidth]{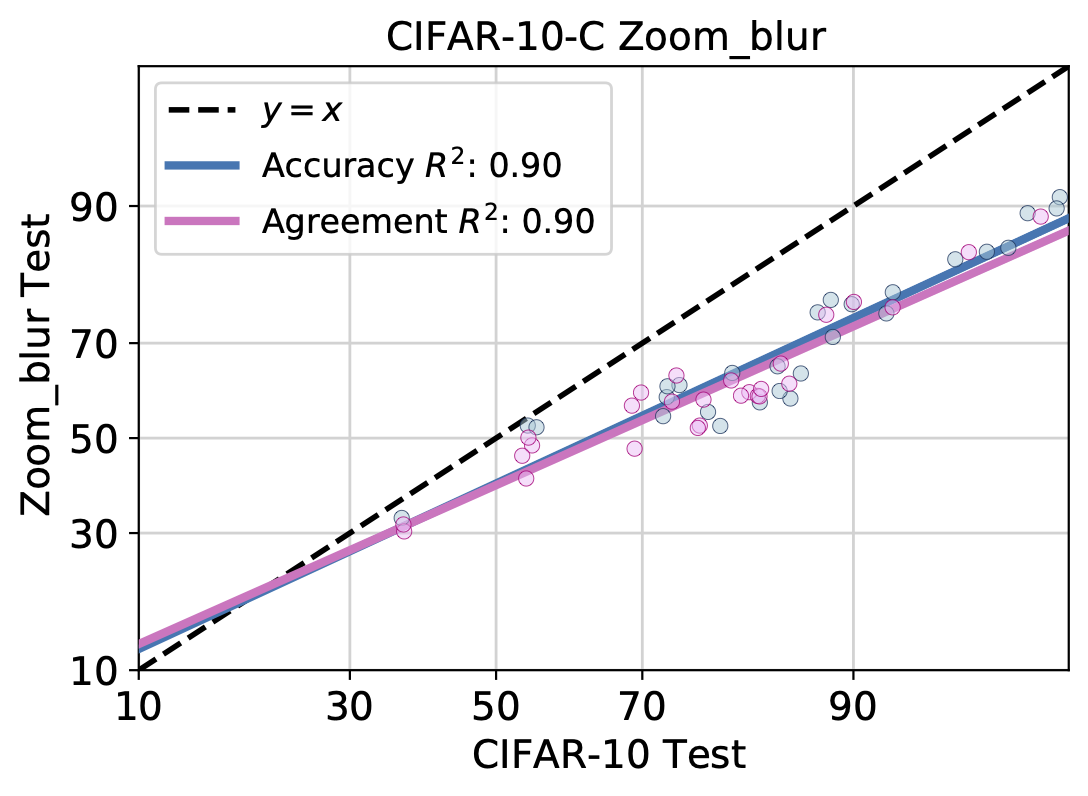}
  \caption{CIFAR10C Zoom Blur}
\end{subfigure}
\medskip
\begin{subfigure}{0.3\textwidth}
  \includegraphics[width=\linewidth]{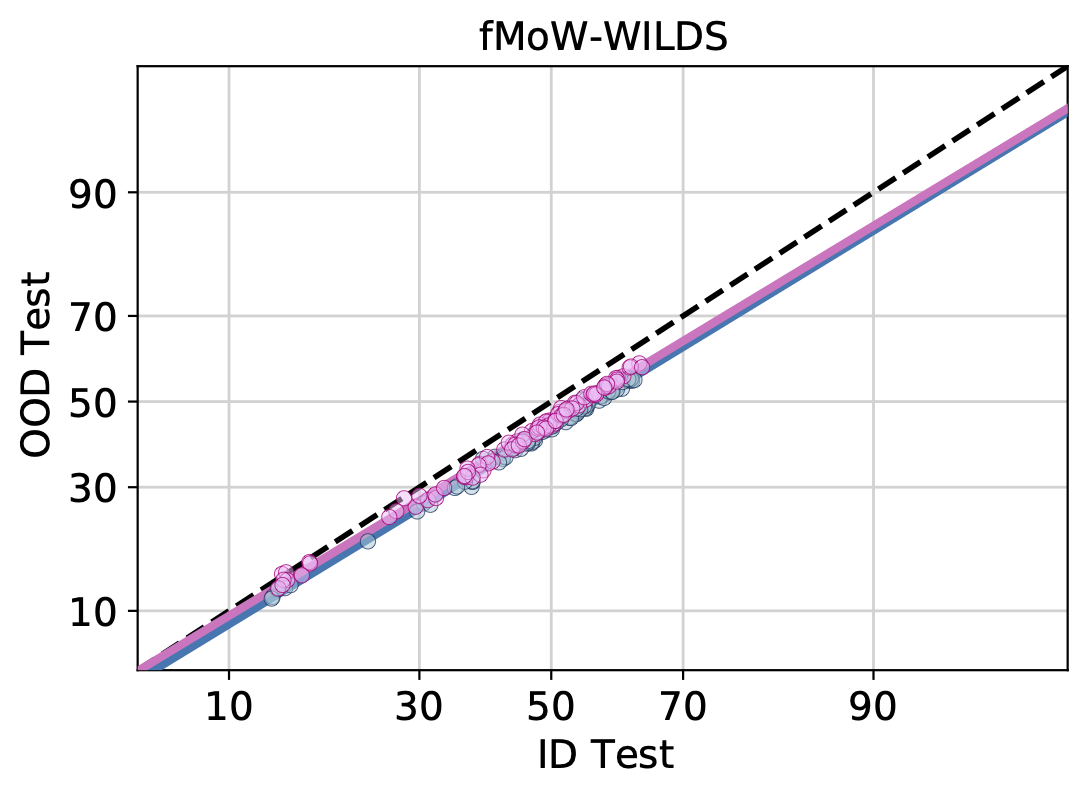}
  \caption{fMoW-\textsc{wilds}}
\end{subfigure}\hfil %
\begin{subfigure}{0.3\textwidth}
  \includegraphics[width=\linewidth]{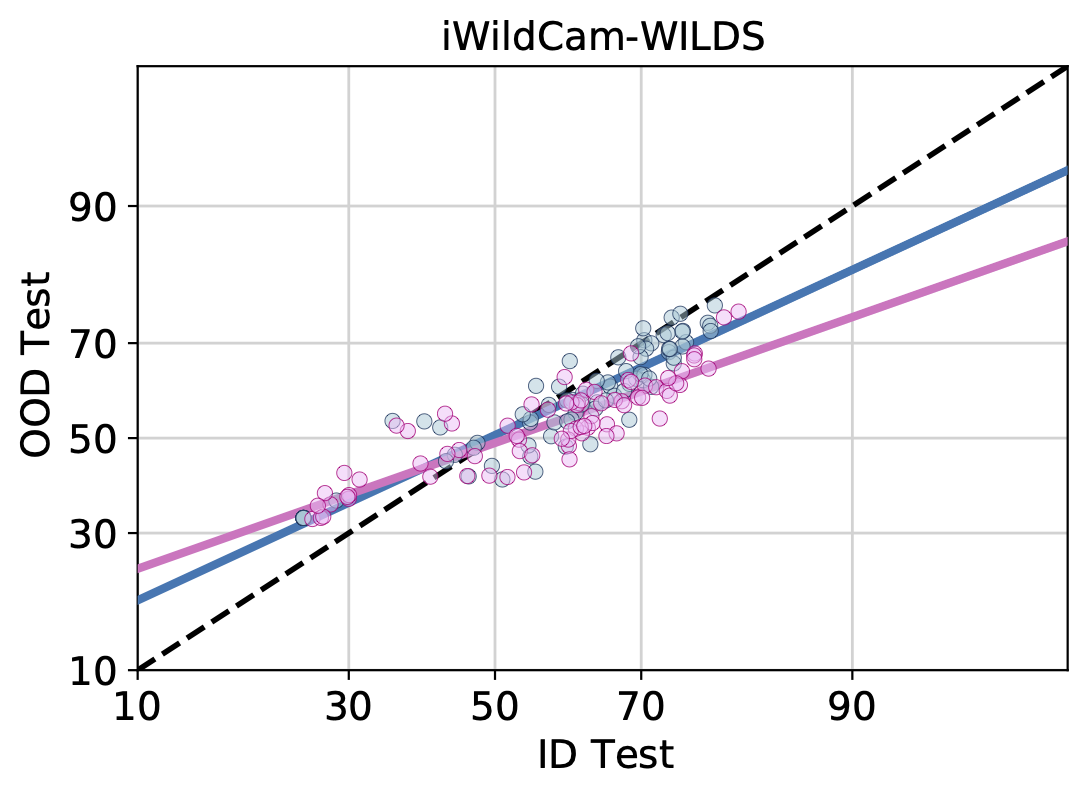}
  \caption{iWildCam-\textsc{wilds}}
\end{subfigure}\hfil %
\begin{subfigure}{0.3\textwidth}
  \includegraphics[width=\linewidth]{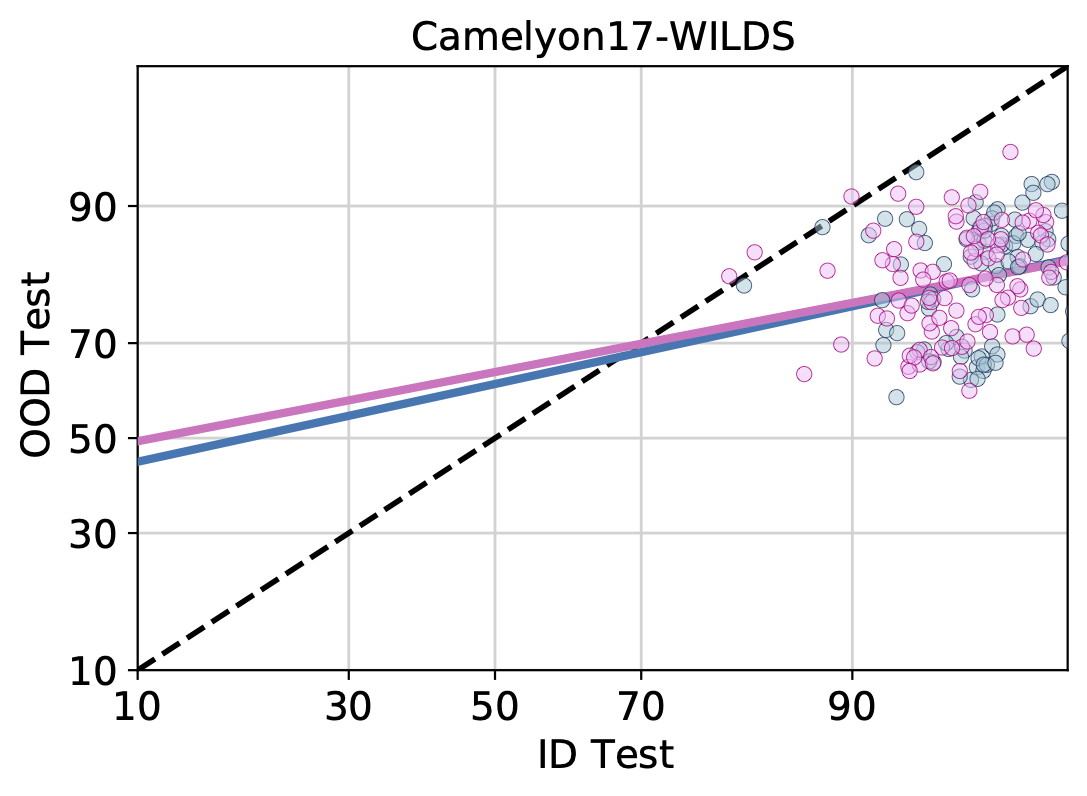}
  \caption{Camelyon17-\textsc{wilds}}
\end{subfigure}
\medskip
\begin{subfigure}{0.3\textwidth}
  \includegraphics[width=\linewidth]{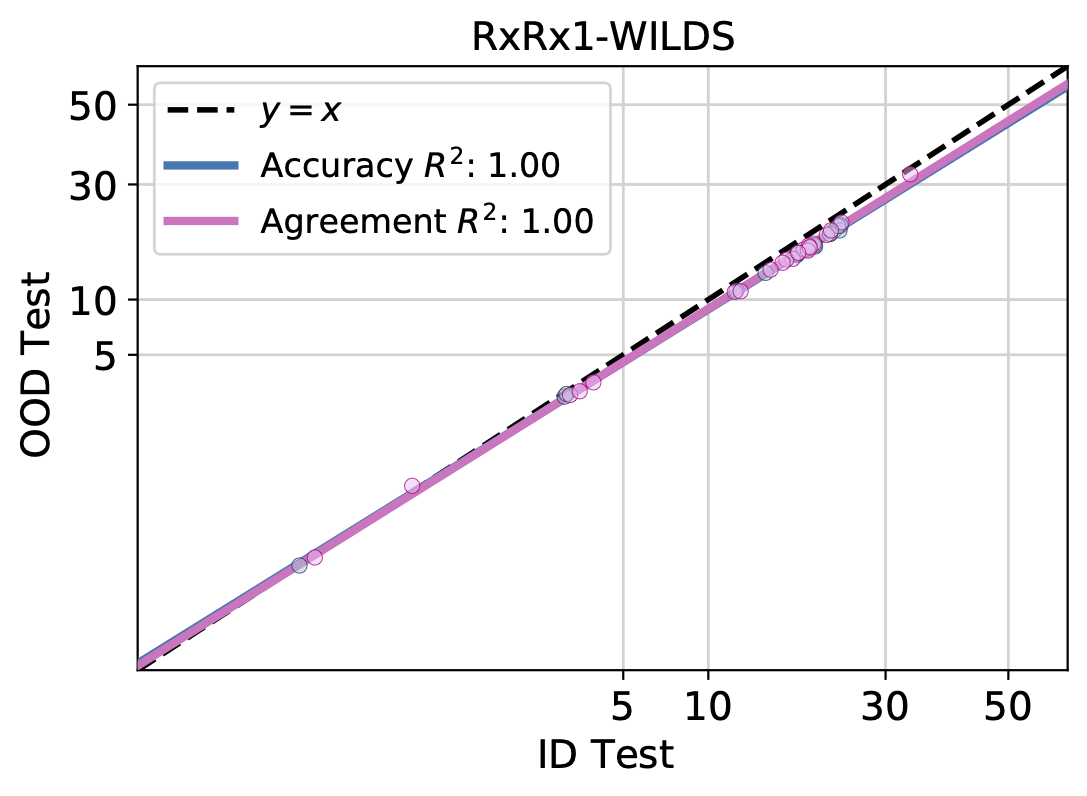}
  \caption{RxRx1-\textsc{wilds}}
  \end{subfigure}
\begin{subfigure}{0.3\textwidth}
  \includegraphics[width=\linewidth]{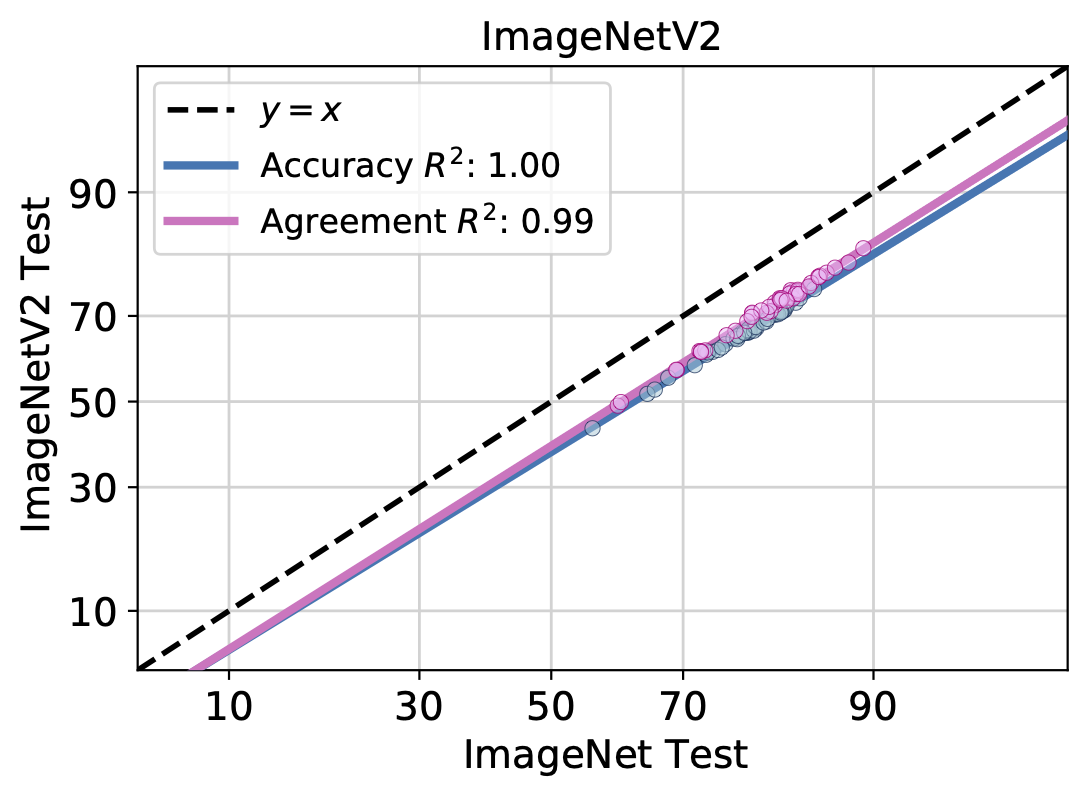}
  \caption{ImageNetV2}
\end{subfigure}
\end{figure}

\subsection{NLP Datasets}
\begin{figure}[ht]
  \centering
  \includegraphics[scale=0.3]{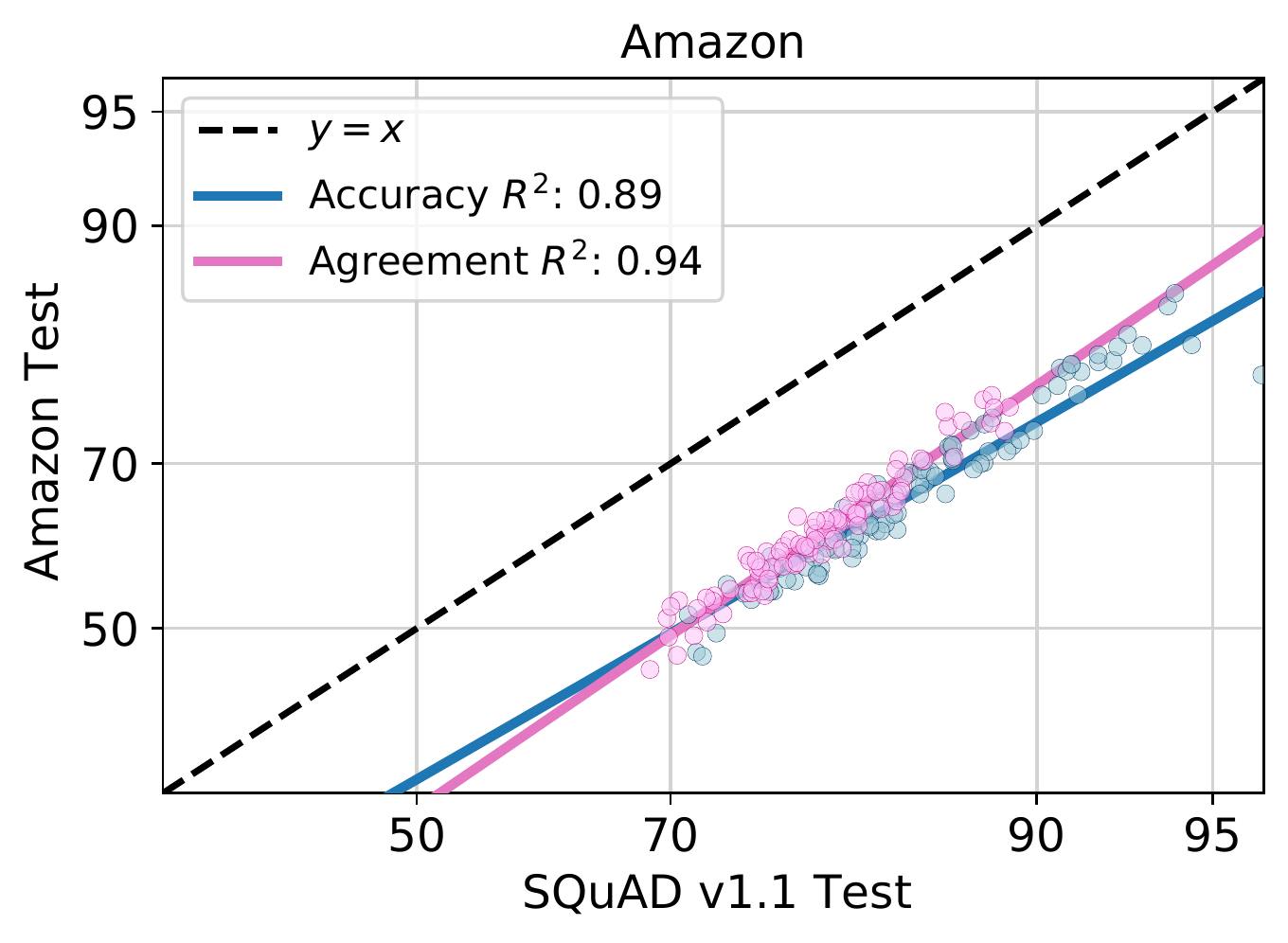}
  \caption{Amazon Q and A}
\end{figure} 

Miller additionally showed in \cite{miller2020effect} that the ID vs OOD F1 score of question-answering models is also often strongly linearly correlated. We plot the ID vs OOD F1 score of the models versus the ID (SQuAD) vs OOD (Amazon) F1-agreement, where F1-agreement is simply the average F1 score where one of the models is treated as the labels or the ground truth. Specifically, 
\begin{align}
    \mathrm{Agr_{F1}(h, h') = \frac{\mathrm{F1}(h, h') + \mathrm{F2}(h', h)}{2}}
\end{align}

Our model collection consists of 99 models from \cite{miller2020effect}, which can also be found in \href{https://worksheets.codalab.org/worksheets/0x787751bd802040ffbea9f6ccbd27175f}{CodaLab}. Note that the linear correlation of agreement is strong with $R^2$ of $0.94$. ALine-D achieves a mean estimation error of 0.028.

\newpage
\subsection{Only Neural Networks}
Here we observe the ID vs OOD accuracy and agreement trend of model families other than neural networks. In the main body of the paper, we illustrated how the agreement-on-the-line phenomenon is specific to neural networks on CIFAR10-Fog, a synthetic shift. Below, we illustrate this further on a data replication shift, CIFAR10.2, and a real-world shift, fMoW-\textsc{wilds}. Note that the slope of the ID vs OOD agreement trend of neural network models is closest to the slope of the ID vs OOD accuracy trend. Excluding neural networks, we observe that the agreement trend of random feature models \cite{randomfeatures} also has a similar slope to that of the accuracy trend for select shifts such as CIFAR-10.2.

\begin{figure}[H]
    \centering
    \includegraphics[scale=0.5]{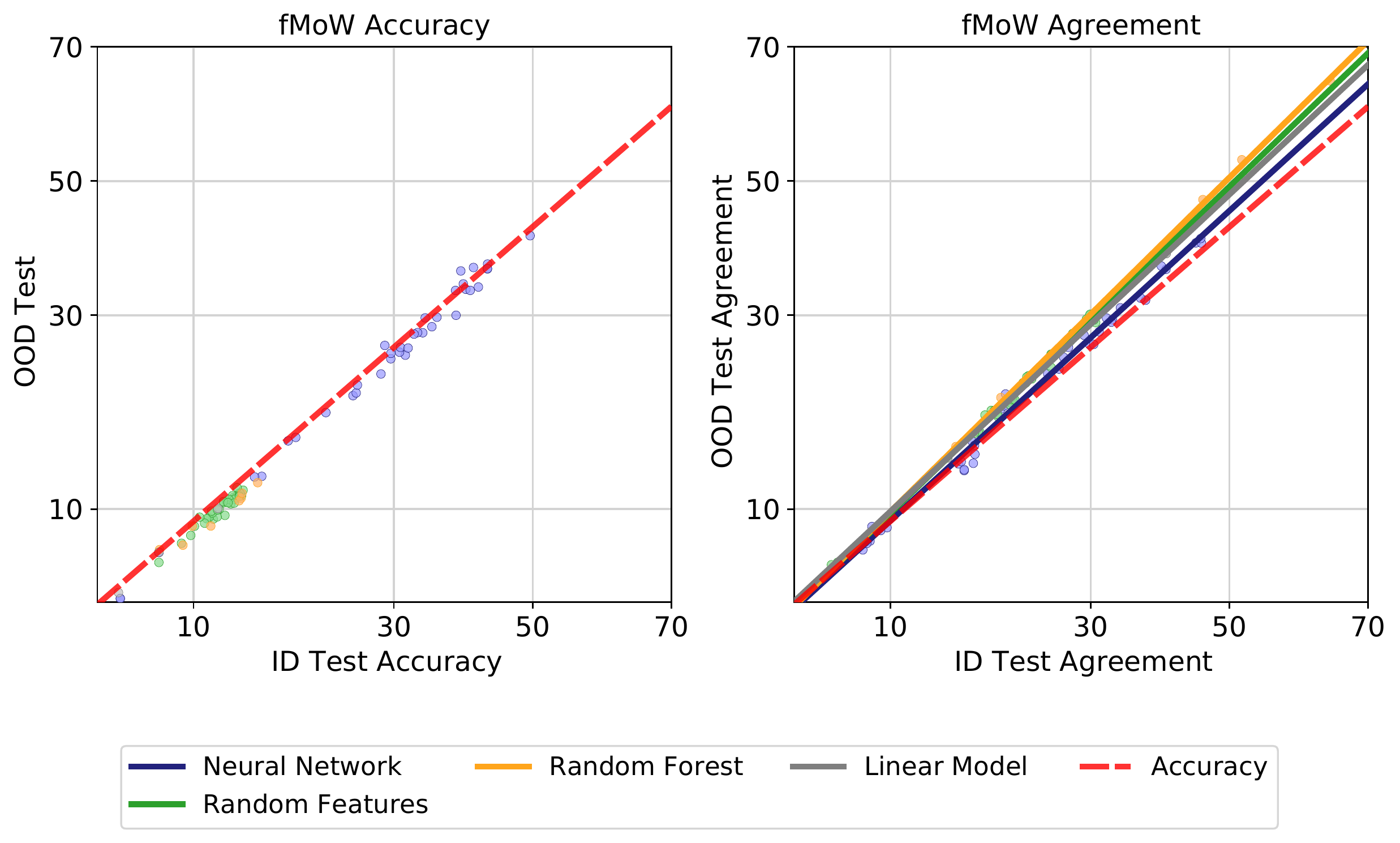}
    \includegraphics[scale=0.5]{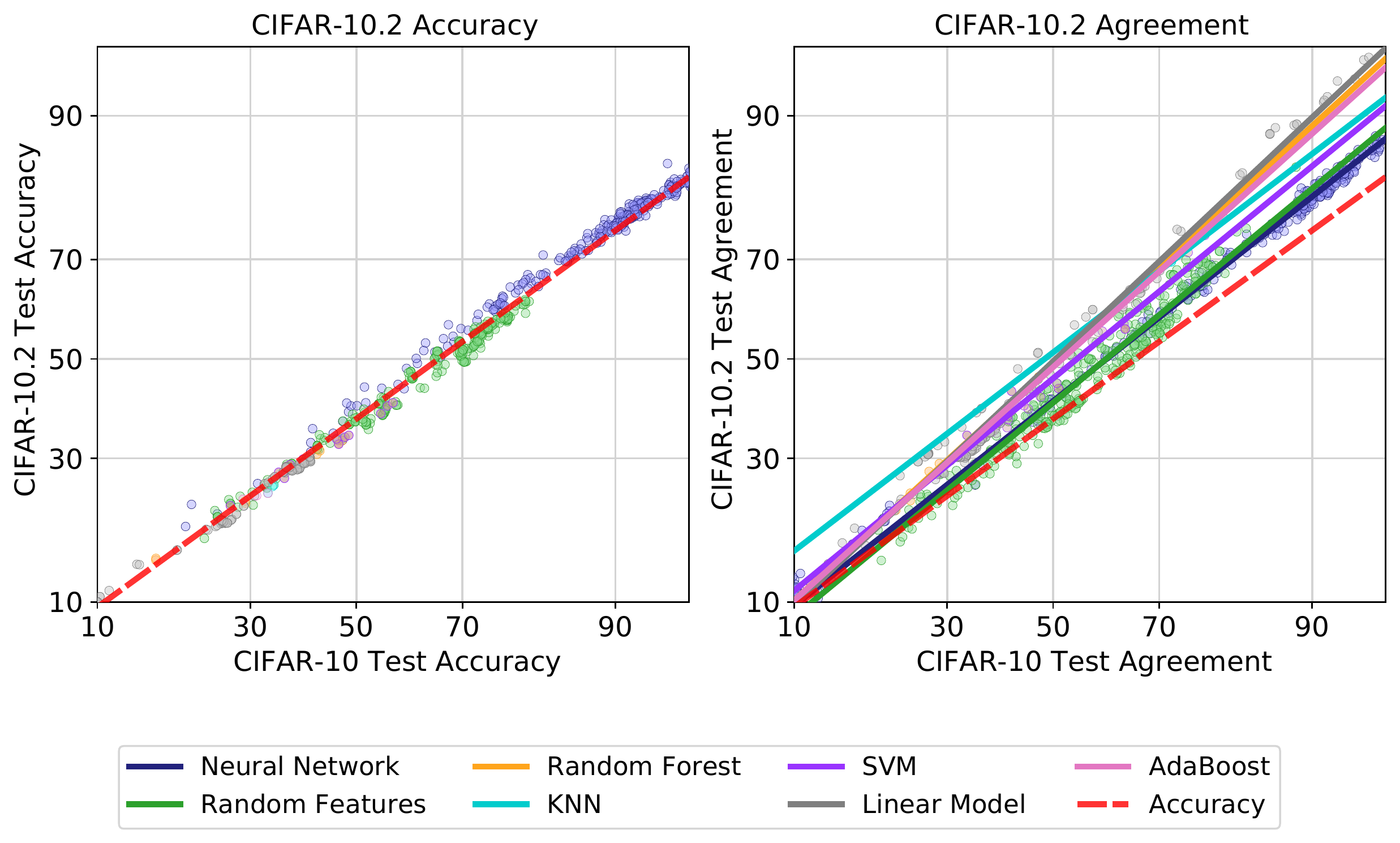}
    \caption{ID vs OOD Accuracy/Agreement of different model families. Given $n$ models, we plot the ID vs OOD accuracy of each model. For agreement, we plot the agreement of $n$ random pairs between models of the same family. The linear trend of ID vs OOD accuracy (in red dashes) is computed using models from all model families.}
\end{figure}
\newpage 
\section{Model Architectures}
\label{app:model}
We list the types of architectures and the corresponding number of models for each testbed. 

\begin{table}[h]
\begin{minipage}[t]{0.45\linewidth}\centering
\caption{CIFAR-10 Testbed \protect{\cite{Miller1}}: 467 models total, out of which 29 are pretrained on ImageNet.}
    \begin{tabular}{cc} \\
         \hline Architecture & Number of models  \\
         \cmidrule(lr){1-1}
         \cmidrule(lr){2-2}
         DenseNet121 \cite{densenet} & 21 \\
         DenseNet169 \cite{densenet} & 8 \\
         EfficientNetB0 \cite{efficientnet}& 13 \\ 
         ResNet18 \cite{resnet}& 13 \\ 
         ResNet50 \cite{resnet}& 18 \\
         ResNet101 \cite{resnet}& 7 \\ 
         PreActResNet18 \cite{preact_resnet}& 63 \\ 
         PreActResNet34 \cite{preact_resnet}& 9 \\
         PreActResNet50 \cite{preact_resnet}& 11 \\ 
         PreActResNet101 \cite{preact_resnet}& 4 \\
         ResNeXT $2\times64$d \cite{resnext}& 12 \\ 
         ResNeXT $32\times4$d \cite{resnext}& 8 \\
         ResNeXT $4\times64$d \cite{resnext}& 1 \\
         RegNet X200 \cite{regnet} & 11 \\ 
         RegNet X400 \cite{regnet} & 13 \\ 
         RegNet Y400 \cite{regnet} & 5 \\
         VGG11 \cite{vgg}& 16 \\
         VGG13 \cite{vgg}& 13 \\
         VGG16 \cite{vgg}& 13 \\ 
         VGG19 \cite{vgg}& 12 \\
         ShuffleNetV2 \cite{shufflenetv2}& 56 \\ 
         ShuffleNetG2 \cite{shufflenet}& 13 \\
         ShuffleNetG3 \cite{shufflenet}& 8 \\
         AlexNet \cite{alexnet}& 2 \\
         MobileNet \cite{mobilenet}& 12 \\
         MobileNetV2 \cite{mobilenet}& 13 \\
         PNASNet-A \cite{pnasnet}& 13 \\ 
         PNASNet-B \cite{pnasnet}& 13 \\
         PNASNet-5-Large \cite{pnasnet} & 3 \\
         SqueezeNet  \cite{squeezenet}& 3 \\
         SENet18 \cite{senet} & 13 \\ 
         GoogLeNet \cite{googlenet} & 20 \\
         DPN26 \cite{dpn} & 8 \\
         DPN92 \cite{dpn} & 2 \\ 
         Myrtlenet \href{https://github.com/davidcpage/cifar10-fast}{[Repo]} & 1 \\
         Xception \cite{xception} & 3 \\
         \hline
    \end{tabular}
    \label{tab:cifar10_models}
\end{minipage}
\hfil
\begin{minipage}[t]{0.45\linewidth}\centering
    \caption{ImageNet Testbed: 49 models total from the timm package \protect{\cite{rw2019timm}} and Torchvision \href{https://pytorch.org/vision/stable/models.html}{[link]}.}
    \begin{tabular}{cc} \\
         \hline Architecture & Number of models  \\
         \cmidrule(lr){1-1}
         \cmidrule(lr){2-2} 
        Adversarial Inception v3 \cite{adversarial_inception}& 1 \\ 
        AlexNet \cite{alexnet}& 1 \\
        BEiT \cite{bao2021beit}& 1 \\
        BoTNet \cite{botnet}& 1 \\
        CaiT \cite{cait}& 1 \\
        CoaT \cite{coat}& 2 \\
        ConViT \cite{convit}& 3 \\
        ConvNeXT \cite{convnext}& 1 \\
        CrossViT \cite{crossvit}& 9 \\
        DenseNet \cite{densenet} & 3 \\
        DLA \cite{dla} & 10 \\
        EfficientNet \cite{efficientnet} & 1 \\
        HaloNet \cite{halonet} & 1 \\
        NFNet \cite{nfnet} & 1 \\
        ResNet \cite{resnet}& 10 \\
        ResNeXT \cite{resnext}& 1 \\
        Inception v3 \cite{inceptionv3} & 1 \\
        VGG \cite{vgg} & 1 \\
        \hline
    \end{tabular}
    \vspace{3mm}
    \caption{FMoW Testbed \protect{\cite{Miller1}}: 161 models total, out of which 37 are pretrained on ImageNet and 2 are CLIP pretrained models.}
    \begin{tabular}{cc} 
         \hline Architecture & Number of models  \\
         \cmidrule(lr){1-1}
         \cmidrule(lr){2-2} 
        ResNet \cite{resnet} & 40 \\
        ResNeXT \cite{resnext} & 18 \\
        AlexNet \cite{alexnet} & 11 \\
        DPN68 \cite{dpn} & 15 \\
        DenseNet121 \cite{densenet} & 11 \\
        GoogLeNet \cite{googlenet} & 8 \\
        Xception \cite{xception} & 11 \\
        ShuffleNet \cite{shufflenet} & 10 \\
        MobileNetV2 \cite{mobilenet} & 8 \\
        VGG \cite{vgg} & 15 \\ 
        PNASNet \cite{pnasnet} & 2 \\
        Squeezenet \cite{squeezenet} & 11 \\ 
        ViT \cite{dosovitskiy2020vit} & 1 \\ 
        \hline
    \end{tabular}
    \label{tab:imagenet_models}
\end{minipage}
\end{table}
\newpage 
\begin{table}[h]
\begin{minipage}[t]{0.45\linewidth}\centering
\caption{iWildCam-\textsc{wilds} Testbed: 157 models total, out of which 81 are pretrained.}
    \begin{tabular}{cc} \\
         \hline Architecture & Number of models  \\
         \cmidrule(lr){1-1}
         \cmidrule(lr){2-2}
         AlexNet \cite{alexnet} & 30 \\ 
         ShuffleNetV2 \cite{shufflenetv2} & 30 \\ 
         ResNeXT \cite{resnext} & 5 \\ 
         ResNet \cite{resnet} & 38 \\
         VGG \cite{vgg} & 5 \\
         SqueezeNet \cite{squeezenet} & 5 \\
         MobileNetV2 \cite{mobilenet} & 30 \\
         PNASNet \cite{pnasnet} & 4 \\
         Xception \cite{xception} & 5 \\ 
         DenseNet \cite{densenet} & 5 \\ 
         \hline
    \end{tabular}
\end{minipage}
\hfil
\begin{minipage}[t]{0.45\linewidth}\centering
\caption{Camelyon17-\textsc{wilds} Testbed: 269 models total out of which 100 are pretrained on ImageNet.}
    \begin{tabular}{cc}
         \hline Architecture & Number of models  \\
         \cmidrule(lr){1-1}
         \cmidrule(lr){2-2} 
        ResNet \cite{resnet} & 29 \\
        SqueezeNet \cite{squeezenet} & 27 \\
        ShuffleNetV2 \cite{shufflenetv2} & 29 \\
        VGG \cite{vgg} & 28 \\
        AlexNet \cite{alexnet} & 29 \\
        MobileNetV2 \cite{mobilenet} & 29 \\
        ResNeXT \cite{resnext} & 27 \\
        DenseNet \cite{densenet} & 28 \\
        Xception \cite{xception} & 28 \\
        PNASNet \cite{pnasnet} & 15 \\ 
        \hline
    \end{tabular}
\end{minipage}

\begin{minipage}[t]{0.45\linewidth}\centering
\caption{RxRx1-\textsc{wilds} Testbed: 36 models total out of which 16 are pretrained on ImageNet.}
    \begin{tabular}{cc}
         \hline Architecture & Number of models  \\
         \cmidrule(lr){1-1}
         \cmidrule(lr){2-2} 
        ResNet18 \cite{resnet} & 9 \\
        ResNet50 \cite{resnet} & 21 \\
        DenseNet121 \cite{densenet} & 6 \\
        \hline
    \end{tabular}
\end{minipage}
\end{table}

\newpage
\section{Section 5: Experimental Details}
\label{app:experiment}
\subsection{Details for Experiment 5.2: Correlation analysis}
We replicate the correlation analysis experiment in Table 1 of \citet{yu2022predicting} to compare the prediction performance of ALine-D versus ProjNorm. Essentially, we want to see how strong the linear correlation is between the estimate of OOD accuracy versus the true OOD accuracy by looking at the coefficients of determination $R^2$ and rank correlations $\rho$ of the fit. We use the GitHub repository of ProjNorm \cite{yu2022predicting} found at \url{https://github.com/yaodongyu/ProjNorm} to replicate their correlation analysis experiment found in their Table 1. Using their repository, we train a base ResNet18 model for 20 epochs with their default hyperparameters

\begin{itemize}
\item{Batch Size:} 128
\item{Learning Rate:} 0.001
\item{Weight Decay:} 0
\item{Optimizer:} SGD with Momentum 0.9
\item {Pretrained:} True
\end{itemize} 
 using cosine learning rate decay \cite{cosinelr}. The repository uses the default implementation of ResNet18 by \href{https://pytorch.org/vision/stable/models.html}{torchvision}. The CIFAR-10 images are resized to be $224 \times 224$, then normalized. To compute ProjNorm, we use the repository to train a reference ResNet18 model using the pseudolabels of the base model for 500 iterations with the same hyperparameters as the base model. We use the ALine-D algorithm to predict the OOD accuracy of the base model. ALine-D requires a model set $\mc{H}$ with at least 3 models so that the linear system of equations (Line 9 in Algorithm 1, Equation 6 in main body) has a unique solution. We use the 29 pretrained models from the CIFAR10 testbed as the other models in the model set.  
\subsection{Details for Experiment 5.3: Performance along a training trajectory}
We train a ResNet18 model from scratch (no pretrained weights) on CIFAR10 for 150 epochs using the following hyperparameters:
\begin{itemize}
\item{Batch Size:} 100 
\item{Learning Rate:} 0.1 
\item{Weight Decay:} $10^{-5}$
\item{Optimizer:} SGD with Momentum 0.9
\item {Pretrained:} False
\end{itemize} 

We decay the learning rate by 0.1 at epoch 75 and epoch 113. We use simple data augmentation consisting of RandomCrop with padding$=4$, RandomHorizontalFlip, and Normalization. We modify ResNet18 to take in $32 \times 32$ input images. 

During training, we save the model weights every 5 training epochs. Given this collection of models, we estimate the CIFAR-10.1 accuracy of each one of these models using the ALine-D procedure. 

\subsection{Hardware}
All experiments were conducted using GeForce GTX 1080 and 2080 graphics cards. 
\newpage
\section{The relationship between agreement and accuracy}
\label{app:calibration}
This work is related to~\citet{jiang2022assessing}, which shows that the agreement between two models of the \textit{same architecture} trained with \textit{different random seeds} is approximately equal to the average ID test accuracy of models if the ensemble consisting of the models is \textit{well-calibrated}. They call this equality between accuracy and agreement Generalization Disagreement Equality (GDE).

Let us ignore the probit transform for a moment and assume the linear correlation between ID vs OOD accuracy and agreement are strong without it. In this simplified scenario, agreement-on-the-line implies that, for datasets where both agreement and accuracy are strongly linearly correlated, if ID agreement of a pair of models is equal to their average ID accuracy, then their OOD agreement is equal to their OOD accuracy. Formally, when a shift satisfies accuracy-on-the-line, we know by agreement-on-the-line that for any two models trained on ID samples $h, h' \in \mc{H}$, the following equations are approximately satisfied (ignoring probit transform)
\begin{align}
     &{\small\frac{\msf{{Acc}_{OOD}}(h) + \msf{{Acc}_{OOD}}(h')}{2} = a \cdot \frac{\msf{{Acc}_{ID}}(h) + \msf{{Acc}_{ID}}(h')}{2} + b }\\
     &{\text{and} \quad \small\msf{{Agr}_{OOD}}(h, h') = a \cdot \msf{{Agr}_{ID}}(h, h') + b } \Longrightarrow \\ 
     &{\small\underbrace{\frac{\msf{{Acc}_{OOD}}(h) + \msf{{Acc}_{OOD}}(h')}{2} - \msf{{Agr}_{OOD}}(h, h')}_{\text{OOD Gap}} =  a \cdot \underbrace{\left(\frac{\msf{{Acc}_{ID}}(h) + \msf{{Acc}_{ID}}(h')}{2} - \msf{{Agr}_{ID}}(h, h') \right)}_{\text{ID Gap}}}
\end{align}
for some slope $a$ and intercept $b$. Thus, if the ``ID gap'' between accuracy and agreement is 0, then the ``OOD gap'' is also 0. This may suggest something about calibration on shifts where accuracy-on-the-line holds: if the ensemble of a pair of models is well calibrated ID, then by agreement-on-the-line GDE also holds OOD.

\begin{figure}[ht]
    \centering
    \includegraphics[scale=0.35]{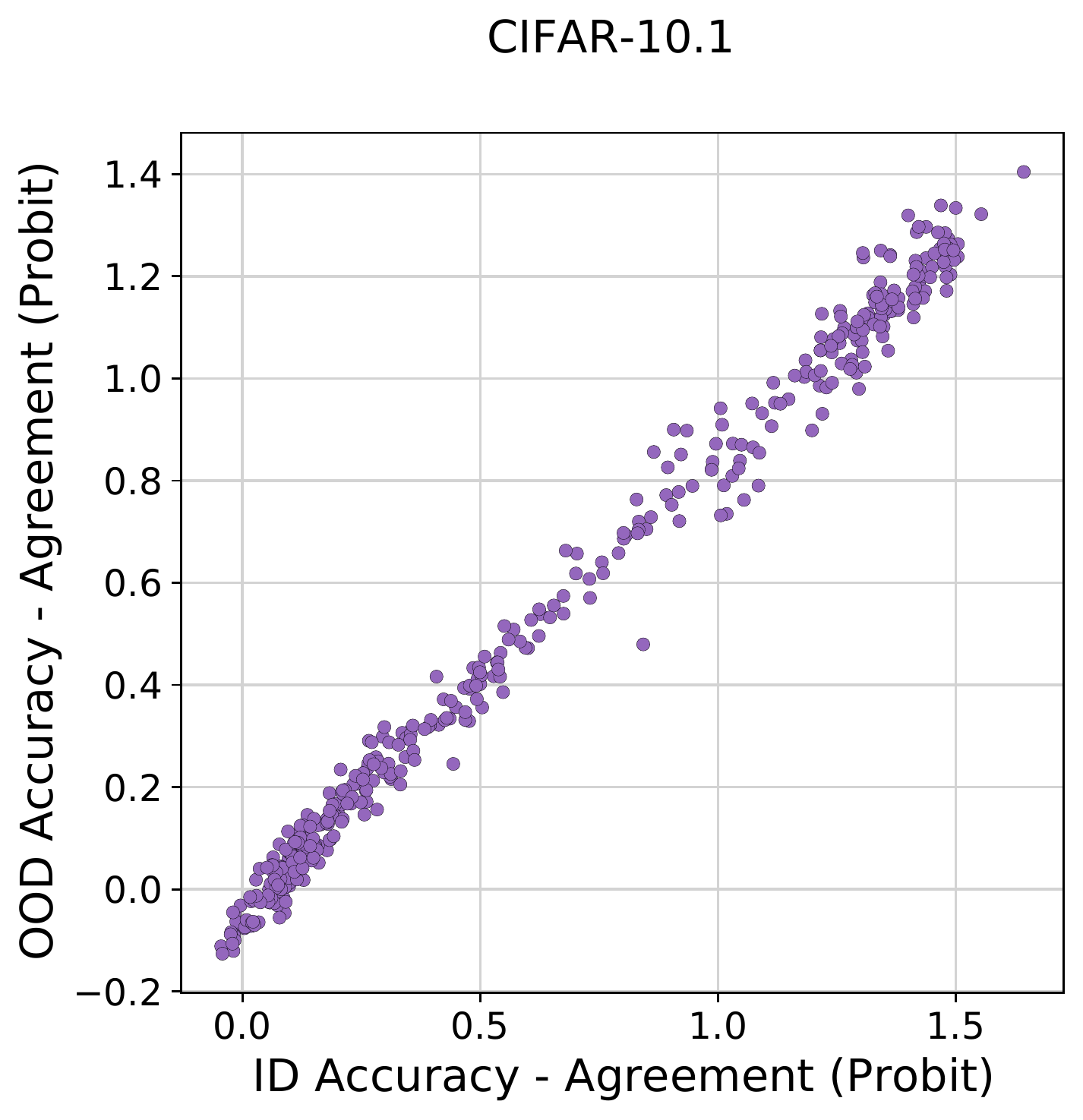}
    \caption{We plot the ID (CIFAR-10 Test) vs OOD (CIFAR 10.1) Test gap for 468 pairs of models randomly sampled from the CIFAR10 testbed. We observe that due to agreement-on-the-line, we observe a strict linear correlation (Eq. 9). However, not all pairs satisfy GDE (ID or OOD gap is not close to 0).}
    \label{fig:no_gde}
\end{figure}

However, agreement-on-the-line goes beyond these results in two ways: (i) agreement between models with \emph{different architectures} and (ii) agreement between different checkpoints on the \emph{same training run} is also on the ID vs OOD agreement line. \citet{jiang2022assessing} does not guarantee GDE holds for these cases. As can be seen in Figure \ref{fig:no_gde}, for most pairs of models, the ID and OOD gaps between probit scaled accuracy and agreement are not equal to 0 i.e. GDE does not occur ID or OOD. Indeed, understanding why agreement-on-the-line holds requires going beyond the theoretical conditions presented in the prior work~\citep{jiang2022assessing} which do not hold for this expanded set of models. 

\begin{figure}[ht]
    \centering
    \includegraphics[scale=0.35]{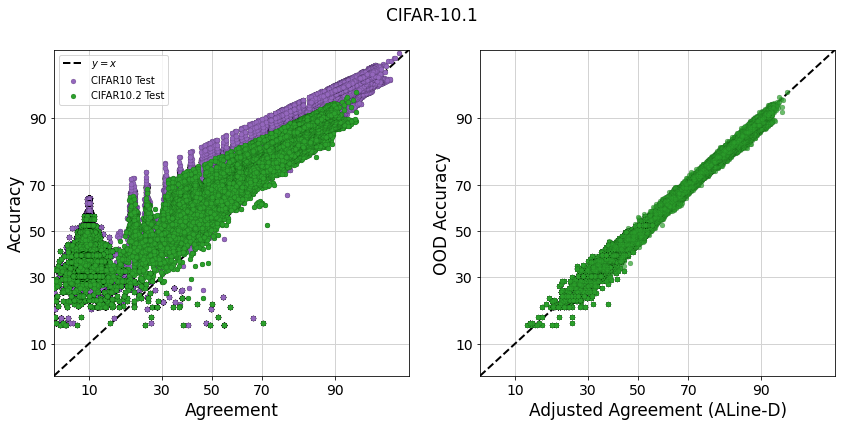}
    \caption{We plot the agreement vs average accuracy (after probit scaling) for every pair of 467 models trained on CIFAR10. We observe that accuracy is not exactly equal to agreement, but the agreement of better performing pairs are closer to its accuracy. However, after the adjustment provided in the left hand side of Equation 6, the agreement matches OOD accuracy.}
    \label{fig:two_no_gde}
\end{figure}

\begin{figure}[ht]
    \centering
    \includegraphics[scale=0.35]{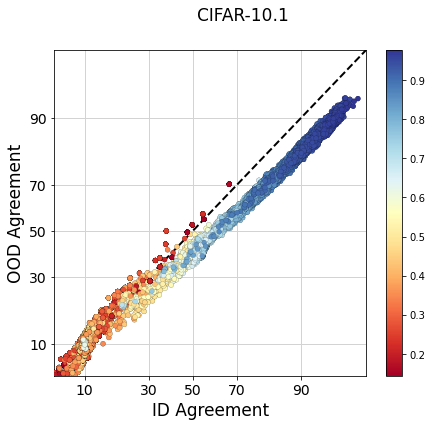}
    \includegraphics[scale=0.35]{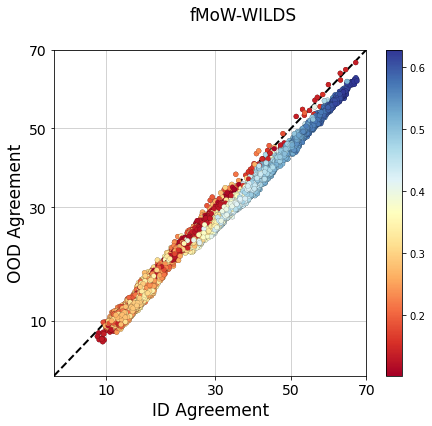}
    \caption{We plot the ID vs OOD agreement of each pair of models in the corresponding testbeds. The color of the dot represents the average ID test accuracy of the pair of models.}
    \label{fig:by_accuracy}
\end{figure}

\textit{Is agreement-on-the-line a result of neural network performing well?} In Figures \ref{fig:two_no_gde} we look at the trend between ID agreement vs accuracy to observe whether the accuracy of the models in the collection plays a role in the success of ALine-D. Note that naturally, the gap between agreement and average accuracy is smaller for pairs of models that perform well. However, this does not necessarily correspond with better ALine-D performance on such models. On the right figure of \ref{fig:two_no_gde}, we plot the right hand side of ALine-D's Equation 6 versus the left hand side ("adjusted agreement"), and observe that all pairs lie close to the diagonal, regardless of how large the gap $\mathsf{Acc}_{OOD} - \mathsf{Agr}_{OOD}$ is before the adjustment. This is to say that, all pairs of varying performance have strong linear correlation between ID vs OOD agreement, which we show in Figure \ref{fig:by_accuracy}.

\newpage
\section{Ablation Study}
\label{app:ablation}
\subsection{How many models are necessary for ALine-D to make accurate predictions?}
\begin{figure}[H]
    \centering %
\begin{subfigure}{0.45\textwidth}
  \includegraphics[width=\linewidth]{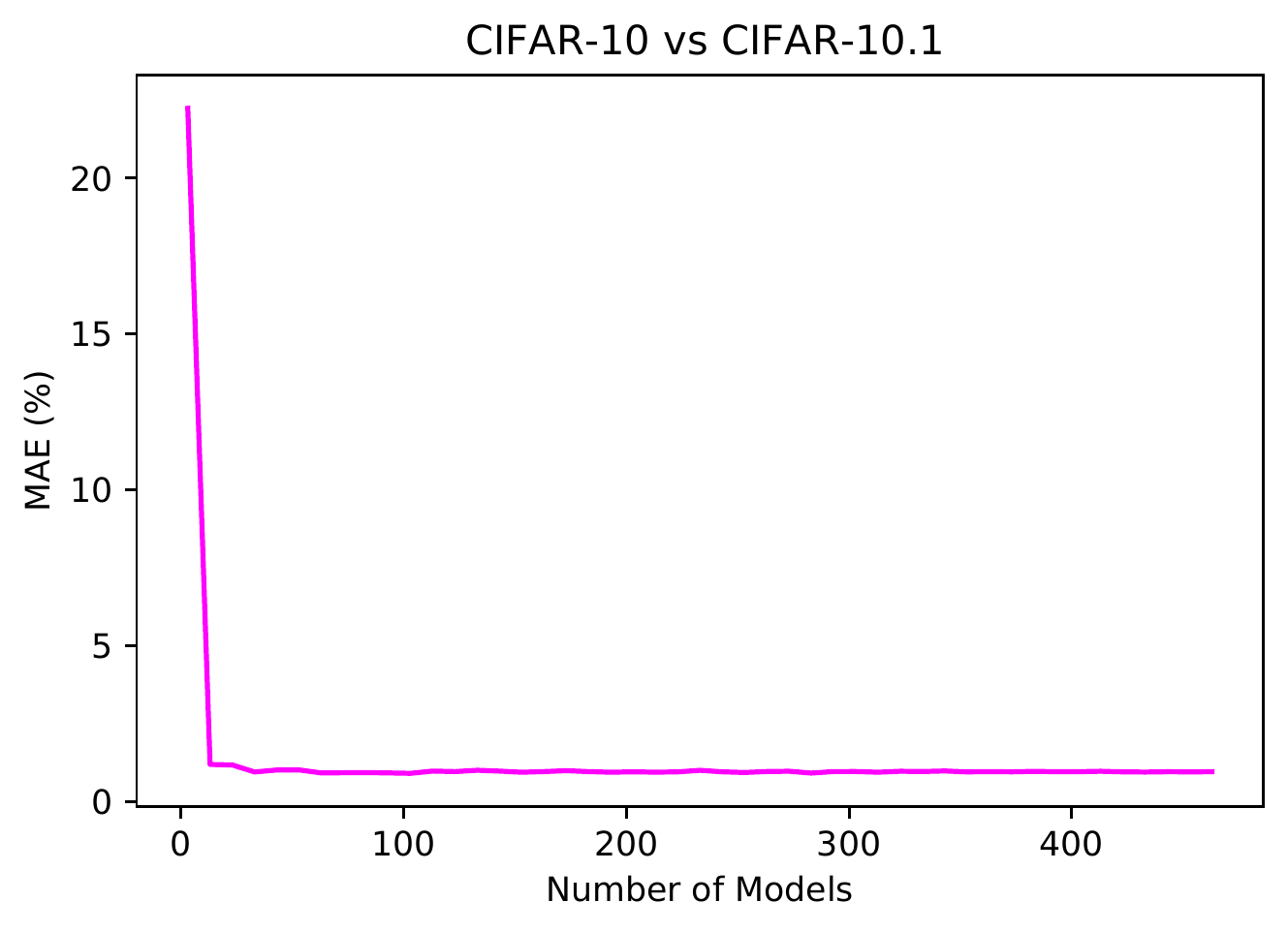}
  \caption{CIFAR-10.1}
\end{subfigure}\hfil %
\begin{subfigure}{0.45\textwidth}
  \includegraphics[width=\linewidth]{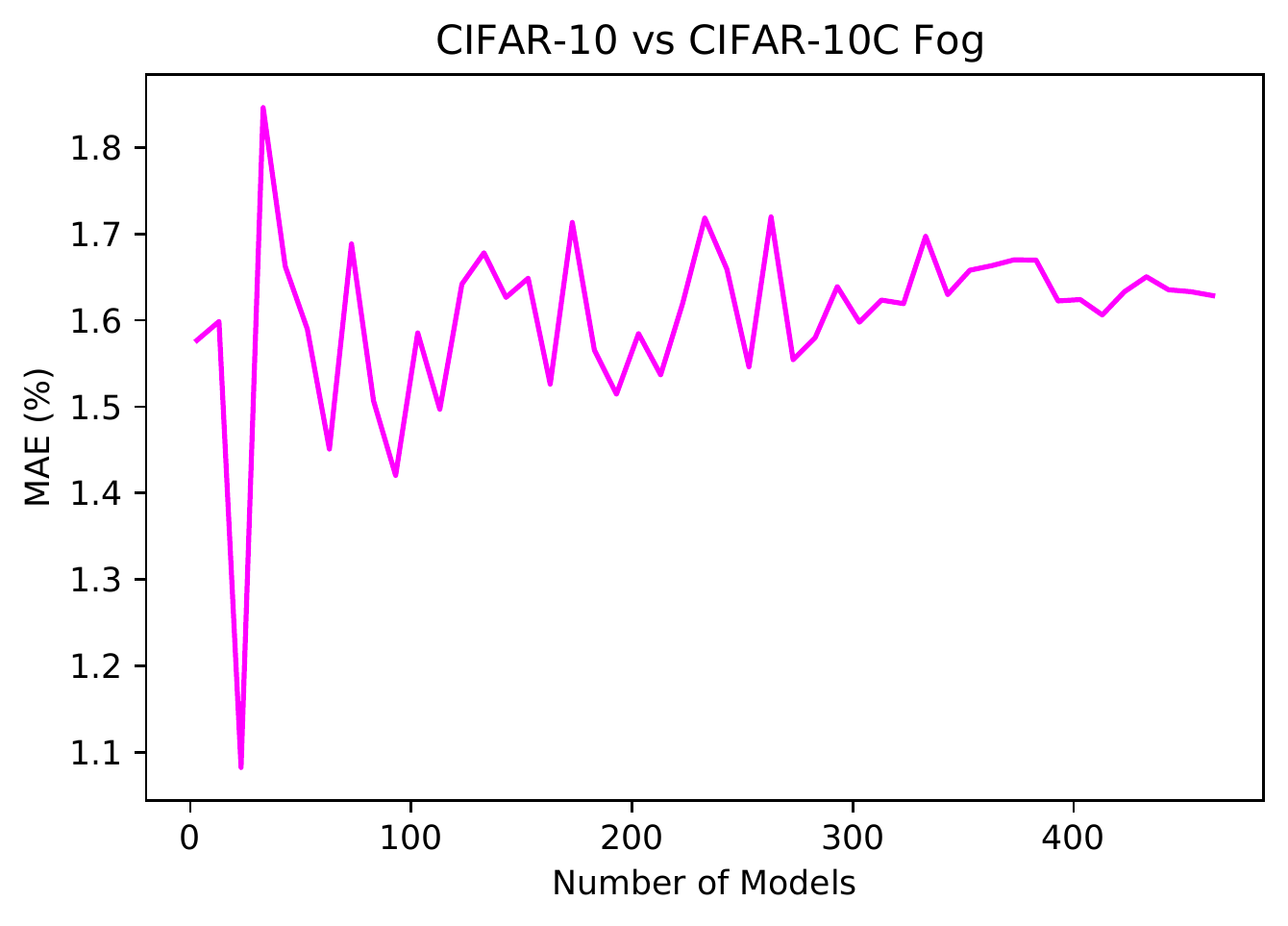}
  \caption{CIFAR-10C Fog}
\end{subfigure}\medskip %
\begin{subfigure}{0.45\textwidth}
  \includegraphics[width=\linewidth]{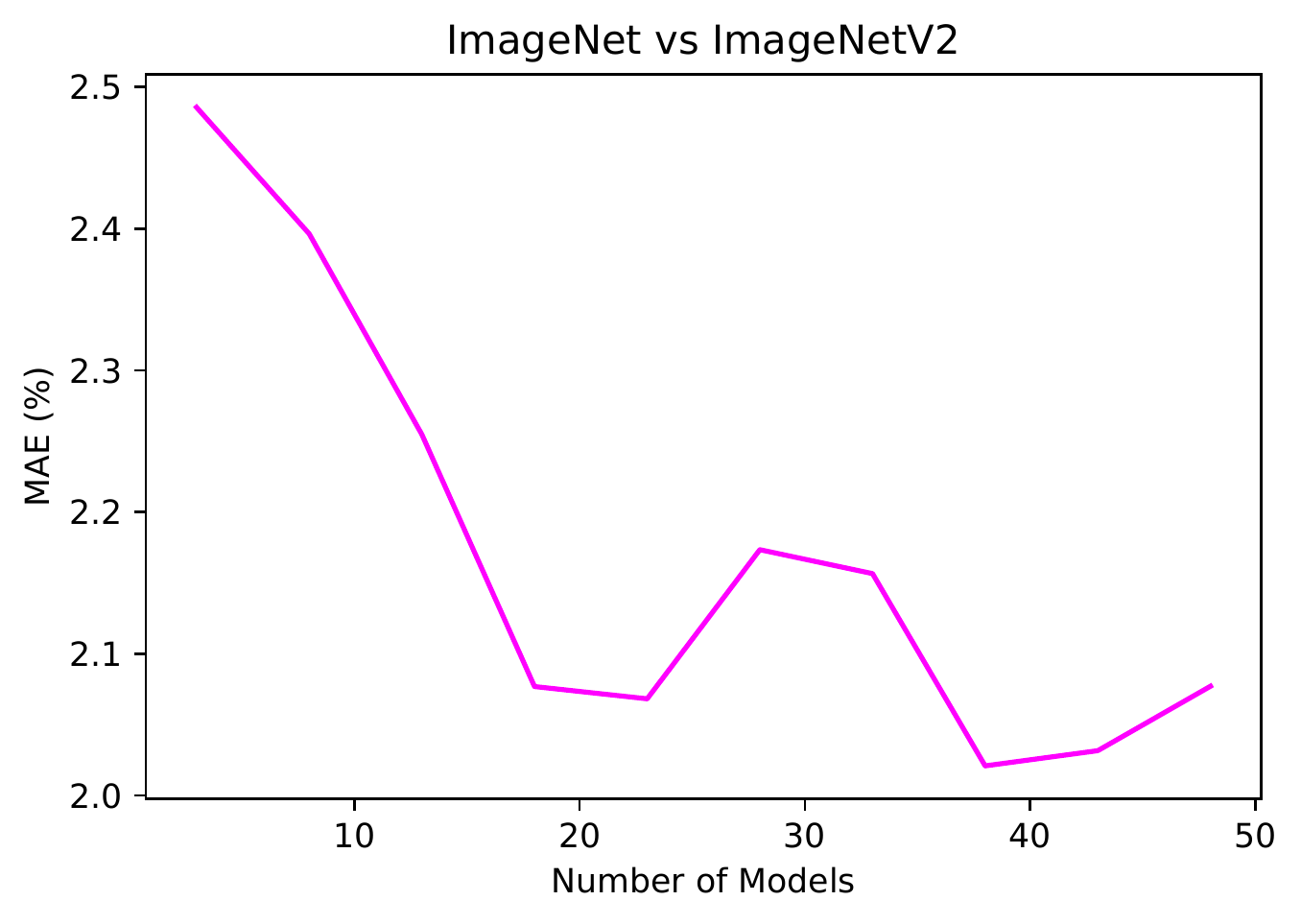}
  \caption{ImageNetV2}
\end{subfigure}

\caption{Performance of ALine-D over model sets of different sizes. We observe the MAE (in $\%$) of the ALine-D estimates of the OOD accuracy of each model in the model set. On the x axis, we vary the size of the model set. We look at CIFAR-10.1, CIFAR-10C Fog, and ImageNetV2 in particular. For each plot, we average the MAE over 3 seeds.} 
\end{figure}

ALine-D is an algorithm that requires at least 3 models in the model set so that the system of linear equations (Line 9 in Algorithm 1) has a unique solution. Additionally, it may generally require more models for the slope and intercept of the agreement trend to match the slope and intercept of the accuracy trend. We observe the MAE (in $\%$) of the ALine-D estimates of the OOD accuracy for model sets of varying sizes. For each distribution shift, we randomly sample $n$ models from the testbed to be our model set. $n$ ranges from 3 to 463 in increments of 10 for CIFAR-10 related shifts, and 3 to 48 in increments of 5 for ImageNetV2. Our ablation study shows that the success of ALine-D is not necessarily tied to the number of models. For CIFAR-10.1, we see a very quick drop in estimation error, and ALine-D performs well even for a small model set. Similarly, in ImageNetV2, we observe a decrease in estimation error as the number of models increases, however, the MAE is already pretty low from the start ($2.5\%$). On the other hand, in CIFAR-10C Fog, the estimation error does not decrease, but the error is quite low (below $1.8\%$) from the start. This short ablation study seems to indicate that ALine-D performs pretty well when agreement-on-the-line holds even for a small number of models ($<15$ models). Additionally, it is not always the case that more models will decrease the estimation error further. 

\subsection{Does a model set of varied architecture perform better than models of the same architecture?}
We study whether the diversity from varying the architecture of the models in the model set improves the performance of ALine-D. We look at the performance of ALine-D on CIFAR-10.1 and CIFAR-10C Fog over two model sets sampled from the CIFAR-10 testbed: (A) 20 PreActResNet18 \cite{preact_resnet} models, (B) 20 models of varying architecture. Similary for fMoW, we look at the performance over (A) 10 DenseNet121 \cite{densenet} models , (B) 10 models of varying architecture. We randomly sample these models from the corresponding testbeds, and average our results over 10 seeds. Our results as shown below depends on the dataset. For CIFAR-10.1 and CIFAR-10C Fog, we see that ALine-D performs better on the diverse set consisting of many architecture types. On the other hand, for fMoW, ALine-D performed better on the uniform set consisting of models of one architecture.

\begin{figure}[H]
    \centering %
\begin{subfigure}{0.3\textwidth}
  \includegraphics[width=\linewidth]{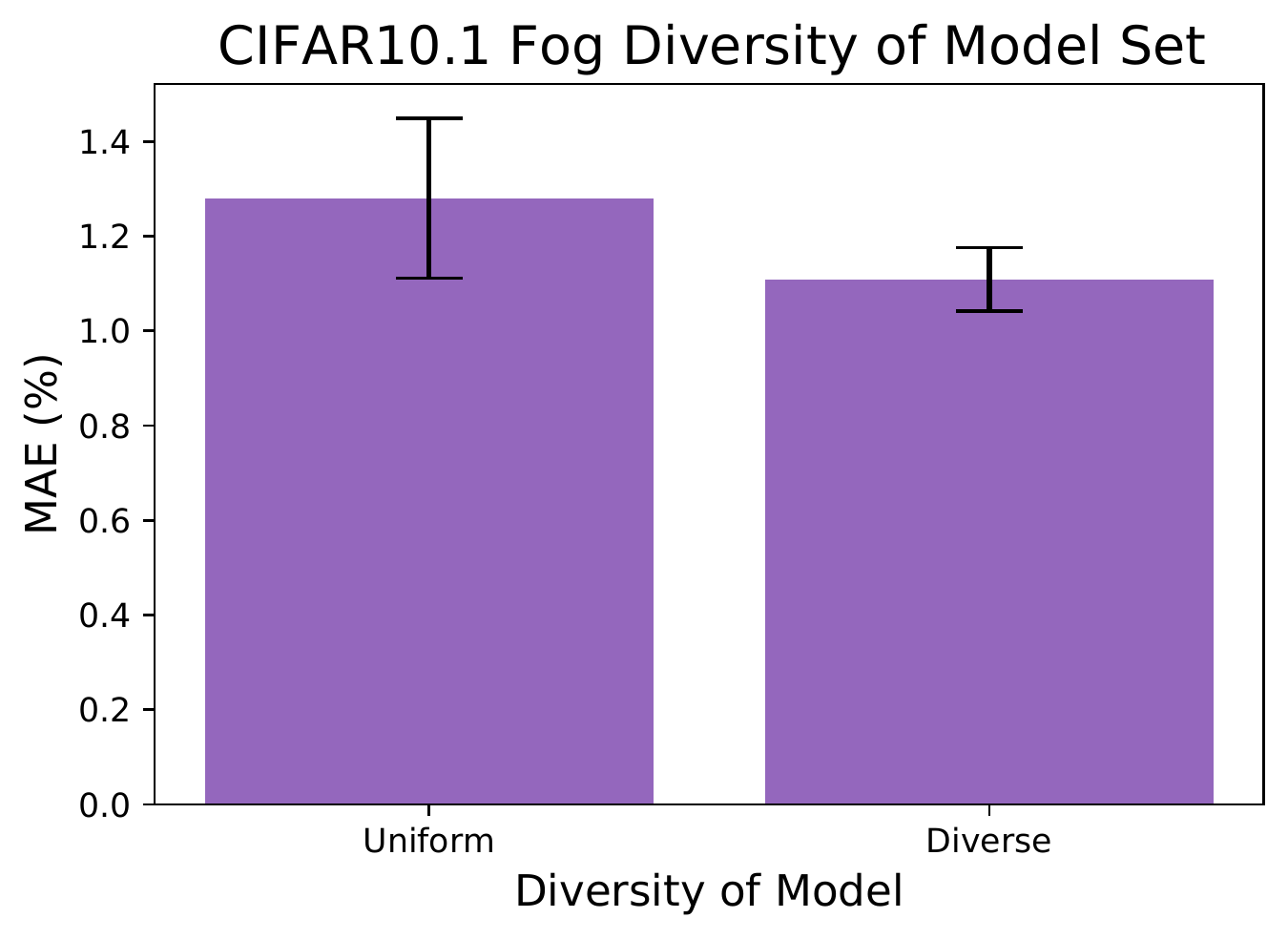}
  \caption{CIFAR-10.1}
\end{subfigure}\hfil %
\begin{subfigure}{0.3\textwidth}
  \includegraphics[width=\linewidth]{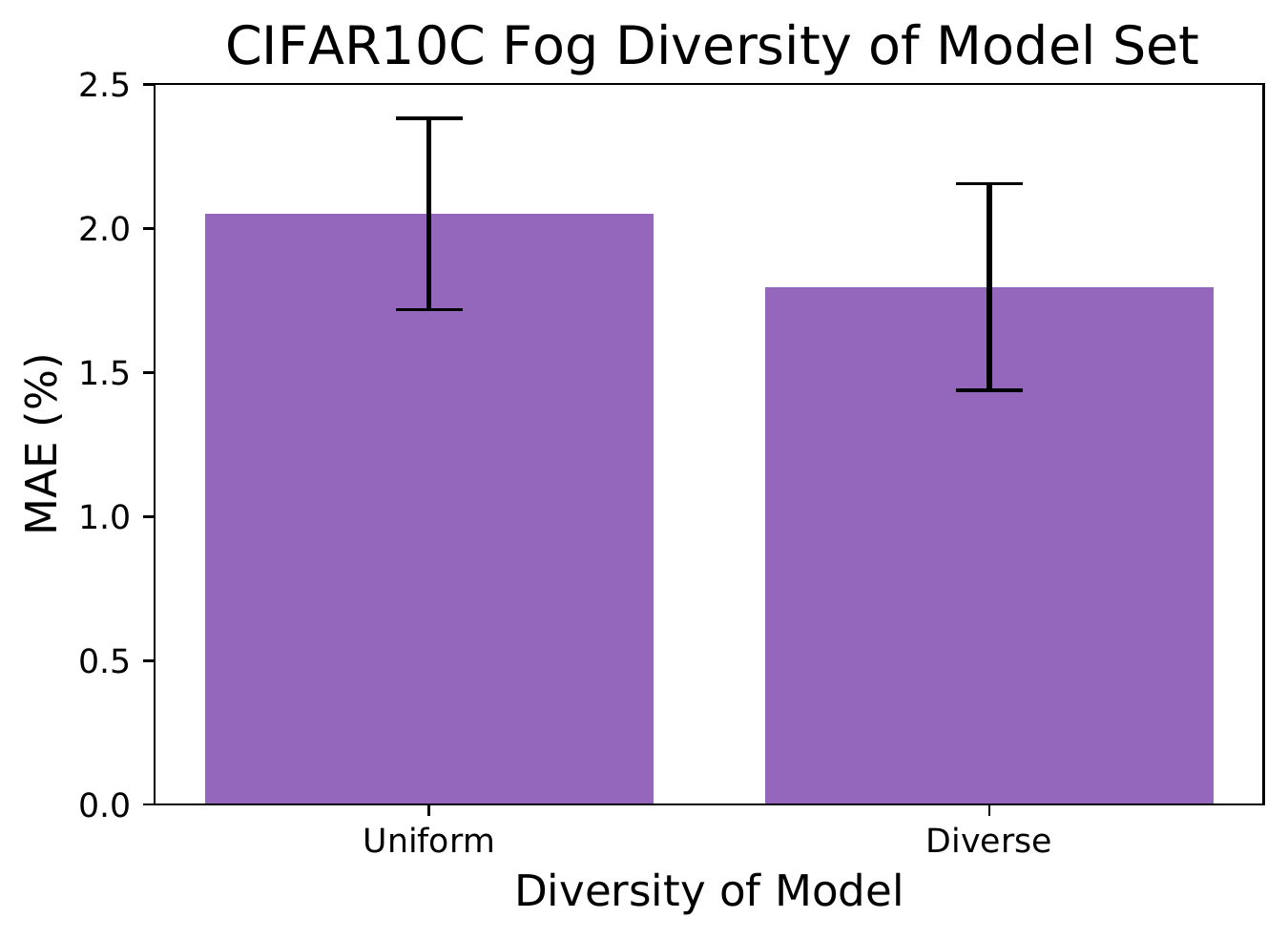}
  \caption{CIFAR-10C Fog}
\end{subfigure}\hfil %
\begin{subfigure}{0.3\textwidth}
  \includegraphics[width=\linewidth]{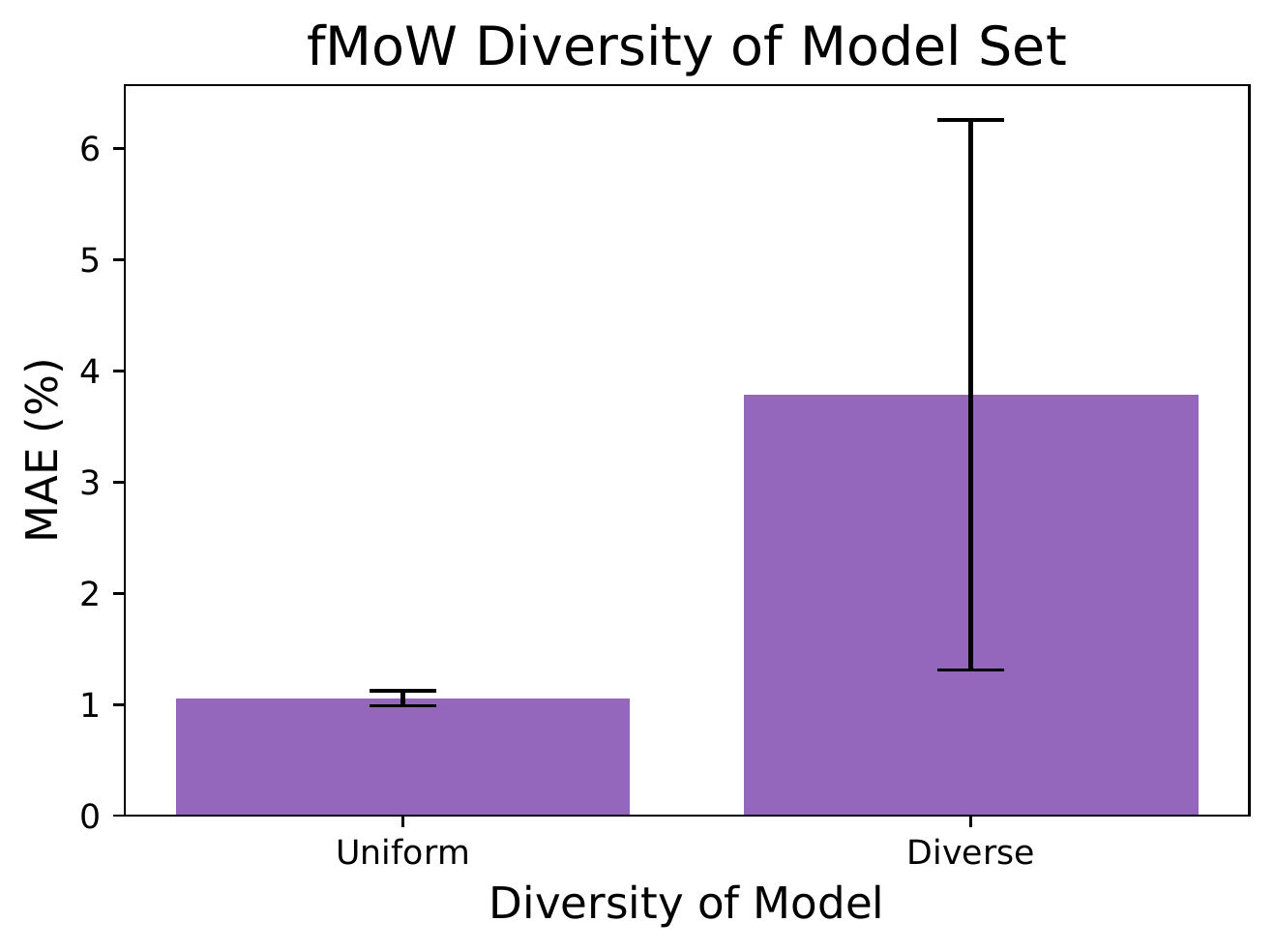}
  \caption{fMoW}
\end{subfigure}
\caption{We compare the performance of ALine-D for a model set consisting of models of many architectures versus a single architecture.}
\end{figure}

\end{document}